Advancing Applications of Satellite Photogrammetry: Novel Approaches for Built-up Area Modeling and Natural Environment Monitoring using Stereo/Multi-view Satellite Image-derived 3D Data

Dissertation

Presented in Partial Fulfillment of the Requirements for the Degree Doctor of Philosophy in the Graduate School of The Ohio State University

By

Shengxi Gui

Graduate Program in Civil Engineering

The Ohio State University

2024

Dissertation Committee

Dr. Rongjun Qin, Advisor

Dr. Alper Yilmaz

Dr. Charles Toth

**Abstract**

With the development of remote sensing technology in recent decades, spaceborne sensors with sub-meter and meter spatial resolution (Worldview & PlanetScope) have achieved a considerable image quality to generate 3D geospatial data via a stereo matching pipeline. These achievements have significantly increased the data accessibility in 3D, necessitating adapting these 3D geospatial data to analyze human and natural environments. This dissertation explores several novel approaches based on stereo and multi-view satellite image-derived 3D geospatial data, to deal with remote sensing application issues for built-up area modeling and natural environment monitoring, including building model 3D reconstruction, glacier dynamics tracking, and lake algae monitoring. Specifically, the dissertation introduces four parts of novel approaches that deal with the spatial and temporal challenges with satellite-derived 3D data.

The first study advances LoD-2 building modeling from satellite-derived Orthophoto and DSMs with a novel approach employing a model-driven workflow that generates building rectangular 3D geometry models. By integrating deep learning for building detection, advanced polygon extraction, grid-based decomposition, and roof parameter computation, we accurately computed complex building structures in 3D, culminating in the development of SAT2LoD2—a popular open-source tool in satellite-based 3D urban reconstruction.

Secondly, we further enhanced our building reconstruction framework for dense urban areas and non-rectangular purposes, we implemented deep learning for unit-level segmentation and introduced a gradient-based circle reconstruction for circular buildings to develop a polygon composition technique for advanced building LoD2 reconstruction. This approach refines building 3D modeling in complex urban structures, particularly for challenging architectural forms.

Our third study utilizes high-spatiotemporal resolution PlanetScope satellite imagery for glacier tracking at 3D level in mid-latitude regions. We developed a novel approach of using a photogrammetric workflow to generate glacier 3D terrain model and track the ice motion for various periods. By generating dense time-series 3D elevation models, we distinguished between seasonal variations and permanent glacier retreat, offering insights into glacier dynamics in response to climate change.

Finally, we proposed a term as "Algal Behavior Function" to refine the quantification of chlorophyll-a concentrations from satellite imagery in water quality monitoring, addressing algae fluctuations and timing discrepancies between satellite observations and field measurements, thus enhancing the precision of underwater algae volume estimates.

Overall, this dissertation demonstrates the extensive potential of satellite photogrammetry applications in addressing urban and environmental challenges. It further showcases innovative analytical methodologies that enhance the applicability of adapting stereo and multi-view very high-resolution satellite-derived 3D data.

# Acknowledgments

I am immensely grateful for the support and guidance I received throughout this PhD dissertation.

Firstly, I am profoundly grateful to my advisor, Dr. Rongjun Qin, whose expertise, understanding, and patience, added considerably to my graduate experience. His willingness to give his time so generously has been very much appreciated. I am indebted to him for having supported and guided me through the intricacies of the research process. I would also like to thank, the members of my dissertation committee, Dr. Alper Yilmaz, and Dr. Charles Toth, for their insightful feedback and encouragement throughout the process. Besides, a heartfelt thanks goes to Dr. Yanlan Liu, who advised my research from 2022 about postfire forest recovery. I am especially thankful for her support during my moments of doubt.

My sincere thanks also goes to my colleagues from Geospatial Date Analytics Lab and collaborators from Ecohydrology Group, German Aerospace Center (DLR), and Weavers Research Group, for their timely and invaluable advice and assistance.

I must acknowledge the profound moral and emotional support my family has provided me through my entire life, but especially during the process of completing my education. In particular, I must thank Mr. Qiang Gui and Mrs. Fang Chen, whose love and encouragement have been my sustenance. Finally, I would like to thank my girlfriend, Huizi Zeng, for all the fun and relaxation she helped bring into my life during this stressful period.

This dissertation stands as a testament to the collaborative effort of so many people, named and unnamed, and it is with heartfelt gratitude that I acknowledge their contribution.

# Vita

**2018**          B.S., Remote Sensing Science and Technology, Wuhan University

**2020**          M.S., Civil Engineering, The Ohio State University

**2020-Present**  Ph.D. student & candidate, Civil Engineering, The Ohio State University

# Publications

**Journal Paper**

[1] **Shengxi Gui**, Rongjun Qin. (2021). Automated LoD-2 model reconstruction from very-high-resolution satellite-derived digital surface model and orthophoto. *ISPRS Journal of Photogrammetry and Remote Sensing*, *181*, 1-19. ([Featured Article of the month](), Editor choice)

[2] **Shengxi Gui**, Shuang Song, Rongjun Qin, Yang Tang. (2024). Remote Sensing Object Detection in the Deep Learning Era – A Review. *Remote Sensing*, *16*(2):327.

[3] Yang Tang, Shuang Song, **Shengxi Gui**, Weilun Chao, Chinmin Cheng, Rongjun Qin (2023). Active and low-cost hyperspectral imaging for spectral analysis in low lighting environment. *Sensors*, *23*(3), 1437.

[4] **Shengxi Gui**, Rongjun Qin, Ningli Xu. Using PlanetScope-derived time-series elevation models to track fast-flowing glacier 3d dynamics. (In preparation)

[5] **Shengxi Gui**, Kaiden Murphy, Mark Tischer, Linda Weavers, Rongjun Qin. Enhanced Remote Sensing of Surface Water Chlorophyl-a: Coupling Dynamic Algae Vertical Movement Modeling with Multispectral PlanetScope Satellite Images. (In preparation)

[6] **Shengxi Gui**, Qian Zhao, Yanlan Liu. (2024). Decadal legacy effect of fires on the spatial structure of forests across CONUS. (In preparation)

**Conference Paper**

[1] **Shengxi Gui**, Philipp Schuegraf, Ksenia Bittner, Rongjun Qin (2024). Unit-level LoD2 building reconstruction from satellite derived digital surface model and orthophoto. *ISPRS. Annals. Photogramm. Remote Sens. Spatial Inf. Sci*. (Accepted)

[2] **Shengxi Gui**, Yanlan Liu. (2023). Decadal Legacy Effect of Fires on the Spatial Structure of Forests Across CONUS. In *AGU Fall Meeting Abstracts* (AGU 2023).

[3] **Shengxi Gui**, Rongjun Qin. (2023). Using PlanetScope-Derived Time-Series Elevation Models to Track Surging Glacier 3D Dynamics in Mid-Latitude Mountain Regions. In *AGU Fall Meeting Abstracts* (AGU 2023).

[4] **Shengxi Gui**, Rongjun Qin, Yang Tang. (2022). SAT2LOD2: a software for automated lod-2 building reconstruction from satellite-derived orthophoto and digital surface model. *The International Archives of Photogrammetry, Remote Sensing and Spatial Information Sciences*, *43*, 379-386.

[5] Ying Zuo, CK Shum, Rongjun Qin, Yuanyuan Jia, Guixiang Zhang, **Shengxi Gui**. (2022). Pan-India Land Use Land Cover Deep Learning Aided Classification Using NICFI Products. In *AGU Fall Meeting Abstracts* (Vol. 2022, pp. G31A-02).

Fields of Study

Major Field: Civil Engineering

Studies in:

    Topic 1: Photogrammetry and Remote Sensing Applications

    Topic 2: LoD-2 level Building 3D model Reconstruction

    Topic 3: Glacier 3D Dynamic Tracking

    Topic 4: Algae Bloom Monitoring via Remote Sensing

# Table of Contents

# List of Tables

## List of Figures

**Chapter 1. Introduction**

This chapter cover then motivation of work presented in this dissertation. Firstly, **Section 1.1** covers the motivations behind the work and challenges for the topic 3D modeling in natural and human environments. Moreover, **Section 1.2** introduces the main research questions and research scopes of this dissertation. The contributions of dissertation are presented in **Section 1.3**. Finally, **Section 1.4** summarizes the structure and workflow of the rest of dissertation.

**1.1 Motivation**

With the development of remote sensing technology in recent decades, spaceborne sensors with sub-meter spatial resolution (World-View2-4, Pleiades, etc.) achieved a comparable image quality to airborne images a few decades ago. With these sensors running 24/7, the volume of data has dramatically grown to the extent that automatic image interpretation, or modelling our environment in 3D level, is becoming necessary. This geographically referenced information has become crucial in numerous civilian applications, offering low-cost, accurate representations of both human-made and natural environments across spatial and temporal dimensions. Furthermore, 3D photogrammetric data, essential for digitizing the environment, are commonly produced through dense stereo matching techniques applied to multi-view optical remote sensing images. Nevertheless, the inherent redundancy in original 3D geospatial data poses significant challenges for their quantitative analysis and environmental modeling. This dissertation outlines three research endeavors designed to streamline environmental modeling both spatially and temporally. These include detecting and vectorizing relevant objects from geospatial datasets and quantifying their changes over time. Addressing the difficulties in modeling various environments, our approach utilizes satellite imagery of very high resolution (Ground Sampling Distance < 5m) to reconstruct detailed 3D surface models, facilitating analysis across different applications.

**LoD2 building reconstruction from satellite-derived data**

Leveraging high-resolution DSMs from satellite imagery, we can generate Level-of-Detail 1 (LoD-1) or even higher-level 3D building models. However, achieving LoD-2 reconstruction remains a significant challenge, mainly due to unclear building boundaries in terms of height and color, as well as the need for robust, model-driven methodologies. Moreover, addressing the lack of accessible, lightweight tools for LoD-2 reconstruction, we try to develop a novel approach and open-source software framework, aimed at enhancing building detection and 3D model reconstruction for satellite-derived DSM and orthophotos, thereby advancing automated LoD-2 modeling techniques.

**Complex and non-rectangular buildings in dense urban region LoD2 reconstruction**

The building reconstructing for duplex and complex buildings in dense building areas remains significant challenges due to segmentation difficulties in distinguishing individual units within weak texture satellite imagery. Duplexes, with separate units sharing walls, and complex buildings, comprising rectangular units with similar features, demand advanced segmentation techniques for accurate LoD2 modeling. Furthermore, it is necessary incorporating buildings with unconventional shapes like circles into 3D models enhances structural representation but is hindered by indistinct edge features and incomplete forms in satellite imagery, necessitating innovative approaches for identification and modeling.

**Glacier 3D dynamics tracking with high resolution optical satellite imagery**

Quantitative evaluation of large-scale glacier dynamics is crucial for understanding global climate change. There is a challenge of monitoring fast-moving glaciers, where optical satellite data is vital yet often insufficient for dynamic analysis within short time revisit. Previous studies have either focused on 2D movement using 10m or lower resolution optical imagery or generated 3D models with SAR, lacking the 3D representative with very high resolution. Additionally, research typically covers short periods, making it difficult to assess long-term trends, including whether volume changes are periodic or indicative of permanent retreat, and the specific climate indicators influencing glacier motion remain to be clarified. Thus, we are going to develop a photogrammetric pipeline to generate a time-series 3D observation of fast-flowing glacier by using very-high resolution satellite images.

**Algae monitoring with remote sensing imagery at lakes**

Remote sensing has emerged as an efficient tool for monitoring algae, offering a way to observe and analyze algal blooms over extensive areas and inaccessible locations. Addressing the challenges of accurately monitoring algae through remote sensing involves recognizing the diurnal vertical movement of algae, as surface Chl-a levels can significantly fluctuate within one day. This fluctuation challenges traditional remote sensing methods, which does not deal with the temporal discrepancies between field and satellite data, indicating models for these movements can be developed for better prediction accuracy. Additionally, monitoring during non-bloom periods presents its own set of challenges due to weak signals in satellite imagery, necessitating advanced remote sensing techniques to capture the full algal growth cycle.

**1.2 Research Scope**

This dissertation aims to develop different approaches to enhance and solve the satellite photogrammetry applications and challenges mentioned above (**Section 1.1**). The research and scientific questions include:

- Is it possible to build up an automatic workflow to generate rectangular buildings LoD2 reconstruction only depends on Orthophoto and DSM? Can this workflow be publicly accessible as open-source tool? (Chapter 2)
- How can we decompose complex buildings into several fundamental models? (Chapter2)

- How to deal with the building reconstruction issue for duplex and very complex building in the dense urban area? (Chapter 3)

- How to reconstruct non-rectangular buildings at 3D level including circular buildings and irregular buildings? (Chapter 3)

- How can we generate high spatial-temporal-resolution 3D models based on optical stereo images for interesting glacier regions? Then how to develop an efficient glacier motion tracking method? (Chapter 4)

- What trend of dynamics of glaciers we can discover based on glacier motion? Are the dynamics associated with climate? (Chapter 4)

- Do algae in lakes have diurnal movement? Is this movement impact Chl-a estimation model developed based on field measurement and remote sensing imagery? And how to describe this potential movement with time? (Chapter 5)

- If we consider the time misalignment issue for remotely sensed Chl-a estimation model, how can we refine the model to improve the accuracy for regression? Does it benefit for the underwater algae monitoring? (Chapter 5)

**1.3 Contribution**

Our research has made a progress on using stereo/multi-view satellite image-derived 3d data to solve issues in remote sensing applications in human and natural environment including urban, ice-covered mountain, and lake areas. These contributions provide an interdisciplinary sight to using satellite photogrammetry in practical use:

For **Building LoD2 reconstruction**, we developed a novel model-driven approach for reconstructing LoD-2 building models from satellite-derived DSMs, overcoming the limitations of existing methods that primarily achieve LoD-1 generation. Our approach features a unique "decomposition-optimization-fitting" paradigm, starting with deep learning-based building detection and advancing through a three-step polygon extraction and grid-based decomposition to accurately model complex and irregular building shapes. Furthermore, incorporating OpenStreetMap and Graph-Cut labeling for orientation refinement, and employing building-specific parameters for 3D modeling, our method significantly enhances the adaptability and quality of satellite-based building reconstruction in LoD2 level. And we also developed the first and most popular satellite-based 3D building reconstruction tool – SAT2LoD2.

For **Unit-level and circular building reconstruction**, we proposed deep learning for unit-level building segmentation from high-resolution satellite imagery, enhancing LoD-2 modeling with a polygon composition technique that regularizes noisy segmentation results for accurate boundary consistency among complex building units. Moreover, we introduced a gradient-based circle detection method for the reconstruction of circular buildings, elevating the detail and accuracy of urban 3D reconstruction. These two contributions enhance the LoD2 building reconstruction workflow in challenge cases.

For **Glacier 3D dynamics tracking**, we utilized time-series PlanetScope satellite imagery to monitor glacier dynamics in mid-latitude mountain regions, distinguishing seasonal variations from permanent glacier retreats. By generating dense and time-series 3D elevation models from 2019 to 2023 for La Perouse, Viedma, and Skamri Glaciers, we developed a novel approach for glacier tracking with photogrammetric workflow, and offer new insights into glacier flow velocity and volume change patterns. Our study benefits high-temporal resolution satellite data in advancing the analyzing of glacier responses to climate change.

For **Algae monitoring considering time effect**, we developed a chlorophyll-a (Chl-a) quantification refinement strategy from satellite imagery by introducing an algal behavior function that corrects for algae fluctuations and timing mismatches between satellite captures and field measurements. This novel approach significantly improves prediction accuracy, especially in high-resolution studies of non-bloom seasons, and precisely estimates underwater algae volumes to get a complete estimation for overall biomass computation in lake ecosystem.

**1.4 Organization**

in this dissertation, we have developed several novel approaches for satellite photogrammetry application issues for human and nature environments. The following structure of this dissertation is: Chapter 2 discusses our novel approach and entire workflow for building 3D reconstruction based on satellite stereo image data, which include building segmentation, 2D footprint generation, and 3D reconstruction. Chapter 3 introduces unit-level segmentation for dense urban building reconstruction, and circular building reconstruction. Chapter 4 mainly focuses on glacier dynamics tracking, introduce a photogrammetric pipeline, and time-series glacier dynamics and their drivers. Chapter 5 discusses our novel approach of using time effect between algae field measurement and satellite imagery to improve the Chl-a estimation methods. Finally, Chapter 6 conclude our dissertation with conclusion and discuss the limitation and future work to further improve satellite photogrammetry applications.

**Chapter 2. LoD-2 Building Model Reconstruction from Satellite-derived Digital Surface Model and Orthophoto**

This chapter is based on the paper "Automated LoD-2 Model Reconstruction from Satellite-derived Digital Surface Model and Orthophoto" that was published in the "ISPRS Journal of Photogrammetry and Remote Sensing (2021)" by Shengxi Gui, Rongjun Qin as Editor Choice paper, and paper "SAT2LOD2: a software for automated lod-2 building reconstruction from satellite-derived orthophoto and digital surface model" that was published in the "The International Archives of Photogrammetry, Remote Sensing and Spatial Information Sciences (2022)" by Shengxi Gui, Rongjun Qin, Yang Tang. **Figure 2.1** shows the main output of building reconstruction from satellite-derived DSM and orthophoto.

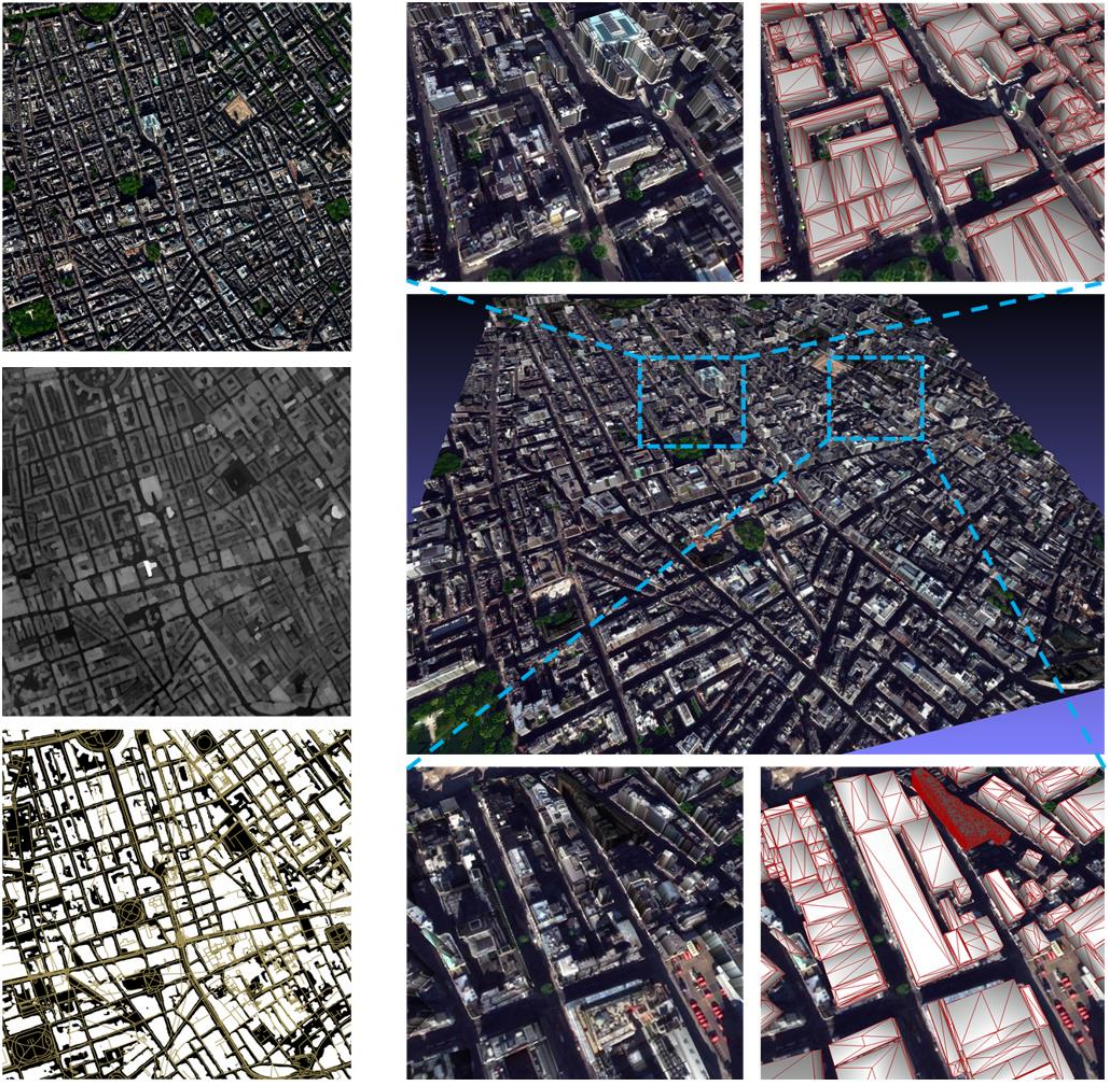

**Figure 2.1** Example of LoD2 building reconstruction. First column from up to bottom: Orthophoto, DSM, building segmentation map using HRNet, with OSM street lines. Second column: 3D building model and detailed view of two specific areas

## 2.1 Chapter Abstract

Digital surface models (DSM) generated from multi-stereo satellite images are getting higher in quality owning to the improved data resolution and photogrammetric reconstruction algorithms. Satellite images effectively act as a unique data source for 3D building modeling, because it provides a much wider data coverage with lower cost than the traditionally used LiDAR and airborne photogrammetry data. Although 3D building modeling from point clouds has been intensively investigated, most of the methods are still ad-hoc to specific types of buildings and require high-quality and high-resolution data sources as input. Therefore, when applied to satellite-based point cloud or DSMs, these developed approaches are not readily applicable and more adaptive and robust methods are needed. As a result, most of the existing work on building modeling from satellite DSM achieves LoD-1 generation. In this paper, we propose a model-driven method that reconstructs LoD-2 building models following a "decomposition-optimization-fitting" paradigm. The proposed method starts building detection results through a deep learning-based detector and vectorizes individual segments into polygons using a "three-step" polygon extraction method, followed by a novel grid-based decomposition method that decompose the complex and irregularly shaped building polygons to tightly combined elementary building rectangles ready to fit elementary building models. We have optionally introduced OpenStreetMap (OSM) and Graph-Cut (GC) labeling to further refine the orientation of 2D building rectangle. The 3D modeling step takes building-specific parameters such as hip lines, as well as non-rigid and regularized transformations to optimize the flexibility for using a minimal set of elementary models. Finally, roof type of building models are refined and adjacent building models in one building segment are merged into the complex polygonal model. Our proposed method have addressed a few technical caveats over existing methods, resulting in practically high-quality results, based on our evaluation and comparative study on a diverse set of experimental dataset of cities with different urban patterns.

**2.2 Introduction**

Remotely sensed satellite imagery effectively acts as one of the preferred ways to reconstruct wide-area and low-cost 3D building models used in urban and regional scale studies (Leotta et al. 2019; Brown et al. 2018; Facciolo, de Franchis, and Meinhardt-Llopis 2017). However, because satellite imagery has a relatively lower spatial resolution than aerial imagery and LiDAR, 3D building model base on satellite data faces challenges in detecting buildings, extracting building boundaries, and reconstructing an accurate building 3D model, especially in regions with small and dense buildings (Sirmacek and Unsalan 2011). Level-of-Detail (LoD) building models defined through the city geography markup language (CityGML), present 3D building model in several levels 0 to 4 (Gröger et al. 2008). An improved LOD standard CityGML 2.0 speciate LoD-0 to LoD-3 as four sub-definition from LOD x.0 to LOD x.3 base on the exterior geometry of buildings (Gröger and Plümer 2012; Biljecki, Ledoux, and Stoter 2016). With relatively fine digital surface models derived from satellite photogrammetry, using satellite-based DSM and building polygon to generate Level-of-Detail 1 (LoD-1) 3D building model with a flat roof is now becoming a standard practice. However, 3D building model reconstruction with prototypical roof structures (LoD2), remains a challenging problem, especially from low-cost data sources such as satellite images (Kadhim and Mourshed 2017; Bittner and Korner 2018). Despite the increasingly higher resolution of satellite images (as high as 0.3-0.5 GSD) on par with past aerial images, the generated 3D information from the satellite images, due to 1) image quality (being high altitude collection distorted by the atmosphere), 2) relatively low resolution in comparison to aerial data, and 3) cross-track collections (Rongjun Qin 2019) with temporal variations, often posses high uncertainties leading to challenges in LoD-2 modeling. Although 3D building modeling from point clouds has been intensively studied, most of the methods are still ad-hoc to specific types of buildings and assume these input point clouds to be highly accurate and dense (i.e. those captured with LiDAR). However, when presented with low-resolution point clouds with relatively high uncertainties as produced by satellite images, these methods no longer produce reasonable results especially for areas with dense buildings, as traditional bottom-up approaches are not able to identify and separate between individual buildings given the low resolution and blurry boundaries both in height (from digital surface models (DSM) or point clouds) and in color (or spectrum), therefore it requires approaches to be sufficiently robust to identify, regularize and extract individual buildings, often following a top-down approach (i.e. model-driven). To extract individual and well-delineated boundaries of the building, one often needs to decompose "overestimated" building footprint candidates, using regularity, spectral or height cues (Brédif et al. 2013; Partovi et al. 2019), and adapt models that are well-suited to the extracted/decomposed footprints. However, among the limited studies focusing on LoD-2 reconstruction based on satellite-derived DSMs (Alidoost, Arefi, and Tombari 2019; Partovi et al. 2019; Woo and Park 2011), caveats exist in the decomposition procedure, which often do not fully explore the use spectrum, DSM and contextual information, resulting in unsatisfactory results, especially in areas where buildings are dense: on one hand the individual buildings are often incorrectly extracted/oriented, on the other hand buildings in clusters are not consistent.

In this paper, we revisit this process and fill a number of gaps by developing a three-step approach that specifically aims to improve the extraction of building polygon and fitting models in such challenging cases, which formed our proposed LoD-2.0 model reconstruction pipeline for satellite-derived DSM and Orthophotos: our proposed method starts with building mask detection by using a weighted U-Net and RCNN (region-based Convolutional Neural Networks), followed by the proposed three-step approach through "extraction-decomposition- refinement" for regularized 2D building rectangle generation. In the "extraction" step, we vectorize building masks as boundary lines, and then regularize lines orientation through line segments from orthophoto. In the "decomposition" step, a grid-based building rectangle generating method is developed by a grid pyramid to generate rectangles and subsequent separating and merging steps to optimize polygons. In the "refinement" step, we post-refine the 2D model parameters through propagating similarities of neighboring buildings (in terms of their orientation and type) through a Graph-Cut (GC) algorithm, with optionally a second refinement step using OpenStreetMap road vector data. The individual building rectangle and the corresponding DSMs are then fitted by taking models from a pre-defined model library that minimizes the difference between the model and the DSM. In this process, the motivation is to fully utilize the spectral and height information when performing the decomposition process and take global assumptions on orientation and building type consistencies of the building clusters to yield results that are superior to state-of-the-art methods. The proposed method is validated using a diverse set of data and although our approach follows and refines the existing modeling paradigm, we find the proposed approach yields robust performance on various types of data and is able to reconstruct areas with dense urban buildings.

This therefore leads to our contributions in this paper:

1) We present a model-driven workflow that performs LoD-2 model reconstruction that yields highly accurate results as compared to state-of-the-art methods;
2) We demonstrate that the use of a combined sematic segmentation and Region-based CNN (RCNN) leverages detection and completeness for object-level mask generation;
3) We propose a novel three-stage ("extraction-decomposition-refinement") approach to perform vectorization of building masks that yields superior performance;
4) We validate that the use of multiple cues of neighboring buildings, and optionally road vector maps, can generally improve the accuracy of the resulting reconstruction.

Besides, due to the implementation triviality and logistics of different libraries, to the authors' knowledge, there are no lightweight open-source tools that often have the capability to reconstruct LoD2 models from Satellite-based products. As a result, researchers either build their flow using a chain of open or commercial tools, with sometimes manual operations, or use open-source tools designed for LiDAR data (Nan and Wonka 2017; Shah, Bennamoun, and Boussaid 2017) to reconstruct building models from satellite data. Thus, a compact open-source tool that researchers can perform a "quick and dirty" test, may greatly facilitate researchers who develop relevant methods.

To this end, we developed SAT2LoD2 based on our previously published work (Shengxi Gui and Qin 2021), a top-down building model reconstruction approach. The aim is to close this community gap and put these complicated processing steps to drive the development of automated LoD2 modelling approaches. SAT2LoD2 is presented as a GUI-based open-source prototype software based on Python that turns an orthophoto and DSM into individual LoD2 building models. This software has a built-in semantic segmentation module to predict building masks but can optionally take customized classification maps. It also implements a scheme to utilize road networks (in shapefile) to improve the reconstruction results. During processing, SATR2LoD2 remains all intermediate process output so that users can exploit building models for various purposes. The codes of the tool are available on GitHub: https://github.com/GDAOSU/LOD2BuildingModel.

**2.3 Related works**

There are many approaches for the detection and reconstruction of 3D building model from airborne photogrammetry and LiDAR data (Cheng et al. 2011; R. Wang 2013), and relatively few on the satellite-derived data (Alidoost, Arefi, and Tombari 2019; Woo and Park 2011) given that dense matching method yielding relatively high quality data was only getting advanced in recent years (Bosch et al. 2016; Leotta et al. 2019; Rongjun Qin 2016a; Rongjun Qin et al. 2019), and SAR data contributes to LOD-1 building reconstruction in state and national scale recent years (Geiß et al. 2019; K. Li et al. 2020; Frantz et al. 2021). Most of satellite-based methods can be classified into three categories: data-driven, model-driven, and hybrid approaches. In data-driven approaches, buildings are assumed to be individual parts of roof planes, by considering the geometrical relationship of point, lines, and surface from DSMs and point cloud. In model-driven approaches, buildings DSMs or point cloud are compared with 3D building model in the model library, to select the most appropriate fitting model and the best parameters of this model. In hybrid approaches, both former approaches have been included that data-driven approaches extract the geometric feature (line, plane) of the building model, and model-driven approaches compute the model parameters and reconstruct the 3D model.

Typical LoD-2 model reconstruction approaches using satellite images take the following steps: The initial step detects and segments building areas from either images or combined images and orthophotos (Qin and Fang 2014). Traditional approaches for building detection used Support Vector Machine (SVM, Gualtieri & Cromp, 1999) to classify building classes as to explore sparse labels, and spatial features exploiting region-wise information such as Length-Width Extraction Algorithm (LWEA, Shackelford & Davis, 2003) are stacked into the feature vectors for classification. Recently, the deep learning models are often used to perform so-called semantic segmentation to fully explore the growingly available labeled satellite datasets, among which U-Net (Ronneberger, Fischer, and Brox 2015) is often the baseline to start with, which predicts labels for every pixel. On the other hand, instance-level prediction is of relevance, as detectors capable of detecting individual buildings may move a step further for building modeling. The fast region-based Convolutional Network (Fast R-CNN, Girshick, 2015) belongs to this class of approach, which took a window-based approach to efficiently identify regions (bounding boxes) containing objects of interest, and its advanced version, the Mask R-CNN (He et al. 2017) concurrently identifies the regions (bounding boxes) of the individual object of interest, as well as performing pixel-level labeling within each bounding box, which normally serves as a baseline approach for any type of detection tasks, which a few variants of this type of approaches are available as well (Cai and Vasconcelos 2019; Y. Zhang et al. 2020).

Once initial building masks are extracted, the next logical step is to vectorize the masks such that the boundaries of individual buildings are modeled with regularized polylines (hereafter we call it building boundaries). Initial steps to pre-processing the detected masks can be to use shape reconstruction methods such as alpha-shape (Kada and Wichmann 2012) or gift wrapping algorithm (Lee et al. 2011) to obtain initial boundaries, which might be followed by line fitting methods to further simplify the lines of the boundaries by using for example, random sampling consensus (RANSAC, (Schnabel, Wahl, and Klein 2007)), least square line fitting (O'Leary, Harker, and Zsombor-Murray 2005), and Douglas–Peucker algorithm (Douglas and Peucker 1973). The processes can be aided by extracting lines directly from orthophotos, such as using LSD (Line Segment Detector, Von Gioi et al. 2008) and KIPPI (KInetic Polygonal Partitioning of Images, Bauchet and Lafarge 2018) algorithms. The proposed method adopts a pipeline based on the Douglas-Peucker algorithm to initially extract building polygons from the mask and uses LSD to refine boundary lines. Another refinement method directly refines the orthogonal boundary lines from Mask R-CNN polygon (Zhao et al. 2018). With refined building boundaries, regularizations or decompositions can be performed to further identify individual buildings for fitting preliminary building models: for example, Partovi et al. (2019) proposed an orthogonal line-based 2D rectangle extraction method that assumed orthogonality and parallelism of building footprint, which aims to decompose a tentative building footprint to rectangle shapes by starting from the longest boundary lines. Once these individual rectangle shapes are extracted, merging operation might be needed to obtain the final simplified shapes (Brédif et al. 2013). Often this process can be aided with supplementary information such as OpenStreetMap or Ordnance Survey data (Haklay 2010) when those with sufficient quality are available. Furthermore, Girindran et al. (2020) proposed a 3D model generation approach use open-source data directly, including OSM and Advanced Land Observing Satellite World 3D digital surface model (AW3D DSM).

The model fitting for satellite-derived data, is usually carried out by fitting a few preliminary models in a model library, based on the resulting 2D building rectangle and the available DSM, and these models normally consist of a few parameters to allow efficient fitting. Girindran et al. (2020) calculated the roof components parameters using an exhaustive search to fit the DSMs. The proposed approach follows a process of a defined model library and model parameters exhaustive search. Other than exhaustive search, (Alidoost, Arefi, and Tombari (2019) presented a deep learning-based approach to predict nDSM and roof parameters of roof based on multi-scale convolutional–deconvolutional network (MSCDN) from the aerial RGB images. Given the low-resolution (and potentially high uncertainty of the 3D geometry), a post-processing to recover complex roofs such as merging of falsely separated building components through for example, eave (Brédif et al. 2013) or ridgeline (Alidoost, Arefi, and Tombari 2019) can be considered. This step is often ad-hoc with different levels of regularization depending on the accuracy of the detected 2D building footprint, complexity of the model and accuracy of the 3D geometry (i.e. DSM). In recent years, end-to-end methods create a solution to directly derive building models from the remotely sensed image. Some of these steps may involve deep learning models for object recognition and meshing (Qian, Zhang, and Furukawa 2021; Y. Wang, Zorzi, and Bittner 2021; W. Li et al. 2021).

## 2.4 Methodology

**Figure 2.2** presents the workflow of our method. We follow the typical reconstruction paradigm as introduced in **Section 2.4**, with a produced DSM and Orthophoto (pre-processing), the process consists of six main steps: 1) building segmentation, 2) building polygon extraction, 3) building rectangle decomposition, 4) orientation refinement, 5) 3D model fitting, and 6) model post-refinement. The input data of our method is DSM and the corresponding orthophoto, which is generated by using a workflow based on RPC stereo processor (RSP) that includes sequentially a level 2 rectification, geo-referencing, point cloud generation, DSM resampling, and orthophoto rectification (Rongjun Qin 2016a). The building segmentation from orthophoto is developed using a combined U-Net based semantic labeling and Mask R-CNN segmentation algorithm, and then further study adapt HRNet-based approach. To extract building polygon, we propose a novel method that first detects the rough boundary of building segments using Douglas–Peucker algorithm and then adjusts and regularizes the boundary lines combining orientation from LSD. Next, to generate and decompose the basic 2D building model from building polygon, especially in the polygon with several individual small buildings, a novel grid-based decomposition method is developed to generate and post-process the building rectangles with serval sub-rectangles in a building segment. The building rectangles are subject to an orientation refinement process using GC optimization and optionally a rule-based adjustment using OpenStreetMap (OSM). LoD-2 model reconstruction is performed by fitting to the most appropriate roof model in a building model library through optimizing the shape parameters that yields minimum RMSE with the original DSM. In the last step, model roof type is refined using GC optimization, and simple building models are merged as complex models based on a few heuristics. The detailed components of each step are given in the following sections.

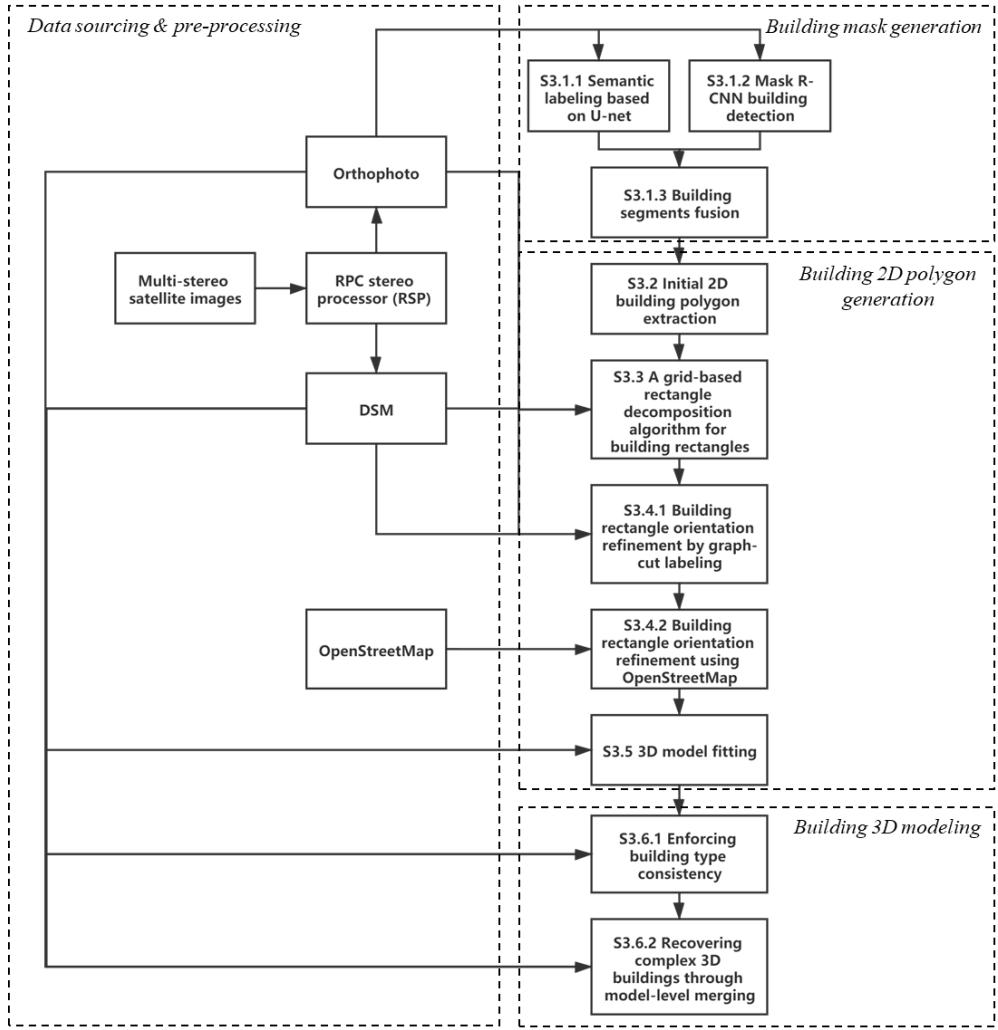

**Figure 2.2** Workflow of our proposed method, each key component is described in subsections in the texts.

### 2.4.1 Building detection and segmentation

The training and validation datasets were combined with satellite and aerial imagery. The satellite dataset was provided by John's Hopkins University Applied Physics Lab (JHUAPL) through the 2019 IEEE GRSS Data Fusion Contest (Le Saux et al. 2019). The aerial dataset was provided by INRIA aerial image labeling benchmark (Maggiori et al.

2017). These datasets have urban buildings from Jacksonville (USA), Omaha (USA), Austin (USA), Chicago (USA), Kitsap (USA), Tyrol (Austria), Vienna (Austria), therefore the trained network is able to work for most of the urban pattern during building segmentation. The images from training and validation datasets are clipped to several 512 × 512 patches with 50% overlap. Thus, there will be enough training data with 31615 patches, and 3512 patches for validation.

*2.4.1.1 Building segmentation by combining UNet and Mask R-CNN*

We used an approach that combines two baselines as introduced in this section: the U-Net and Mask R-CNN. U-Net (Ronneberger, Fischer, and Brox 2015) with its structure designated for preserving details provides well-delineated boundaries of objects, while Mask R-CNN (He et al. 2017) with its original structure although less complex to preserve good boundaries, it has the ability to perform instance level segmentation thus often to have better recall, while given the nature of the region-based detection for Mask R-CNN, it may omit certain large-sized buildings. These two are complementary to each other, therefore we perform a segmentation-level fusion for results generated by both networks. In the following sections, we introduce training details of U-Net, in which we used a revised loss (Qin et al. 2019), and the Mask R-CNN, both of which are trained using satellite datasets.

We used the training dataset provided by John's Hopkins University Applied Physics Lab's (JHUAPL) through the 2019 IEEE GRSS Data Fusion Contest (Le Saux et al. 2019). In total, five classes are considered in this data: ground, tree, roof, water, and bridge and elevated road. While in practice, those five objects are not balanced, so the number of training patches is adjusted to make the number of each category has similar patches (Qin et al. 2019; Liu and Qin 2020).

For the prediction part, the data splitting and fusion method are performed as well. The original testing images with a size of 1024 × 1024 are divided into 512 × 512 patches with 80% overlap. Moreover, four predictions are performed by rotating the patches to four directions for each patch. The final prediction can be derived by fusing these predictions with the following strategy: the size of each patch is 512*512, a buffer with a width of 64 pixels in the patch border will not be considered (weight = 0), otherwise pixels will have a weight equal to 1. In the merging process, a voting strategy is used to calculate the summed weight of each patch (predictions of four directions are rotated to the original direction), and class of pixels will be assigned as the class with the highest weight in that pixel. Thus, the segmentation result can be developed from orthophotos.

For Mask R-CNN, the only category we need is building, and this group is masked for training, validating, and testing. The vertexes of building masks in samples are extracted by using the building polygon extraction method, which is described in **Section 2.4.2** and **Section 2.4.3**. Furthermore, the bounding-boxes are generated from these vertexes as well. In the study, the pre-train weight focus on building segmentation provided by CrowdAI (Mohanty 2018) is used to Mask R-CNN processing. Furthermore, the dataset of 2019 IEEE GRSS Data Fusion Contest (Le Saux et al. 2019) is used for training and validation, to achieve a specific weight of network for building segments of orthophoto. During the training step, a total number of 120 epochs training has been performed, with 30 epochs for RPN training, 60 epochs for FCN training, and 30 epochs for all layers. Thus, each epoch contains 500 training steps and 50 validation steps.

As mentioned above, we used a segmentation-level fusion to obtain the initial building masks. The fusion method is rather heuristic, as U-Net tends to produce sharp object boundaries and Mask R-CNN tends to capture individual objects. Therefore, we start by using the U-Net result as the primary segmentation and use a decision weight $w$ to decide if the label of a segment from Mask R-CNN needs to be used for replacing the result of UNet, as shown below:

$$w = \frac{1}{area_{bbox}} \cdot \frac{area_{class}}{area_{bbox}} = \frac{area_{class}}{(area_{bbox})^2} \qquad (2.1)$$

where $w$ indicates the decision weight of a bounding box, $area_{class}$ is the area of the masked pixel cover in the bounding box, $area_{bbox}$ is the area of the bounding box. The rationale of this formulation follows the observation that the Mask R-CNN tends to perform well with smaller buildings (possibility due to the large number of small buildings in the training set), and as long as detection has a good filling within its bounding box, it may be used as a confident detection of individual buildings. The threshold of determination is empirical (here we variably used 0.1 to 0.2 to yield reasonable results for this 0.5 m resolution data), which is stable in one region. If the average size is large, the threshold will be close to 0.1. Otherwise, the threshold will be close to a larger value. **Figure 2.3**, shows the decision process. Firstly, use the semantic labeling result as primary classification. Secondly, calculate the segment weight of each segment from Mask R-CNN result, select segments with weight exceeding threshold as potential segments. Finally, use the potential segments in Mask R-CNN result to replace the corresponding mask of semantic labeling result.

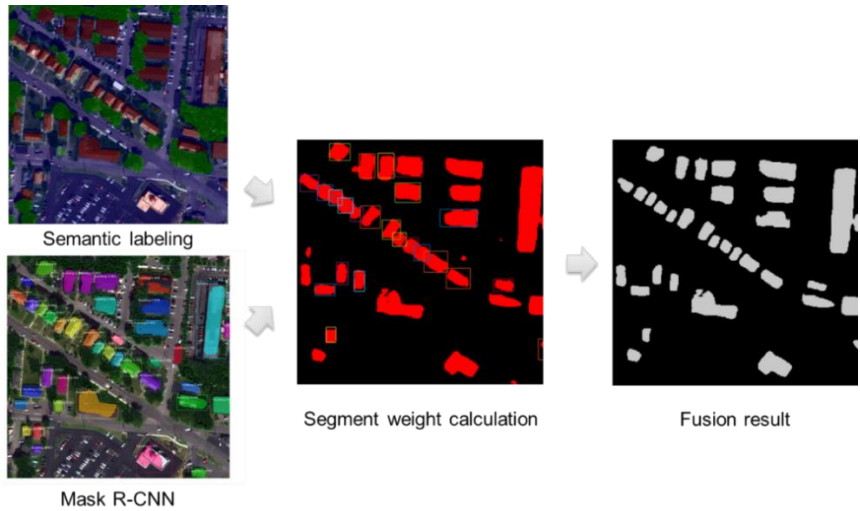

**Figure 2.3** A schematic workflow for segmentation-level fusion

*2.4.1.2 Building segmentation with HRNet*

Our approach has a trained network that harnesses the existing available benchmark datasets. We used an approach of HRNetV2 (K. Sun et al. 2019) to get building segments by using orthophoto with RGB bands. The training process is a binary classes segmentation in which there are only building and non-building labels for training and validation. During the training step, a total of 30 epochs have been performed with 5000 iterations per epoch, and we adapted the weight from the last epoch for our segmentation function. The training environment is on a GPU of Nvidia RTX 2070s, and the batch size is two during training.

In the prediction part of the software, input orthophoto is divided into 512 × 512 patches and then predicted those patches. Moreover, the final segmentation is developed by merging the segment of the predicted individual patch. Thus, the image size processing capability of software depends on the RAM rather than the GPU memory of the user's PC. The output file of this step will be a building segmentation image, which can be used for building detection or building area computation.

**2.4.2 Initial 2D building polygon extraction**

Building polygon extraction aims to vectorize building boundaries as polylines to assist regularized building model fitting. These polylines need to follow constraints as man-made architectures do, such as complying for orthogonality and parallelism. For coarse resolution data, this is simplified as polylines consisting of parallel or orthogonal lines, the extraction of which in this work follows three steps: initial line extraction, line adjustment, and regularization; the result of each step by example is shown in **Figure 2.4**.

The *initial line extraction* performed by using the well-known Douglas–Peucker algorithm (Douglas and Peucker 1973). It starts at the boundary points of the building mask and extracts lines recursively. Due to the irregularity of the building masks, the initial line segments present errors, sharp angles and are short. The *line adjustment* process aims to use the concept of main orientations to connect and turn these short line segments to consistently long and straight lines: One or more main orientations are defined for the polygon to separate fitted lines into bins presenting orthogonal line pairs. First, we draw a histogram of orientations varying from 0° to 90° with 10° of interval, and each bin present two angles that are 90° different: for example, for the bin present 30°, it would include line segments close to -60° as well. This is to facilitate maximizing the line adjustment to contain mainly orthogonal and parallel lines as for regularity. Then we sum up the length of lines for each bin and select the bins whose summed length is bigger than a given threshold (here we set 120 pixels), in which we conclude the main orientation by taking the weighted average (weight being the length of each segment) of the orientations in that bin. The orientation for bin is computed as the weighted average of lines' orientation in this group (Note: orthogonal orientations are deemed identify thus only represented by angles within 0°-90°). The adjustment is done by iterating through each line segment, reassigning to the closest main orientation, intersect and reconnect with neighboring segments; clear differences can be noted between in **Figure 2.4 (a)** and **Figure 2.4 (b)**.

The *line regularization* process further optimizes line orientation by utilizing directly detected lines from the orthophoto: Line-Segment Detector (LSD) algorithm (Von Gioi et al. 2008) is used to extract line segments and the orientation of each line segments is used to readjust orientations of nearest line of the initially extracted line segments from the building mask to enable consistencies between the detected footprint and the image edges. The resulting building footprint after this process is regarded as the initial building polygon for reference.

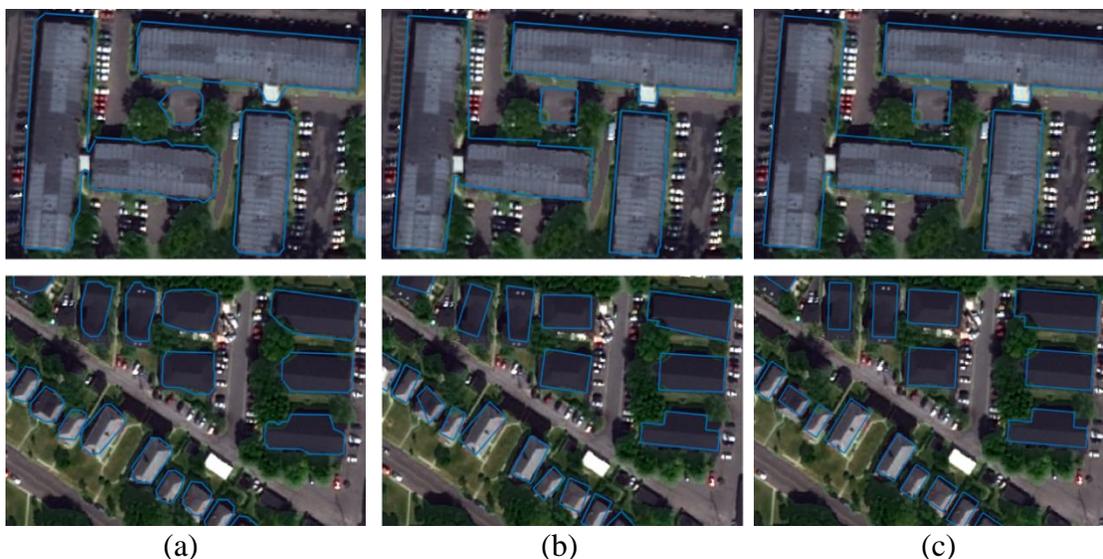

(a)                  (b)                  (c)

**Figure 2.4** Two examples showing boundary extraction of building polygon. (a) boundary after initial line extraction; (b) boundary after line adjustment; (c) boundary after line regularization.

### 2.4.3 A grid-based rectangle decomposition algorithm for building rectangles

The polyline based initial building polygon can be overcomplex for the limited number of simplex models to fit, thus it requires to decompose these building polygons (as produced following method introduced in **Section 2.4.2**), into simple shapes, which in this work as basic rectangles on which the model fitting can base on (we call it building rectangle hereafter). Existing methods mostly follows a greedy strategy for decomposition, which identifies the main orientation following the longest straight line, extends parallel lines gradually to get decomposed rectangles to fill the polygon (Kada 2007; Partovi et al. 2019). However, often the results of such a greedy-decomposition method does not match the actual building components. We therefore present our grid-based decomposition approach, in which information from the orthophoto and DSMs can be flexibly integrated to guide this process.

The grid-based decomposition approach assumes that multicomplex building polygons generally follows Manhattan type and are composed of simplex rectangles, the seamlines of which can be approximately either horizontal or vertical if line on a 2D grid. The workflow is shown in **Figure 2.5**. To start, we first rotate the 2D building polygon by aligning its main orientation to the x axis (**Figure 2.5**), and the same transformation will be performed on the Orthophoto and DSM. The decomposition of the polygon to rectangles starts with a DSM and orthophoto based initial decomposition: first, gradients of the DSM in both the horizontal and vertical directions are computed and the those larger than a threshold (0.3 m), followed by a non-maximum suppression (with a window size of 7 pixels) are selected as the candidate of lines for separating the polygon, which are further filtered by considering the color information: for each candidate separating line, we create a small buffer on both side and only lines with color differences of between these two buffers bigger than a threshold $T_d$ (we used 10 for an 8-bit image). For the segments after this initial decomposition, a maximum inner rectangle extraction (Alt, Hsu, and Snoeyink 1995) is performed to extract individual rectangles. Note this above process is performed on the coarsest layer an image pyramid to reduce impacts of the noise, and the results are then interpolated to the finer layers for further analysis. For our 0.5m resolution data, we kept three layers (grid level 1-3, **Figure 2.5**), meaning that the top layer is ¼ of the original resolution.

The individual rectangles will then be projected to the original resolution grid and post-processed by using a merging operation, as the above process might over separate the building masks. Here the merging operation will only be performed on rectangles that share one edge, as the merged the rectangles must be rectangles as well to enable fitting. To start, we first identify adjacent rectangles by dilating their boundaries for 7 pixels and define adjacency as long as there are overlap and common edge has similar length (difference shorter than 5 pixels in length). The criterion used to decide if two adjacent rectangles should be merged are in below:

$$\begin{cases} merge, & |\overline{C_1} - \overline{C_2}| < T_d \ \cap \ |\overline{H_1} - \overline{H_2}| < T_{h1} \cap max|\Delta H_{edge}| < T_{h2} \\ not\ merge, & otherwise \end{cases} \quad (2.2)$$

meaning that the mean color differences ($|\overline{C_1} - \overline{C_2}|$) of the two rectangles (projected onto the orthophoto) is smaller than a threshold $T_d$ (here we define as 10 for an 8-bit image); 2) mean height difference ($|\overline{H_1} - \overline{H_2}|$) is smaller than a threshold $T_{h1}$ (1 m); 3) There is no dramatic height changes in a buffered region that cover the common edge, to avoid narrow streets in between, and this criterion is defined as the height gradient ($\Delta H_{edge}$) in the overlapped region (after the dilation) smaller than $T_{h2}$ (0.2 m).

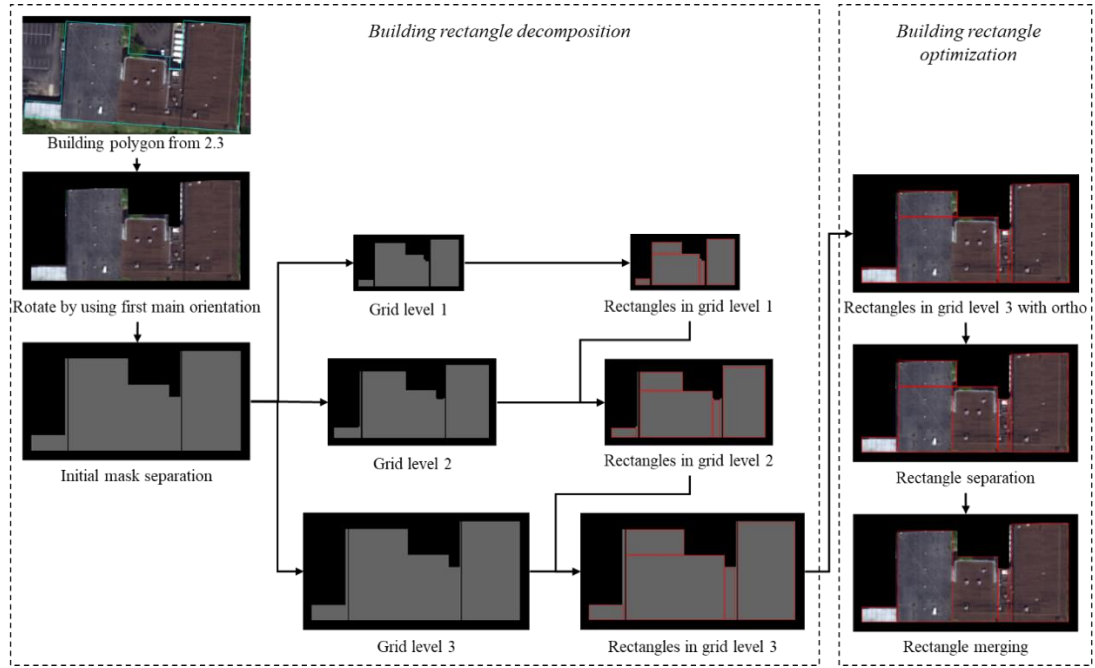

**Figure 2.5** The proposed grid-based decomposition algorithm. Details are in the texts.

**2.4.4 Orientation refinement for building rectangles**

The rectangular building elements produced based on the process as described above are independent of each other, and on the other hand, the orientation of these rectangle footprints can be easily impacted by noises of the initial building mask and the orthophotos & DSM. Considering that most neighboring buildings follow consistent orientations, such as those in the same street block. Therefore, we optimize the orientations for buildings using the Graph-Cut algorithm (Boykov and Jolly 2001), and as an optional step to refine the orientation using available OpenStreetMap road networks (covering 80% of the cities worldwide (Barrington-Leigh and Millard-Ball 2017).

*2.4.4.1 Building rectangle orientation refinement by graph-cut labeling*

We formulate the orientation adjustment problem as a multi-labeling problem, by assigning each building rectangle, possible labels as discrete angle values, ranging from 0° to 180°, with a 2° as the interval, resulting in approximately 90 labels (0° pointing to the north). We aim to assign these labels such that similar neighboring building rectangles have the same or similar labels, where similarity is defined by using texture and height differences. GC is well known for solving multi-label problems in polynomial time (Kolmogorov and Zabih 2002). It aims to minimize a cost function consisting of a data term and a smoothness term, in which the smoothness terms aims to enforce consistency between nodes and the data terms encode a priori information, shown in

$$E(\mathcal{L}) = \sum_{x_i \in P} R(x_i, \mathcal{L}_i) + \lambda \sum_{(x_i, x_j) \in \mathbb{N}} B(x_i, x_j) \delta(\mathcal{L}_i, \mathcal{L}_j) \qquad (2.3)$$

$$R(x_i, \mathcal{L}_i) = 1 - e^{-|\theta_{x_i} - \theta_{\mathcal{L}_i}|} \qquad (2.4)$$

$$B(x_i, x_j) = W(i, j) \qquad (2.5)$$

$$\delta(\mathcal{L}_i, \mathcal{L}_j) = \begin{cases} 0, & if\ \mathcal{L}_i = \mathcal{L}_j \\ 1, & if\ \mathcal{L}_i \neq \mathcal{L}_j \end{cases} \qquad (2.6)$$

where $\mathcal{L} = \{0,1,2,\ldots,89\}$ is the label space indicating the orientation of the rectangle being $\theta_\mathcal{L} = 2 \times \mathcal{L}$ degrees, and $\mathcal{L}_i$ refers to the optimized label (or $\theta_{\mathcal{L}_i}$ as the optimized orientation) for the building rectangle $x_i$. $\theta_{x_i}$ refers to the initial orientation of the building rectangle and the data term $R(x_i, \mathcal{L}_i)$ tends to keep the initial orientation unchanged and is defined as a normalized value (smaller than 1) that gets bigger as the optimized orientation differs more. $B(x_i, x_j)$ is the smooth term, which defines the similarity of neighboring building rectangles (set ℕ), and $\lambda$ is the weight that leverages the contributions of the smoothness term. The smoothness term $B(x_i, x_j)$ is determined by using affinity matrix $W(i,j)$, and which adapts an exponential kernel (Liu & Zhang, 2004):

$$W(i,j) = e^{-dist(i,j)/2\sigma^2} \qquad (2.7)$$

$$dist(i,j) = \sqrt{\begin{array}{c} w_r(\boldsymbol{nr_i} - \boldsymbol{nr_j})^2 + w_\theta(n\theta_i - n\theta_j)^2 + w_S(\boldsymbol{nS_i} - \boldsymbol{nS_j})^2 \\ + w_C(\boldsymbol{nC_i} - \boldsymbol{nC_j})^2 + w_\sigma(\boldsymbol{n\sigma_i} - \boldsymbol{n\sigma_j})^2 \end{array}} \qquad (2.8)$$

where $W(i,j)$ is the affinity weight matrix for neighboring building rectangle $i$ and $j$, with $0 < W(i,j) < 1$, $\sigma$ is the bandwidth of exponential kernel. $dist(i,j)$ is the similarity between building rectangle $x_i$ and $x_j$, calculated as a weighted (and normalized) combination of a few factors including the distance of two rectangle center $\boldsymbol{nr_i}$, the difference of the orientation, the shape $\boldsymbol{nS_i}$, the color $\boldsymbol{nS_i}$, and that of the color variations $\boldsymbol{n\sigma_i}$, respectively defined as $\boldsymbol{nr_i} = (nX_i, nY_i)$, where $nX_i$ and $nY_i$ referring to the location of the rectangle center, $\boldsymbol{nS_i} = (nL_{1,i}, nL_{2,i})$ referring to the length $nL_{1,i}$ and width $nL_{2,i}$ of the rectangle, $\boldsymbol{nC_i} = (nC_{R,i}, nC_{G,i}, nC_{B,i})$ referring the mean value of RGB that rectangle covers, $\boldsymbol{n\sigma_i} = (n\sigma_{R,i}, n\sigma_{G,i}, n\sigma_{B,i})$ referring to the standard deviation of RGB. We give a higher weight value $w_r = 3$ to location, to ensure consistencies are mainly optimized locally; the weight of orientation $w_\theta$ and the weight of shape $w_S$ are set by default as 1, and the weight of RGB color $w_C$ and the weight of RGB standard deviation $w_\sigma$ are set as 0.3, since both the $\boldsymbol{nC}$ and $\boldsymbol{n\sigma}$ have three components. Note all measures for computing similarities are normalized through an exponential kernel similar to **Equation (2.7)**. These weights (for location, orientation & shape, color-consistency to decide similarity of two neighboring buildings), are set to be intuitive and often can be fixed in the optimization based on the quality of the data. It is possible to automatically tune these parameters based on prior information of the dataset, or to use the meta-learning method, or to decide the parameter based on the distribution of the data (while only slight difference from our empirical values (Kramer 1987). Here since these parameters are intuitive and explainable, these weights are not proceeding an automated tuning.

*2.4.4.2 Building rectangle orientation refinement using OpenStreetMap*

OpenStreetMap is a publicly available vector database and it was known to have covered more than 80% (Barrington-Leigh and Millard-Ball 2017) of the road networks globally, thus it may serve as a valid source to refine building orientations for well geo-referenced dataset. We made a simple assumption that the buildings always have the same direction as their surrounding road vectors (Zhuo et al. 2018). Therefore, we perform the OSM based orientation refinement following simple heuristics or reassigning orientation using nearest criterion: for each building rectangle, the nearest road vector is found by computing the distance between the center of the building rectangle and each road vector, and if the intersection angle between the main orientation line of a building polygon (introduced in **Section 2.4.2**) and the nearest road vector is smaller than 30°, the orientation of the rectangle will be adjusted, otherwise kept unchanged. This process will allow minor adjustment of the orientation when the OSM data is available. Two examples are shown in **Figure 2.6** showing comparisons when the GC optimization and the OSM based adjustment are respectively applied, resulting in adjusted rectangles (in blue and green).

Once OSM is unavailable in certain region or orientation difference between building and OSM street line, orientation refinement using OSM will not execute and there will be sole refinement by using GC.

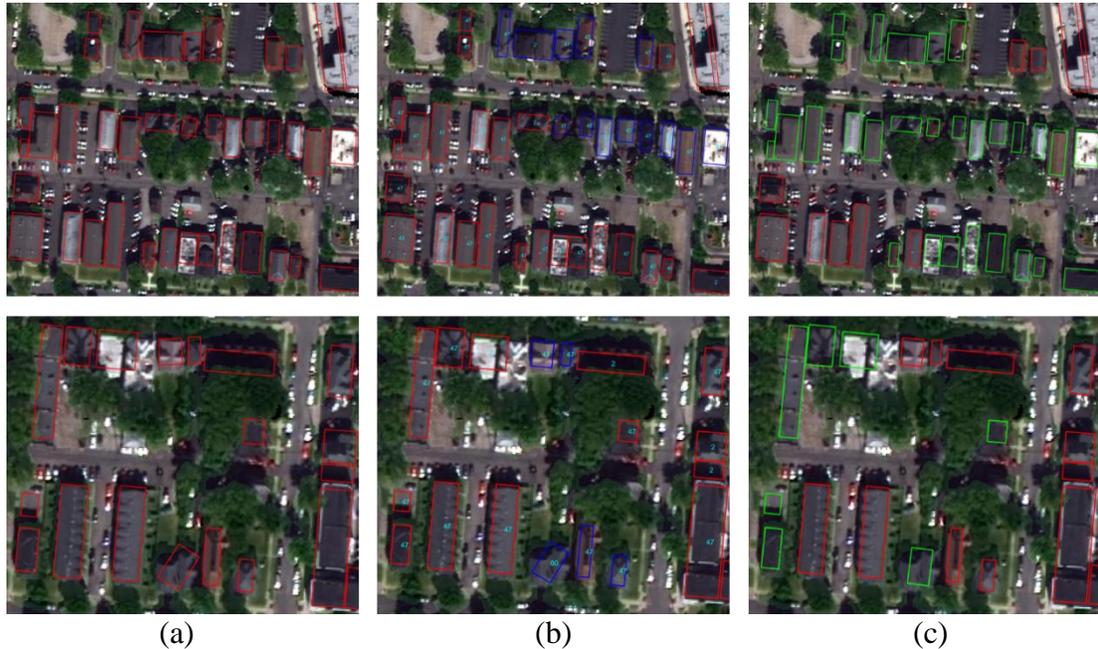

(a) (b) (c)

**Figure 2.6** Two examples showing building rectangle orientation respectively adjusted by GC (b) and OSM (c). Blue and Green rectangles refer to those changed during the decomposition. (a) shows polygons before the orientation refinement.

**2.4.5 Rectangular-based 3D model fitting**

The extracted building polygons are in rectangle shapes; together with DSM/nDSM, these can be easily fitted by simplex building models. Here we only consider five types of building models namely: flat, gable, hip, pyramid, and mansard, a brief illustration of which is shown in **Figure 2.7**. It can be noted the complexity of the roof topology increases from left to right, along with more parameters to be considered for fitting.

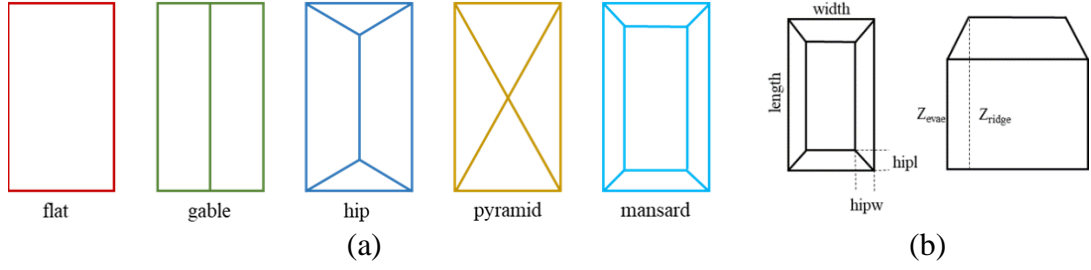

**Figure 2.7** (a) Model roof library with five types and geometrical parameters, color-coded for subsequent demonstration of results. (b). Tunable parameters to define these model types.

Each building model can be described by using several geometrical parameters. we adopt the parameterization based on (Partovi et al. 2019), where the geometrical parameters $\psi$ are defined as follows:

$$\psi \in \Psi; \ \Psi = \{P, C, S\} \tag{2.9}$$

where $\Psi$ includes three subsections, $P, C, S$. With $P = \{x_0, y_0, Orientation\}$ are the location parameters of the building model, and $C = \{length, width\}$ are the contour parameters of the building model, and $S = \{Z_{ridge}, Z_{eave}, hipl, hipw\}$ are the shape parameters of the building model, and the geometrically meaning of these parameters are shown in **Figure 2.7 (b)**. Each type of building model may be special cases for this type (shown in **Figure 2.7**): for example: for flat building $Z_{ridge} = Z_{eave}$, and for hip building, $hipw = width/2$ etc. Details of different models under this rationale can be found in **Table 2.1,** where "width", "length" come from the fitted polygons, and "$\overline{Height}$" calculated as the mean elevation (from DSM) based on the building polygon (i.e. rectangle).

**Table 2.1** Initial parameters of building model, parameters with * means these are constants

| Model type | $hipl_{(0)}$ | $hipw_{(0)}$ | $Z_{eave(0)}$ | $Z_{ridge(0)}$ |
|---|---|---|---|---|
| Flat building | 0* | 0* | $\overline{\text{Height}} - 0.5$ m | $\overline{\text{Height}} - 0.5$ m |
| Gable building | 0* | 1/2 width* | $\overline{\text{Height}} - 0.5$ m | $\overline{\text{Height}}$ |
| Hip building | 1/4 length | 1/2 width* | $\overline{\text{Height}} - 0.5$ m | $\overline{\text{Height}}$ |
| Pyramid building | 1/2 length* | 1/2 width* | $\overline{\text{Height}} - 0.5$ m | $\overline{\text{Height}}$ |
| Mansard building | 1/4 length | 1/4 width | $\overline{\text{Height}} - 0.5$ m | $\overline{\text{Height}}$ |

The optimization specifically updates the parameter set $\boldsymbol{S} = \{Z_{ridge}, Z_{eave}, hipl, hipw\}$ based on the DSM, where the starting terrain height is computed by taking the local minimum of building height. Given the noisiness of the DSM, we consider directly perform an exhaustive search over the parameters set and choose the model and parameter set with the smallest RMSE. As can be seen from **Table 2.1**, the enabled parameters for exhaustive search increase as the model gets more complex. We observe that since one building only covers a few hundreds of pixels, even with exhaustive search the computational time is still reasonable to scale up. Ranges of parameters for searching are shown in **Table 2.2**, the value of "searching step size" decided by the resolution of satellite-derived DSM (0.5-meter ground sampling distance (GSD)).

**Table 2.2** Optimization parameters with their range and step

| Parameter | Parameter range | Searching step size |
|---|---|---|
| $Z_{eave}$ | $(Z_{eave(0)} - 3, Z_{eave(0)} + 3)$ | 0.2 |
| $Z_{ridge}$ | $(Z_{eave} + 0.5, Z_{eave} + 4)$ | 0.2 |
| Hipl | $(hipl_{(0)} - 1/8 \text{ length}, hipl_{(0)} + 1/8 \text{ length})$ | 0.4 |
| Hipw | $(hipw_{(0)} - 1/8 \text{ width}, hipw_{(0)} + 1/8 \text{ width})$ | 0.4 |

### 2.4.6 Model post-refinement

*2.4.6.1 Enforcing building type consistency*

Since the type of models are individually fitted and the nosiness of the DSM may result in inconsistent types of building for neighboring and similar building objects (an example shown in **Figure 2.8 (a)**). Here we use the same idea to enforce the consistencies of the building types through GC optimization (details in **Section 2.4.4.2**), in which label space contains five labels representing the used building types, and the smooth term $B(x_i, x_j)$ keeps the same (**Equation (2.5)~(2.8)**) as to enforce label consistency for neighboring buildings with similar color and height, and the data term $R(x_i, \mathcal{L}_i^{'})$ for building type follows a binary representation being a constant if the target label equals to the initial label, otherwise a large number:

$$E(\mathcal{L}^{'}) = \sum_{x_i \in P} R(x_i, \mathcal{L}_i^{'}) + \lambda \sum_{(x_i, x_j) \in \mathbb{N}} B(x_i, x_j) \delta(\mathcal{L}_i^{'}, \mathcal{L}_j^{'}) \qquad (2.10)$$

$$R(x_i, \mathcal{L}_i^{'}) = 1 - e^{-D(x_i, \mathcal{L}_i^{'})} \qquad (2.11)$$

$$D(x_i, \mathcal{L}_i^{'}) = \begin{cases} 0, & Type_{x_i} = Type_{\mathcal{L}_i^{'}} \\ 1, & Type_{x_i} \neq Type_{\mathcal{L}_i^{'}} \end{cases} \qquad (2.12)$$

where $\mathcal{L}^{'} = \{flat, gable, hip, pyramid, mansard\}$ is the label space indicating the building type of the rectangle for type refinement, and $\mathcal{L}_i^{'}$ refer to the optimized label (same as $Type_{\mathcal{L}_i^{'}}$) for building rectangle $x_i$. $Type_{x_i}$ refers to the initial building type of building rectangle and the data term $R(x_i, \mathcal{L}_i^{'})$ tends to remain the same type as initial, the value of which is normalized using an exponential kernel. Data term $R(x_i, \mathcal{L}_i^{'})$ is determined by a binary function $D(x_i, \mathcal{L}_i^{'})$, which represents the building type difference between building rectangle $x_i$ and potential label $\mathcal{L}_i^{'}$, and it is set as zero if the building types are the same as the initial type.

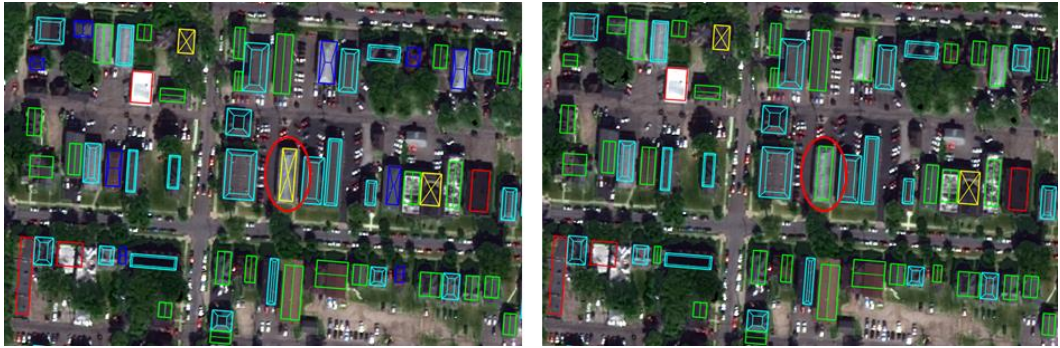

**Figure 2.8** Fitted building types before (left) and after (right) the label consistency enforcement through GC. Different building types are color-coded differently. An example of such a correction is shown in red circle.

It can be seen from **Figure 2.8** that the type of enforcement through GC has yield more consistent building types for neighboring buildings. It should be noted this process can be again optional depending on the quality of the DSM.

*2.4.6.2 Recovering complex 3D buildings through model-level merging*

Large buildings with complex roof structures after decomposition, need to be recovered to 3D topological models. Once individual 2D building rectangles are fitted with one of the five types of models, we consider a model level merging operation to recover potential 3D buildings with complex shapes through line intersection and rule-based merging process, as described in below:

To start, we first identify adjacent building rectangles and determine whether or not to perform the merging using criteria as used in **Section 2.4.3** (**Equation (2.2)**). The model-level merging is performed at the 2D level and re-fit the model following algorithm from (Brédif et al. 2013) (an example of the process is shown in **Figure 2.9**): with two building rectangles (**Figure 2.9** (a)), it first extends the side of the rectangle and seeks for intersections of these sides (**Figure 2.9** (b)), followed by the extraction of the enclosed polygon (**Figure 2.9** (c)) as the base of the more complex shapes. Here we assume the type of the merged polygon may change depending on the original model type of the basic building model. The determination of the merged model types follows heuristics of the building topology, denoted as type conversion matrix shown in **Table 2.3**. For example, Pyramid roofs often represent individual building and when two of them are identified for merging, it will skip; when one of the two basic building models has the type of pyramid, the type of the resulting merged model will be identified through the converting matrix, as if the pyramid roof has the same type of the other basic building model. The optimization of the merged roof parameters ($Z_{ridge}, Z_{eave}, hipl, hipw$) follows the method described in **Section 2.4.5** (initial parameters follow the basic model with larger footprint), resulting in the final merged model (**Figure 2.9** (d)).

Table 2.3 Roof type decision matrix during model merging

| Roof type | flat | gable | hip | pyramid | mansard |
|---|---|---|---|---|---|
| Flat | flat | gable | hip | flat | mansard |
| Gable | gable | gable | hip | gable | mansard |
| Hip | hip | hip | hip | hip | mansard |
| Pyramid | flat | gable | hip | Not merge | mansard |
| Mansard | mansard | mansard | mansard | mansard | mansard |

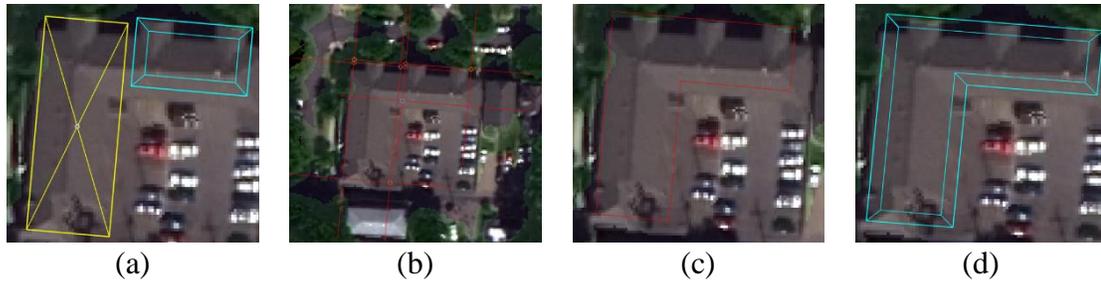

(a) (b) (c) (d)

**Figure 2.9** An example of model-level merging from two basic building models (a); (b) intermediate step for 2D-level merging by extending and intersecting sides of rectangles; (c) merged 2D-level polygon; (d) merged model.

*2.4.6.3 Model of irregular building shapes*

With building models parameters extracted from **Section 2.4.3 & 2.4.4** in the 2D level from **Section 2.4.5** in 3D level, a 3D building mesh can be generated by using these parameters. The processing of converting buildings from DSM to vectorized mesh decreases the redundancy of data, and clean and straightforward geometry with tiny size. However, some buildings may have irregular shapes; in other words, multi-complex models may cause obviously incorrect detections. Therefore, a decision strategy using the Intersection over Union (IoU) based on decomposed rectangles and building masks from segmentation is adapted to discover an irregular shape building model: If IoU between building mask and rectangles in a segment is lower than 0.65, and the building mask has an area larger than 5000 pixels, DSM covered by building polygon in **Section 2.4.2** is transferred into mesh directly, and then mesh is simplified less than 1000 faces. These numbers are tuned specifically for 0.5-meter resolution data. This step ensures that irregular buildings will not be vectorized into an incorrect shape.

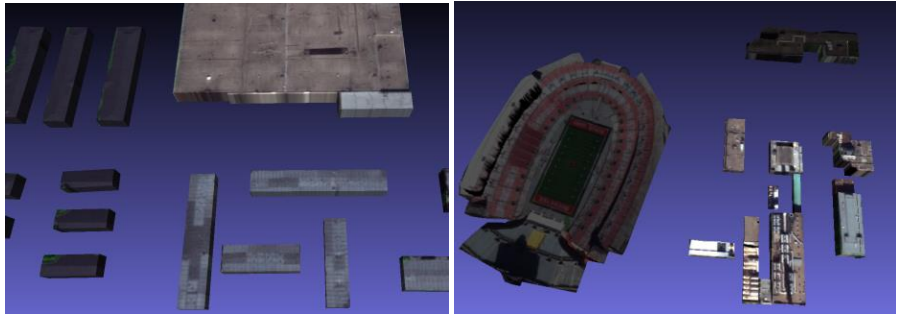

**Figure 2.10** Left: example of the regular (rectangular) building model, right: example of the irregular (non-rectangular) building model

## 2.5 Experiment

### 2.5.1 Study area

The experiments are performed in three cities with different geographical location and urban patterns, one located in a typical U.S. city (City of Columbus, Ohio), the other one in South America (Buenos Aires, Argentina), and the last one located in a European Megacity (London, UK). We selected two representative regions for each city (totaling six regions) as shown in **Figure 2.11** (first row, overlaid with ground-truth labels): The Columbus-area-1 shows a typical residential/commercial region with relatively dense buildings in similar sizes and Columbus-area-2 a typical industrial region containing buildings with varying sizes; Buenos-Aires-area-1 presents a very challenging scene that contains both large-sized building and small & dense buildings; Buenos-Aires-area-2 presents a region with relatively sparse and single houses, with disturbances such as swimming pools and regular playgrounds; The London-area-1 shows a typical block region with compact buildings; London-area-2 shows a region with numerus complex buildings.

Each of these regions covers approximately 0.5 to 2.25 km² area with Orthophoto and DSM at a 0.5 m GSD (ground sampling distance). The orthophoto and DSMs are produced by using a multi-view stereo approach (Qin 2016; Qin 2019; Qin 2017) based on five worldview-3 stereo pairs for Buenos Aires dataset and 12 World-view stereo pairs for the Columbus dataset. The accuracy of the DSMs were analyzed systematically in the work of (Qin 2019), which reported to have achieved sub-meter vertical accuracy in terms of RMSE (root-mean-squared error). The orthophotos are pan-sharpened using an 8-band multispectral image, while for simplicity, converted to 3-band RGB for building detection and processing. The evaluation of the geometry and detection are evaluated separately using both a 2D Intersection over Union (IOU2) and 3D Intersection over Union (IOU3) based on manually created reference data for building footprint and LiDAR-based DSM for 3D geometry (as shown in **Figure 2.11**). The IOU2 and IOU3 are defined following (Kunwar et al. 2020) as follows:

$$IOU2 = \frac{TP}{TP + FP + FN} \quad (2.13)$$

$$IOU3 = \frac{TP_{3D}}{TP_{3D} + FP + FN} \quad (2.14)$$

where $TP$ is the number of true positives pixels that determined as extracted and manually labeled building footprint simultaneously, $FP$ the number of false positives and $FN$ the number of false negatives. $TP_{3D}$ is $TP$ pixels whose 3D vertical difference from the ground-truth LiDAR is within 2 m.

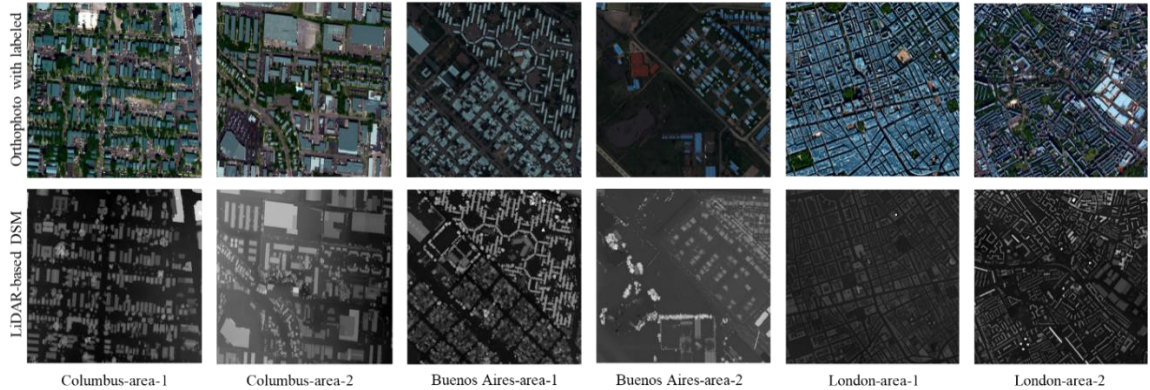

**Figure 2.11** First row: Orthophotos of the study areas, overlaid with manually drawn masks; second row: corresponding ground-truth LiDAR data.

**2.5.2 Experimental results**

The orthophotos study area are shown in **Figure 2.11**, first row, overlaid with manually drawn building masks results of the building 2D polygon and rectangle extraction (methods described in **Section 2.4.1-2.4.4**) and building 3D model fitting (methods described in **Section 2.4.5-2.4.6**) are shown in **Figure 2.12,** which specifically includes building mask detection, initial building polygon detection, decomposition & merging, and orientation refinement with GC and OSM. We show the model fitting results of the six experimental regions in the first row of **Figure 2.12,** by projecting the wireframes of the model on the orthophoto; in the second row of **Figure 2.12,** we demonstrate intermediate results of the first region (i.e. Columbus-area-1) including initial building polygon extraction, building rectangle decomposition and refinement, as well as the final fitted more, with the third row of **Figure 2.12** highlighting some of the results. It can be seen that the proposed method has detected most of the buildings and correctly outlined the building boundaries. In addition, most of the buildings initially detected to be connected are successfully decomposed as individual rectangles. Minor errors are observed for small and complex buildings where decompositions may fail: an example is shown in the third row (circled in red), which has erroneously separated buildings into a thin rectangle thus being reconstructed incorrectly.

**Figure 2.12** also shows the final result of building modeling for the six regions with roof type, and classes other than building are set as terrain DSM. **Figure 2.13** display the LoD-2 reconstructed buildings in six experimental areas. The result indicates that the proposed approach performs well on community building in Columbus-area-1, Columbus-area-2, and Buenos-Aires-area-2, while in very complex block, like Buenos-Aires-area-1 and London-area-1, some buildings are not reconstructed so accurate, since building mask from **Section 2.4.1** not detect perfect segments.

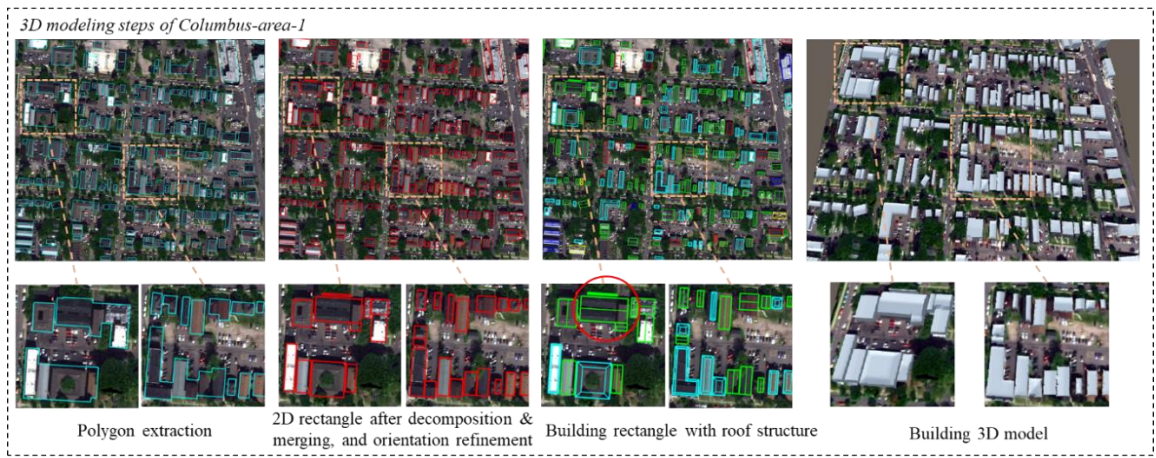

**Figure 2.12** LoD-2 model reconstruction steps for Columbus-area-1. First row: Intermediate results of the "Columbus-area-1" region including initial building polygon extraction, polygon decomposition and refinement, model fitting and the last figure of this row shows the 3D visualization. The second row enlarges part of each figure of the second row for visualization. The building circled in red shows an example of erroneous reconstruction.

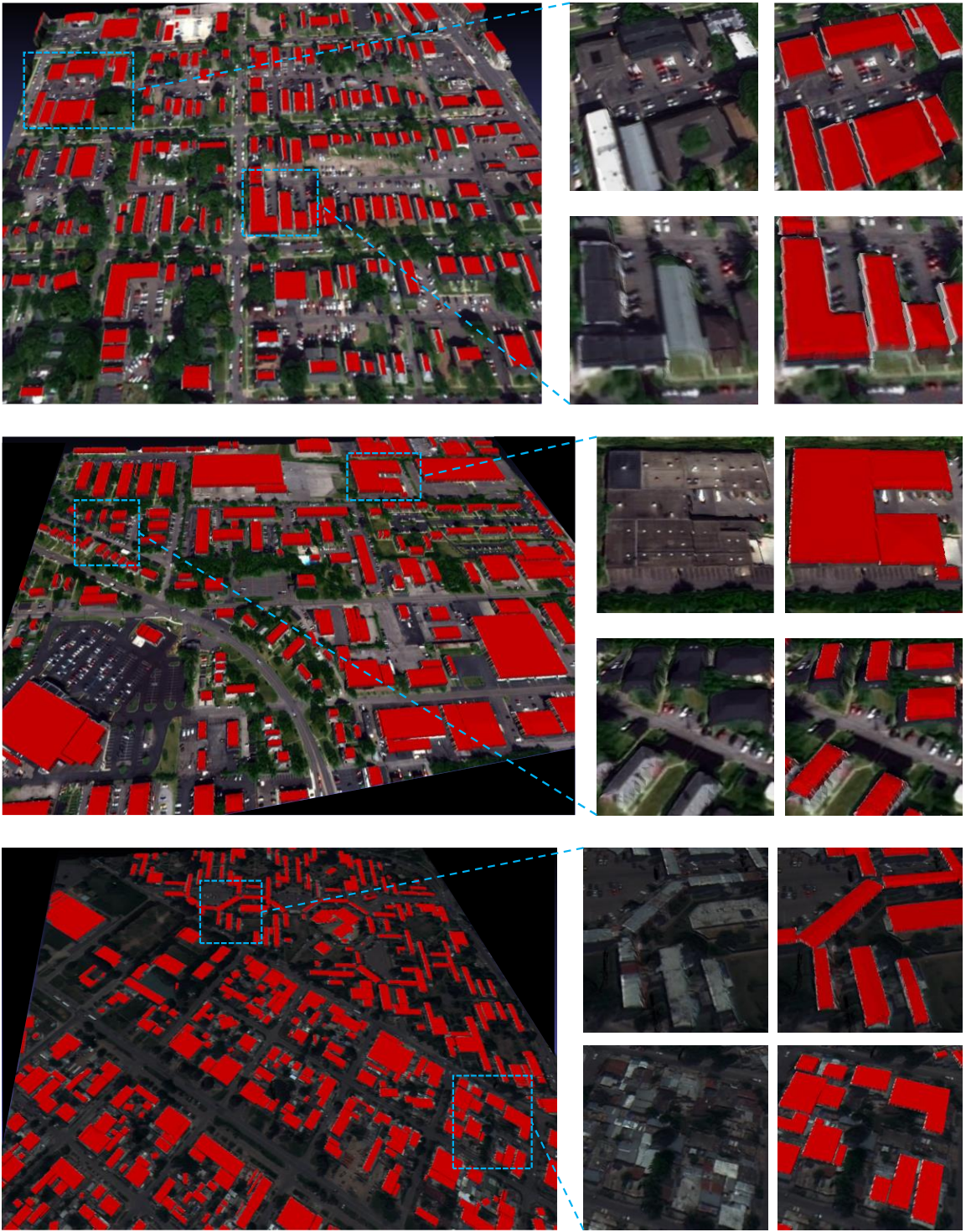

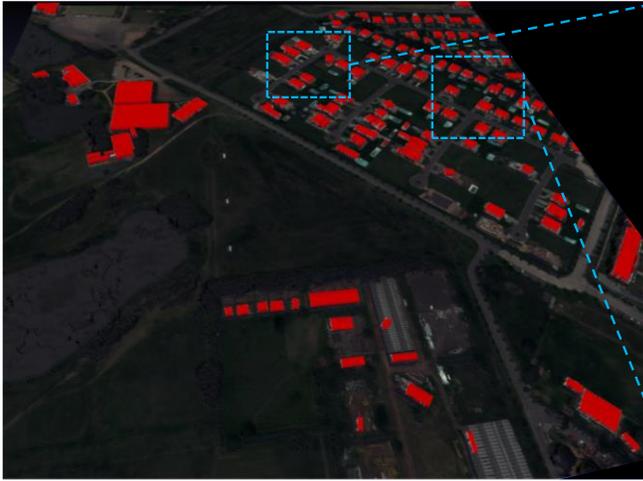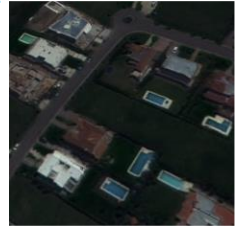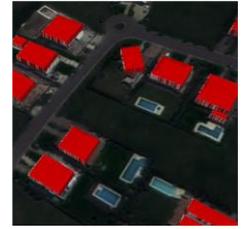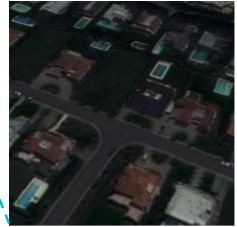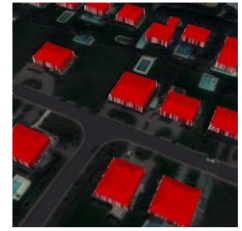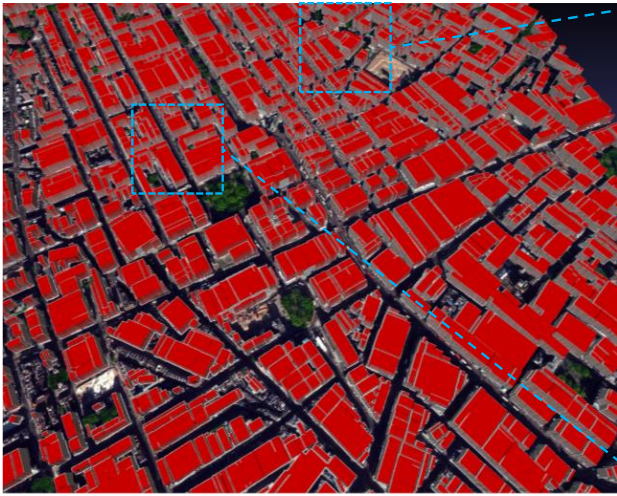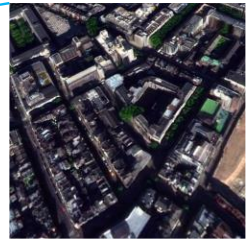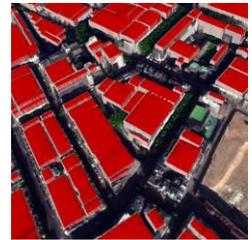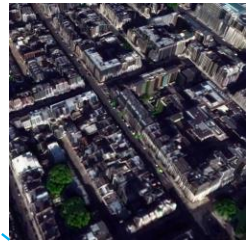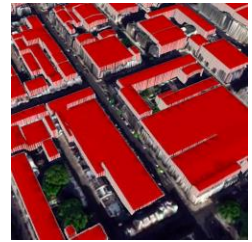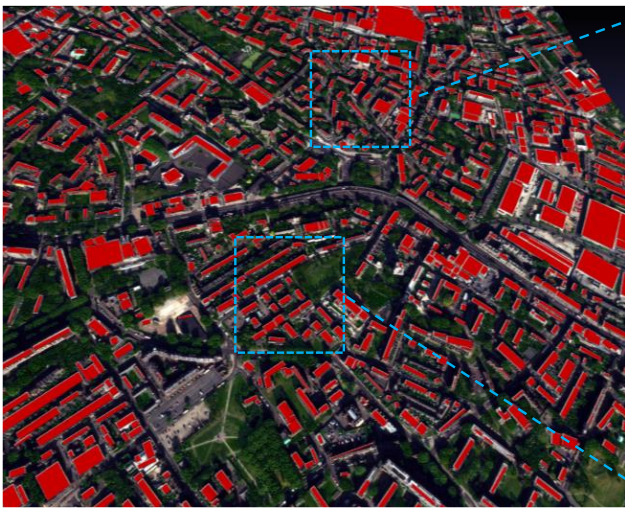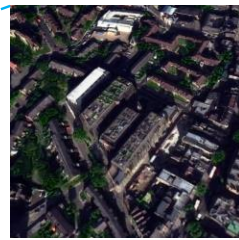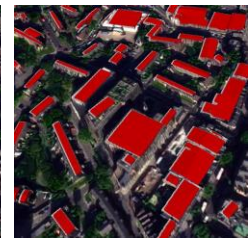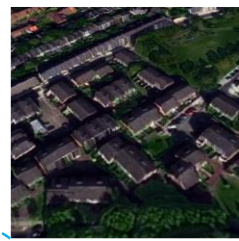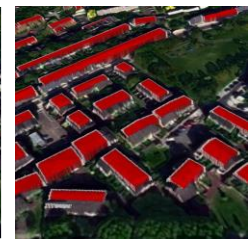

**Figure 2.13** LoD-2 model reconstruction results of the six experimental regions. Left column: 3D model with roof covered by red mask and façade covered by white; medium column: DSM with texture in a close view, right column: 3D model with roof covered by red mask in a close view. From top to bottom: Columbus-area-1, Columbus-area-2, Buenos-Aires-area-1, Buenos-Aires-area-2, London-area-1, and London-area-2.

**2.5.3 Accuracy evaluation of experiment areas**

The accuracy of the resulting models is evaluated using the IOU2 and IOU3 metric, respectively to assess the 2D building footprint extraction accuracy and the 3D fitting accuracy. The accuracy about 'DSM' represents the raw measurements to be compared with the ground-truth. We have ablated the results with and without the GC and OSM refinements, statistics against the ground truth are shown in **Table 2.4**, and visualized sample comparison in **Figure 2.14**, where "OSM" or "GC" indicates that the orientation is refined using OSM or GC alone, and "OSM+GC" indicates the orientation being refined by OSM first and followed by GC. It can be seen that in both cases OSM and GC refinement had a positive impact on the resulting metrics in three out of the six areas, and it is possible that incorrect OSM might adversely impact the final IOU. It should be noted that although the statistics show only a marginal improvement of the GC and OSM refinement (largely due to that these adjustments are small and the building segments are rather accurate), it visually shows a much better consistency in occasions where the buildings are mis-oriented (e.g. **Figure 2.7**). The fitted DSM has a larger error distribution than the original DSM, since often fitting will result in reduced accuracy given that the regularized shape may introduce errors, as for an example, a raw measurement curved surface may be approximated by multiple piece of planar surface, and the same applies to the model fittings.

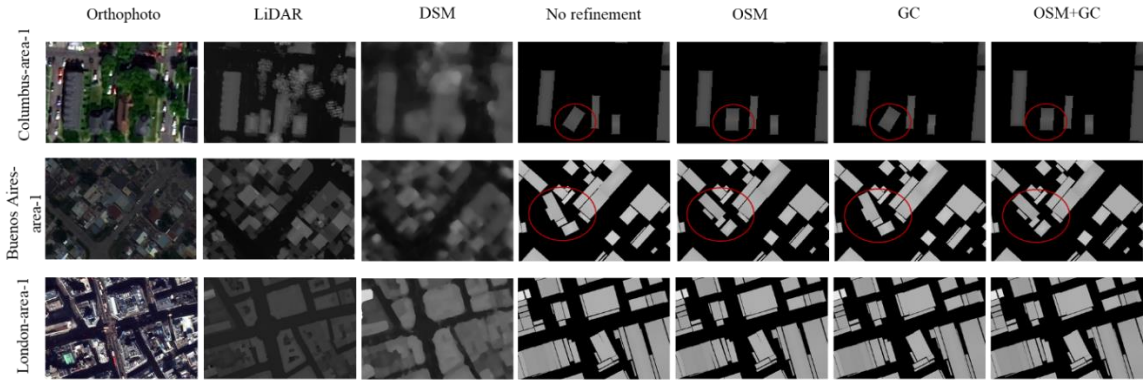

**Figure 2.14** Building model height result with different orientation refinement option. The red circled regions highlight notable differences among different strategies.

**Table 2.4** Accuracy evaluation of the resulting building models

| Region | Accuracy | DSM | No refinement | OSM | GC | OSM+GC |
|---|---|---|---|---|---|---|
| Columbus area 1 | IOU2 | 0.6929 | 0.6348 | *0.6359* | 0.6349 | **0.6362** |
|  | IOU3 | 0.6632 | 0.5780 | *0.5794* | 0.5782 | **0.5796** |
| Columbus area 2 | IOU2 | 0.8360 | 0.8076 | *0.8084* | 0.8076 | **0.8085** |
|  | IOU3 | 0.8287 | 0.7950 | *0.7961* | 0.7951 | **0.7962** |
| Buenos Aires area 1 | IOU2 | 0.6709 | 0.6378 | 0.6377 | *0.6380* | **0.6381** |
|  | IOU3 | 0.6551 | 0.5892 | 0.5888 | *0.5893* | **0.5894** |
| Buenos Aires area 2 | IOU2 | 0.3539 | *0.5301* | 0.5298 | **0.5302** | *0.5301* |
|  | IOU3 | 0.3281 | **0.4829** | 0.4825 | *0.4828* | 0.4827 |
| London area 1 | IOU2 | 0.6830 | 0.5797 | **0.5799** | 0.5796 | *0.5798* |
|  | IOU3 | 0.6176 | 0.4711 | **0.4714** | 0.4707 | *0.4712* |
| London area 2 | IOU2 | 0.6004 | **0.5053** | *0.5052* | 0.5048 | *0.5052* |
|  | IOU3 | 0.5626 | **0.4180** | **0.4180** | 0.4160 | 0.4163 |

**2.5.4 Evaluation of semantic segmentation input**

The input file of semantic segmentation for building class impacts the final accuracy of LoD-2 building model reconstruction. Therefore, the Columbus and London datasets evaluate the quantitative relationship between segmentation results and the building model. The accuracy of the resulting models is evaluated using the IOU2 and IOU3 metrics, respectively, to assess the 2D building footprint extraction accuracy and the 3D fitting accuracy.

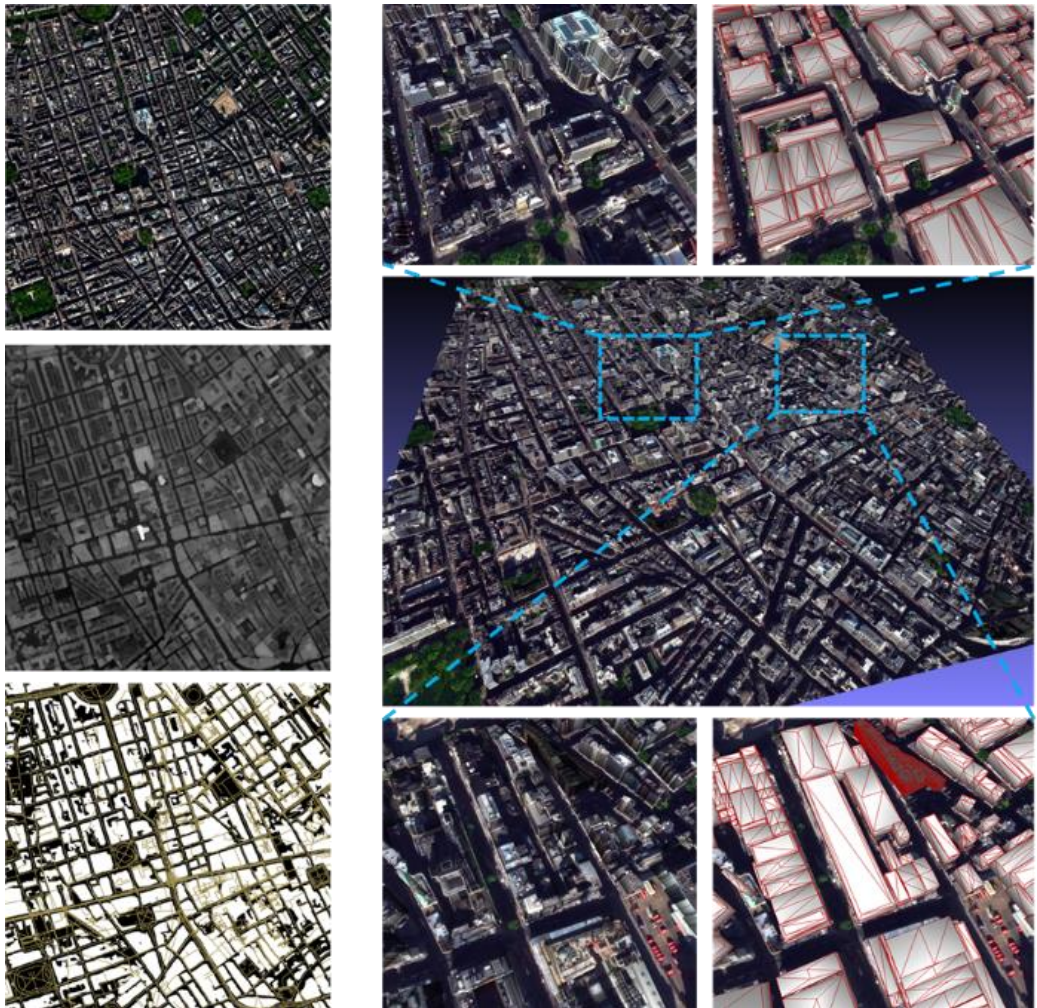

**Figure 2.15** Example result for London-area-1. First column from up to bottom: Orthophoto, DSM, building segmentation map using HRNet, with OSM street lines. Second column: 3D building model and detailed view of two specific areas

HRNet is a capable segmentation model to train and predict building masks, while alternatives such as UNet, Swin-transformer, or other recent semantic segmentation methods can be used as basic building masks to reconstruct building 3D models. We have trained the semantic segmentation network of UNet (Rongjun Qin et al. 2019), HRNetV2 (K. Sun et al. 2019), and Swin-transformer (Z. Liu et al. 2021) with the same training dataset from 2019 IEEE GRSS Data Fusion Contest (Le Saux et al., 2019) and INRIA aerial image labeling benchmark (Maggiori et al., 2017). In order to ensure that only input building classification maps have a difference in evaluation, OSM shapefiles are included for all experiments, and the reconstruct parameters adapt the default values of $\{T_l, T_d, T_{h1}, T_{h2}\}=\{90, 10, 0.5, 0.1\}$ to generate the building model. **Figure 2.16** indicates the orthophoto and segmentation results from three methods. It can be seen in **Table 2.5** that the building model reconstruction accuracy highly depends on the accuracy of the initial building segmentation result.

**Table 2.5** Accuracy evaluation of different segment input

| Region | Accuracy | UNet | HRNet | Swin-T |
|---|---|---|---|---|
| Columbus area 1 | IOU2 (segment) | 0.6927 | 0.7154 | 0.6320 |
|  | IOU2 | 0.6310 | 0.5664 | 0.6198 |
|  | IOU3 | 0.5636 | 0.5129 | 0.5459 |
| Columbus area 2 | IOU2 (segment) | 0.8355 | 0.8495 | 0.7577 |
|  | IOU2 | 0.7984 | 0.7572 | 0.7620 |
|  | IOU3 | 0.7836 | 0.7465 | 0.7416 |
| London area 1 | IOU2 (segment) | 0.5866 | 0.6854 | 0.6996 |
|  | IOU2 | 0.4019 | 0.5421 | 0.6547 |
|  | IOU3 | 0.3578 | 0.4345 | 0.4908 |
|  | IOU2 (segment) | 0.4856 | 0.6035 | 0.4484 |
| London area 2 | IOU2 | 0.3845 | 0.4638 | 0.4182 |
|  | IOU3 | 0.3128 | 0.4092 | 0.2856 |

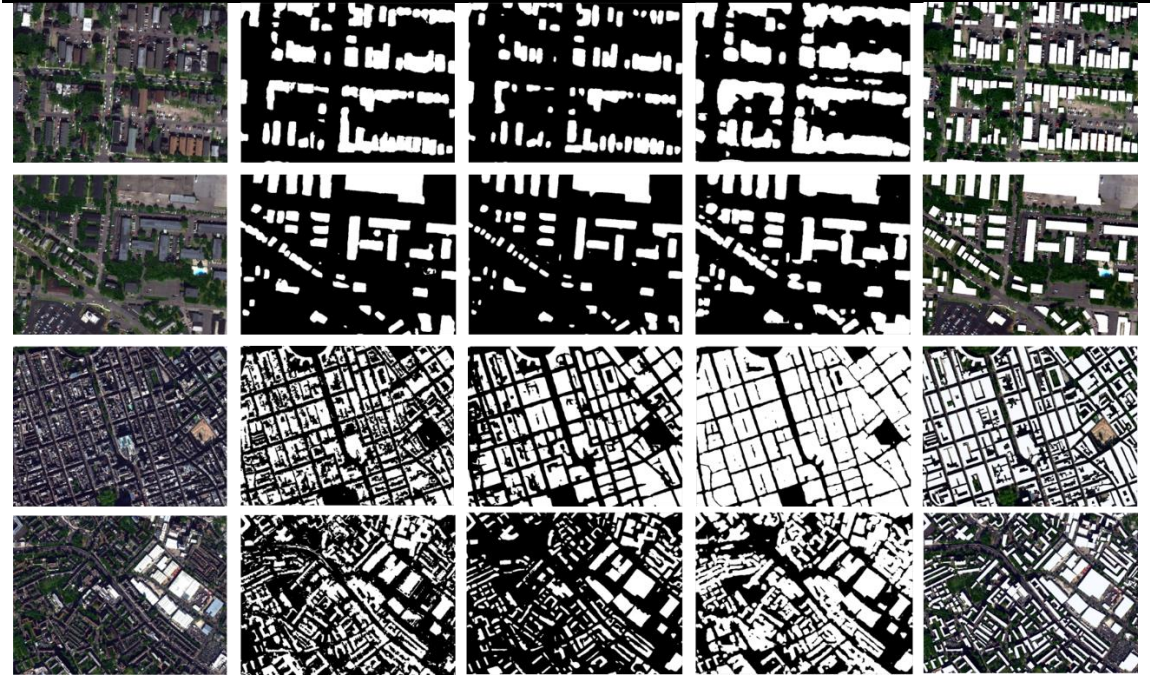

**Figure 2.16** Orthophoto and building segmentation map in example areas of Columbus-area-1, Columbus-area-2, London-area-1, and London-area-2, from left to right: orthophoto, segmentation from UNet, segmentation from HRNet, segmentation from Swin-transformer, and ground truth

### 2.5.5 Comparative study

We compare our results with results generated by state-of-the-art methods, and given the nature of LoD-2 model reconstruction method being component-rich, trivial and often ad-hoc, we re-implement and compare the key components of some of the existing methods on their algorithms for building polygon extraction and decomposition, these methods being: the method in (Le Saux et al. 2019), (Wei, Ji, and Lu 2019a), (Arefi and Reinartz 2013a), and that in (Li et al. 2019). The two key components we are comparing against existing methods are: 1) building polygon extraction (process described in **Section 2.4.2**), and 2) building polygon decomposition (process described in **Section 2.4.3**). The building polygon extraction methods include 1) a generalization of line segments-based building outline extraction method (Partovi et al. 2019), which generates building mask by applying SVM classification to gradient feature from PAN image and extract primitive boundary points by using SIFT algorithm (Lowe 2004a), and then fit boundary line segments from points and regularized them by finding building's orientation; 2) a toward automatic building footprint delineation by using CNN and regularization method (Wei, Ji, and Lu 2019b), which firstly segments buildings via FCN with multiple scale aggregation of feature pyramids from convolutional layers, and next regularizes polygon by adapting a coarse and fine polygon adjustment; 3) a minimum bounding rectangle (MBR) based method (Arefi and Reinartz 2013b), which approximates the remaining polygon by calculating bounding rectangle of building segment and the difference mask between bounding rectangle and building mask, with MBR-based algorithm for rectilinear building and RANSAC-based approximation algorithm for non-rectilinear building. The rectangle decomposition methods include 1) a parallel line-based building decomposition method (Partovi et al. 2019), which moves line segments until it meet the buffer of another parallel line segment and then generates rectangles using these two parallel line segments; 2) a primitive-based 3d building modeling method (Li et al. 2019), which cascades a building into a set of parts via mask R-CNN, and uses a greedy approach to select and move instance with largest IOU to decompose the building into a set of shapes. To evaluate the performance of building polygon extraction methods and rectangle decomposition method individually, polygon extraction methods share the same building mask input from **Section 2.4.1**, and rectangle decomposition methods share the same building polygon from **Section 2.4.2** and IOU2 is used as the metric for evaluation. The building polygon extraction from (Partovi et al. 2019) was realized by using SIFT algorithm to detect key points and those points are selected as boundary points only around the building mask from **Section 2.4.1**. For (Wei, Ji, and Lu 2019b) and (Arefi and Reinartz 2013b), primitive building segment is replaced by using building mask from **Section 2.4.1**.

**Figure 2.17** shows the sample result of all methods for building polygon extraction, **Figure 2.18** shows the sample result of all methods for building polygon decomposition to building rectangles, and **Table 2.6** gives the statistics. It can be seen that our method in both of the tasks clearly outperforms all the existing methods, especially in **Figure 2.17-Figure 2.18,** our methods due to the nature of deeply integrating decision criteria on the image color similarity and DSM continuity through a grid-based approach, the detection and decomposition algorithms are able to identify regularized polylines, and identify separating boundaries for building rectangles. As compared to all other existing approaches, which either miss detections or incorrectly locate or identify building rectangles. Examples showing the limitation of our method can be found in **Figure 2.17,** the last row, in which our method has failed to separate a connected building and created a minor artifact due to shadows, whereas other methods in this example are found to be worse in decomposition. **Table 2.6** shows that among the six regions we have experimented on, the IOU2 of our results achieves the highest for both tasks, except for Buenos-Aires-area 2 and London-area 2, which is marginally lower.

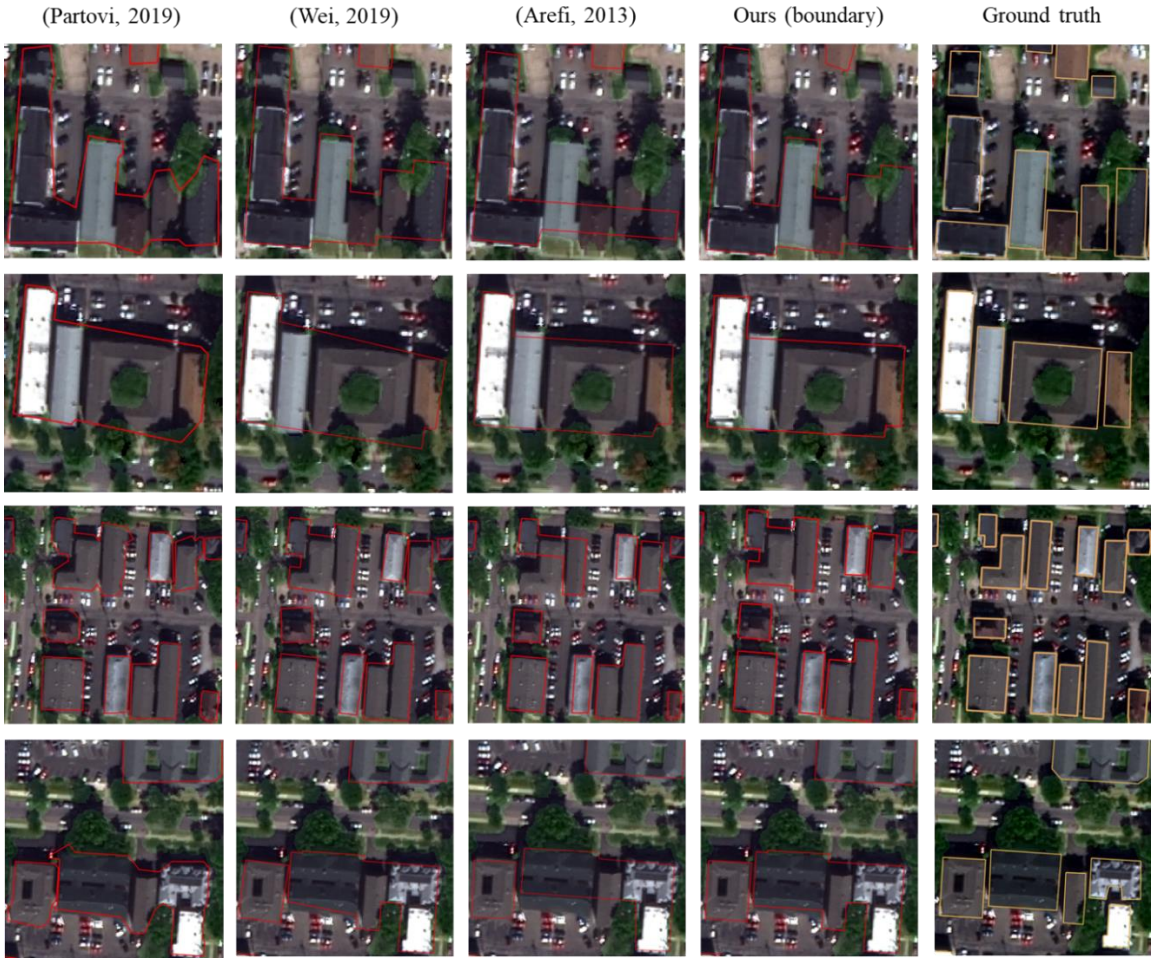

**Figure 2.17** Comparative results on building polygon extraction with other state of the art methods.

**Table 2.6** IOU2 values of comparing results in building footprint extraction and decomposition (Bold values indicate the best performing approach and italic the second best)

| Method | Columbus area1 | Columbus area2 | Buenos Aires area 1 | Buenos Aires area 2 | London area 1 | London area 2 |
|---|---|---|---|---|---|---|
| (Partovi, 2019) | *0.6827* | 0.8157 | 0.6456 | **0.6109** | *0.6623* | 0.5744 |

| | | | | | | | |
|---|---|---|---|---|---|---|---|
| Building Polygon Extraction | (Wei, 2020) | 0.6811 | *0.8179* | *0.6543* | 0.6093 | 0.6601 | **0.5765** |
| | (Arefi, 2013) | 0.6433 | 0.7761 | 0.5381 | 0.5778 | 0.4545 | 0.4639 |
| | Ours | **0.6831** | **0.8195** | **0.6549** | *0.6099* | **0.6624** | *0.5749* |
| Rectangle Decomposition | (Partovi, 2019) | *0.6579* | *0.7679* | 0.5976 | 0.5532 | 0.5745 | 0.4848 |
| | (Li, 2019) | 0.5768 | 0.7618 | *0.5980* | *0.5533* | *0.5781* | *0.5110* |
| | Ours | **0.6587** | **0.8133** | **0.6375** | **0.5969** | **0.5796** | **0.5136** |

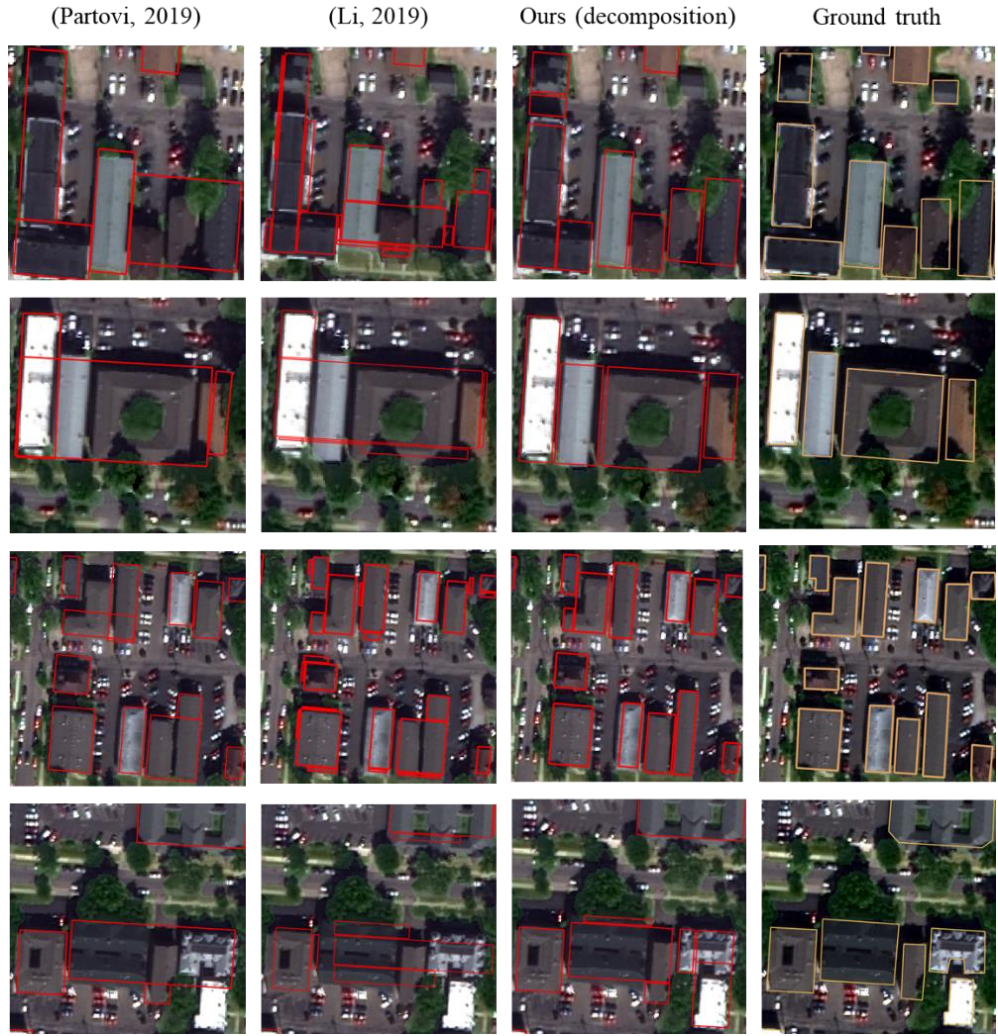

**Figure 2.18** Comparative results on building polygon decomposition with other state-of-the-art methods.

### 2.5.6 Parameter analysis

With several tunable parameters in the proposed approach, it is worth discussing the sensitivity of the parameters in **Table 2.7**. All five thresholds listed in **Table 2.7** contribute to the 2D building shape in the approach, and are assigned based on the empirical studies in the experiment. Two areas in Columbus are taken to analyze the impact of each threshold. A pair of initial thresholds are separately assigned to those two experimental areas, with $\boldsymbol{T_1} = \{T_w, T_l, T_d, T_{h1}, T_{h2}\} = \{0.16, 120, 10, 1, 0.2\}$ for Columbus-area-1 and $\boldsymbol{T_2} = \{T_w, T_l, T_d, T_{h1}, T_{h2}\} = \{0.2, 120, 10, 1, 0.2\}$ for Columbus-area-2. IOU2 of building rectangles after building polygon decomposition is calculated to represent the performance of different pairs of thresholds. The initial IOU2 of Columbus-area-1 with threshold $\boldsymbol{T_1}$ equals to 0.6584, and 0.8131 for Columbus-area-2 with threshold $\boldsymbol{T_2}$. **Figure 2.19** shows the relationship and difference between IOU2 with individually changed parameters $\boldsymbol{T'}$ and IOU2 with initial parameters $\boldsymbol{T_1}$ and $\boldsymbol{T_2}$, equals to $\boldsymbol{T'} - \boldsymbol{T_i}$, with $\boldsymbol{T'}$ means the pair of threshold that solely change one threshold based on $\boldsymbol{T_i}$, and $i = \{1,2\}$. **Figure 2.19** (a) shows the decreasing trend of IOU2 when the weight threshold $T_w$ close to 0.1 in Columbus-area-1; **Figure 2.19** (b) to (e) show that the tuning thresholds leads to an influence lower than 0.005 of IOU2, which represents there are minors influences once those thresholds are assigned as different values. Therefore, the sensitivity of several major threshold is reliable for proposed approach.

**Table 2.7** List of tunable parameters of proposed approach

| Parameter | Section | Description |
| --- | --- | --- |
| Weight threshold $T_w$ | Section 2.4.1.1 | A threshold of decision weight of a bounding box, to determine whether to use building segment from Mask R-CNN. |
| Length threshold $T_l$ (pixel) | Section 2.4.2 | A threshold of summed up length for determining building main orientations. |
| Color difference threshold $T_d$ (RGB) | Section 2.4.3 | A threshold of mean color differences ($|\overline{C_1} - \overline{C_2}|$) of the two rectangles to decide whether to merge two nearby rectangles in building decomposition. |
| Mean height difference threshold $T_{h1}$ (m) | Section 2.4.3 | A threshold of mean height difference ($|\overline{H_1} - \overline{H_2}|$) between two nearby rectangles to decide whether to merge two nearby rectangles in building decomposition. |
| Gap threshold $T_{h2}$ (m) | Section 2.4.3 | A threshold of dramatic height changes in a buffered region that cover the common edge between two nearby rectangles between two nearby rectangles. |

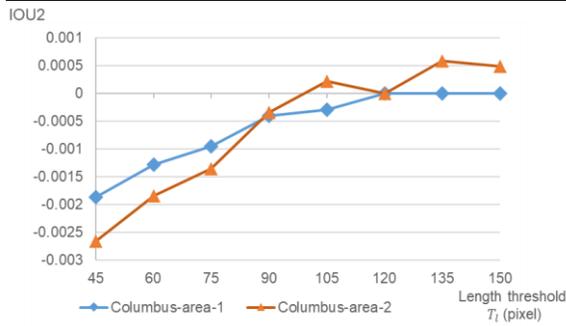

(a)

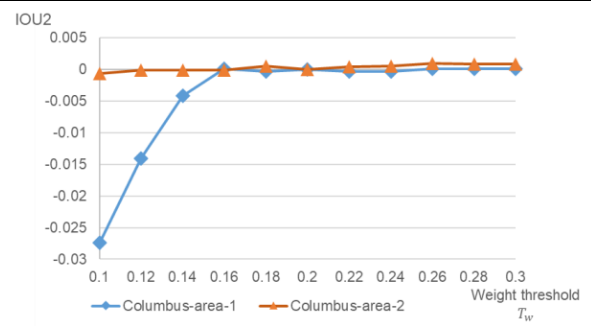

(b)

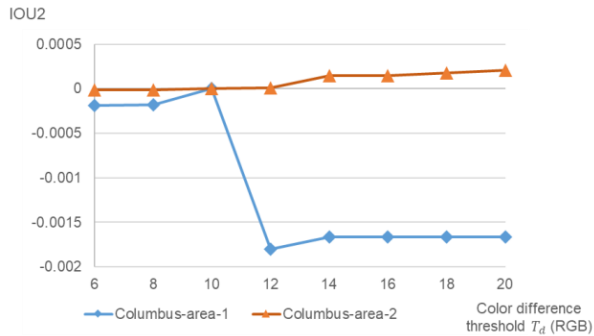

(c)

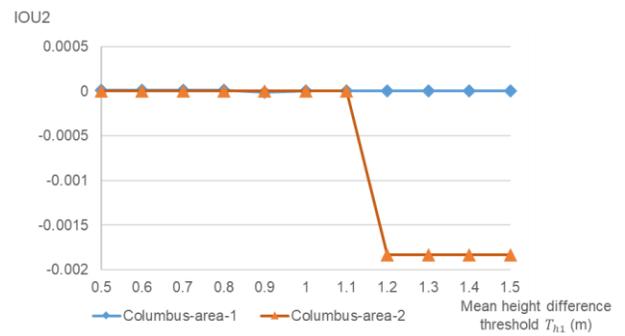

(d)

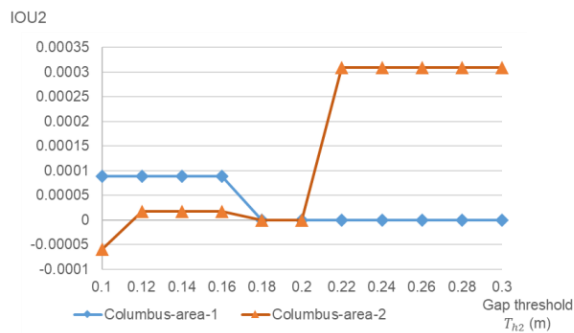

(e)

**Figure 2.19** The IoU2 of decomposed building rectangles difference between initial setting and changed thresholds. Columbus-area-1 initial $IOU2 = 0.6584$, Columbus-area-2 initial $IOU2 = 0.8131$. (a) the relationship between weight threshold $T_w$ and IOU2; (b) the relationship between length threshold $T_l$ and IOU2; (c) the relationship between color difference threshold $T_d$ and IOU2; (d) the relationship between mean height difference threshold $T_{h1}$ and IOU2; (e) the relationship between gap threshold $T_{h2}$ and IOU2;

**2.6 Conclusion**

In this part of section, we propose a LoD-2 model reconstruction approach performed on DSM and orthophoto derived from very-high-resolution multi-view satellite stereo images (0.5-meter GSD). The proposed method follows a typical model-driven paradigm that follows a series of steps including: instance level building segment detection, initial 2D building polygon extraction, polygon decomposition and refinement, basic model fitting and merging, in which we address a few technical caveats over existing approaches: 1) we have deeply integrated the use of color and DSM information throughout the process to decide the polygonal extraction and decomposition to be context-aware (i.e., decision following orthophoto and DSM edges); 2) a grid-based decomposition approach to allow only horizontal and vertical scanning lines for computing gradient for regularized decompositions (parallelism and orthogonality). Six regions from three cities presenting various urban patterns are used for experiments and both IOU2 and IOU3 (for 2D and 3D evaluation) are evaluated, our approaches have achieved an IOU2 ranging from 47.12% to 80.85%, and an IOU3 ranging from 41.46% to 79.62%. Our comparative studies against a few state-of-the-art results suggested that our method achieves the best performance metrics in IOU measures and yields favorably visual results. Our parameter analysis indicates the robustness of threshold tuning for the proposed approach.

Given that our method assumes only a few model types rooted in rectangle shapes, the limitation is that the proposed approach may not perform for other types of buildings such as those with dome roofs and may to over-decompose complex-shaped buildings. It should be noted the proposed method involves a series of basic algorithms that may involve resolution-dependent parameters, and default values are set based on 0.5 meter resolution data and can be appropriately scaled when necessarily processing data with higher resolution, while the authors suggest when processing with higher resolution data that are potentially sourced from airborne platforms, bottom-up approaches or processing components can be potentially considered to yield favorable results. The proposed approach developed in this paper, is specifically designed for satellite-based data that rich the existing upper limit of resolution (0.3-0.5m GSD) to accommodate the data uncertainty and resolution at scale. In the region with numerous compact blocks, the proposed approach capability is limited to reconstruct the roof structure of those blocks.

# Chapter 3. Unit-Level Lod2 Building Reconstruction for Complex and Dense Urban Regions

This chapter is based on the paper "Unit-Level Lod2 Building Reconstruction from Satellite Derived Digital Surface Model And Orthophoto" that was accepted by the "ISPRS. Annals. Photogramm. Remote Sens. Spatial Inf. Sci (2024)" by Shengxi Gui, Philipp Schuegraf, Ksenia Bittner, Rongjun Qin, and the unpublished work to achieve circular-based building model reconstruction.

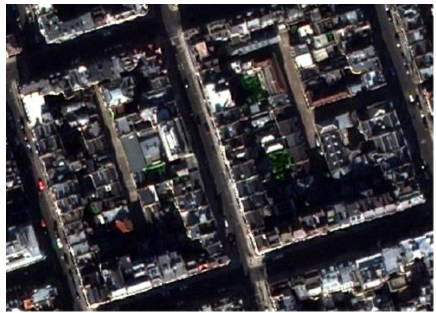
(a) Orthophoto for dense urban region

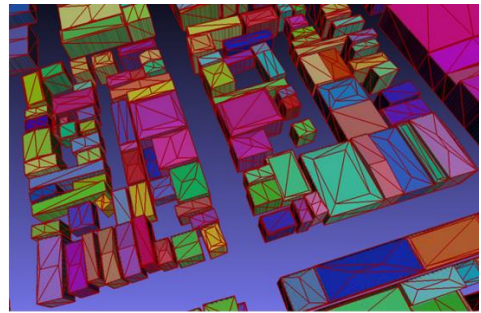
(b) Unit-level 3D building models

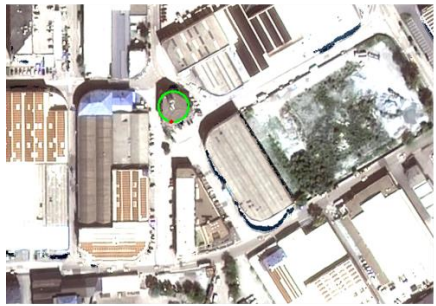
(c) Orthophoto for circular building

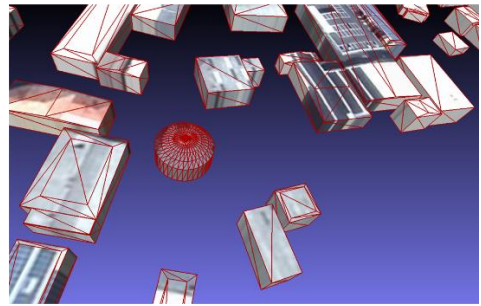
(d) 3D building models for circular building

**Figure 3.1** Sample figure for unit-level LoD2 building reconstruction. (a) Orthophoto for weak texture buildings; (b) 3D building models by using unit-level reconstruction; (c) Orthophoto for region include circular building; (d) 3D building model for circular building

## 3.1 Chapter Abstract

Recent advancements in deep learning have enabled the possibility to identify unit-level building sections from very-high-resolution satellite images. By learning from the examples, deep models can capture patterns from the low-resolution roof textures to separate building units from complex or duplex buildings. This section demonstrates that such unit-level segmentation can further advance level of details (LoD)2 modeling. We extend a building boundary regularization method by adapting noisy unit-level segmentation results. Specifically, we propose a novel polygon composition approach to ensure the individually segmented units within a duplex building or dense adjacent buildings are consistent in their shared boundaries. Besides, other than rectangular building units, circular buildings are significant for urban region 3D reconstruction. Therefore, we propose a novel gradient-base circle detection and roof reconstruction method to generate the LoD2 model for circular, sector, or ring shape buildings. Results of the experiments show that, our unit-level LoD2 modeling has favorably outperformed the state-of-the-art LoD2 modeling results from satellite images.

## 3.2 Introduction

### 3.2.1 Background

Level of Detail (LoD)2 building models describe architectural features and topological structures of building roofs (Gröger et al. 2008; Biljecki, Ledoux, and Stoter 2016), therefore, are of high interest in various applications such as mapping, urban planning, architectural design, virtual reality environments, and risk management. Typically, creating high-quality LoD2 models involves a manual and very expensive process, while recent research efforts aim to automate this process. Out of many sources, very-high-resolution (VHR) satellite stereo imagery (with ground sampling distance (GSD) < 1m) is beneficial due to its global coverage and low cost per unit area (Facciolo, de Franchis, and Meinhardt-Llopis 2017; S. Li et al. 2023; Huang, Qin, and Chen 2018). Past literature has shown that it is possible to reconstruct LoD2 (Shengxi Gui and Qin 2021; S. Gui, Qin, and Tang 2022; Partovi et al. 2019) from such data, which typically follow a standard process takes preprocessed digital surface model (DSM) and orthophotos from stereo satellite imagery as input data: first, perform building detection to obtain building masks; second, vectorize individual building masks with topologically consistent line primitives, third, determine the types of roofs and then join individual small buildings into more complex building models. Although these methods produce reasonable results for individual and single structured buildings, due to the lack of resolution of satellite images, reconstructing models in densely built areas and duplex buildings remains a significant challenge (Chen et al. 2018).

Challenges of reconstructing duplex buildings, or buildings in densely built regions, arise from the difficulties of building segmentation algorithms to identify distinct boundaries for duplex and adjacent buildings (Xingliang Huang et al. 2023) based on the mere low-resolution orthophoto and DSM. Duplex building consists of two or more separate units, typically side-by-side or stacked on top of each other, sharing a common wall but operating independently. Complex buildings usually consist of several rectangular units with similar roof material and texture. Complex/duplex buildings are often constructed from smaller, and contextually separate building units, while it is traditionally extremely challenging to infer information at the unit-level. The recent work in (Schuegraf et al. 2023) has shown that, by learning from examples, it is possible to infer unit-level segmentation from the low-resolution image textures, which, if successfully applied, can be used to extract unit-level models for LoD2 model reconstruction.

In addition, while the predominant building shape in most regions is rectangular, there exists a noteworthy presence of buildings with unconventional shapes, including circular, which, although not abundant, play a significant role in the architectural diversity of the area. Consequently, 3D reconstruction is essential for accurately including circular buildings to represent the complex urban building features. However, reconstructing circular buildings presents unique challenges: 1) the edge features in satellite imagery are less distinct than in natural images, complicating the direct detection of circles from orthophotos; 2) many circular buildings exhibit incomplete circular forms, including sectors, rings, C-shapes, and S-shapes, further complicating their identification and modeling (San and Turker 2010).

In this part, we integrate unit-level building segmentation with building model reconstruction, introducing an effective composition method for level of details LoD2 building model generation including rectangular and circular buildings. This method preserves boundary consistency among segmented units in duplex buildings. Upon evaluation in seven distinct regions, the experiments show that our unit-level LoD2 modeling approach significantly outperforms existing LoD2 models derived from satellite imagery in terms of accuracy and detail.

### 3.2.2 Related works

As mentioned earlier, reconstructing building models from photogrammetric data typically entails a multi-stage process that starts with the detection of building masks. This is followed by the extraction of 2D parameters (regularized footprint), and concludes with the derivation of 3D parameters (roof primitive) using specialized algorithms (Alidoost, Arefi, and Tombari 2019; Partovi et al. 2019).

**Building Segmentation**: Building segmentation has moved towards unit-level segmentation. Hence, recent studies have not only tackled the segmentation of each pixel as building or nonbuilding, but also into instances. PolyMapper (Li et al., 2019) directly predicts building and road vectors on the instance level. Approximating shapes in images with polygons (ASIP) (Z. Li, Wegner, and Lucchi 2018) surpasses the performance of PolyMapper on the CrowdAI (Mohanty et al. 2020) benchmark dataset for building instance segmentation. Then, Frame Field Learning set new standards for building instance segmentation on the CrowdAI benchmark by first predicting a pixel-level segmentation of buildings and building borders together with a map of two tangent directions per pixel. The tangent directions, called Frame Field, are used in an iterative optimization procedure to produce building polygons with regular appearance. Exceeding the performance of Frame Field Learning on CrowdAI, PolyWorld (Zorzi et al. 2022) is an end-to-end trainable building instance segmentation approach. It includes multiple steps of extracting vertices and learning the adjacency matrix that is used to connect the vertices. This procedure is error-prone, since a false negative vertex can strongly alter the appearance of the predicted polygon. Missing links in the adjacency matrix can cause missing polygons. Furthermore, PolyWorld does not separate directly adjacent buildings. Tackling this issue, Schuegraf et al. (2023) predicts separation lines between buildings together with the building segment and use the watershed transform to robustly predict building instances. The results of Schuegraf et al. (2023) surpass those of Frame Field Learning for complex urban scenarios.

**Footprint regularization**: The process of extracting regularized building footprints begins with the vectorization of images into regularized polylines, designated as building boundaries, subsequently generating rectangular-shaped building footprints. The preliminary processing of building segments employs shape reconstruction methods, such as alpha-shape (Kada and Wichmann 2012) and Hough Transform (San and Turker 2010), to establish initial building boundary formation. These are further refined through polyline simplification techniques, including the Random Sampling Consensus (RANSAC) Schnabel et al. 2007 and the Douglas–Peucker algorithm (Douglas and Peucker 1973). Next, post-processing of line segments from polyline can be facilitated based on orthophoto and algorithms like line segment detector (LSD) (Von Gioi et al. 2008), KInetic Polygonal Partitioning of Images (KIPPI) (Bauchet and Lafarge 2018), and PolyCity (W. Li et al.

2023). Subsequent steps involve further decomposition to delineate individual buildings, aligning them with preliminary rectangular or circular 2D models. An illustrative method is the orthogonal line-based 2D rectangle extraction technique by (Partovi et al. 2019), which decomposes building footprints into rectangle shapes starting from the longest boundary lines.

**Model Reconstruction**: The methods for building 3D reconstruction from images are generally classified into two distinct strategies: bottom-up and top-down. The bottom-up, or datadriven approach, treats buildings as collections of roof planes and other elements, assembling them based on geometric relationships observed in DSMs and point clouds. This strategy may employ techniques such feature filling (Zhou, Cao, and Zhou 2016) and region growing (S. Sun and Salvaggio 2013) to merge the structural components. Conversely, the top-down, or model-driven approach, relies on a predefined library of 3D building models (Lafarge et al. 2010; H. Huang, Brenner, and Sester 2013). It selects the most suitable model for a given set of data (like DSMs or point clouds), but this method often requires complex processes or adaptable parameters to match the diverse nature of building architectures. Advanced techniques, including deep learning for object recognition and meshing, are increasingly being incorporated into primitives estimation (Q. Li et al. 2023, 3; Mao et al. 2023; Y. Wang, Zorzi, and Bittner 2021).

**Circular building detection and reconstruction**: Hough transformer is the most popular methods for circle detection from digital image (Yuen et al. 1990), which has been instrumental applied in the field of circular building detection (San and Turker 2010). It is a feature extraction technique used widely for detecting simple shapes such as circles and lines in images. Notably, (Yuen et al. 1990) work on the optimization of the Hough Transform for circle detection has laid the groundwork for subsequent advancements in this area. For circle detection, it maps points in the image space to circular curves in the parameter space, identifying the parameters of circles (center coordinates and radius) present in the image based on the accumulation of votes. (Ok and Başeski 2015) detected circular oil tanks in single panchromatic satellite images by leveraging radial symmetry, introducing an automated thresholding technique to isolate circular areas, and employing the circle support ratio to verify detected circles. (Turker and Koc-San 2015) developed a method for automatically extracting both rectangular and circular-shaped buildings from high-resolution optical spaceborne images, utilizing a blend of support vector machine (SVM) classification, Hough transformation, and perceptual grouping, to detect circular building including ring, S-shape, and C-shape.

### 3.3 Methods for unit-level building reconstruction

The method described for unit-level building reconstruction employs pre-processed satellite-derived DSM and Orthophoto data, along with image processing techniques and deep models, to create 3D geometric models of buildings (LoD). The input data, DSM and Orthophoto, can be generated through standard photogrammetric workflow using the provided Rational Polynomial Coefficients (RPC), as for example, our input data are generated by using the RSP (RPC stereo processor) software (Rongjun Qin 2016a). As shown in **Figure 3.2**, With this input data, the proposed workflow initiates with unit-level building segmentation and then reconstructing building models in 2D and 3D. Specifically, the unit-level semantic segmentation process aims to detect and segment discernable building units from Orthophoto and DSM, which stands for standard single-unit buildings, or multiple units of duplex buildings. The LoD2 building footprint extraction process extracts regularized rectangular building footprints from these individual building segments, further dividing bigger segments into basic building units. Finally, it utilizes the most appropriate building model with 3D primitives to represent the building units at 3D level.

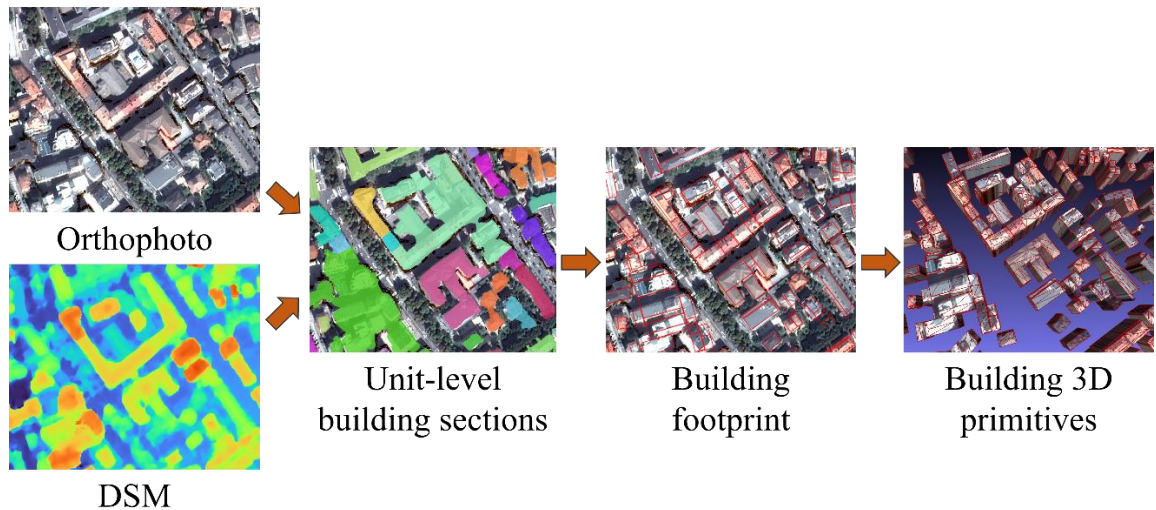

Figure 3.2 Workflow of unit-level building LoD2 model reconstruction

### 3.3.1 Unit-level semantic segmentation

Buildings in large cities exhibit complex structures comprising of various interconnected units and components. To perform as detailed as possible 3D reconstruction, the modeling of each building component as a separate unit is a correct way to proceed. Supporting the LoD2 modeling methodology, this paper employs a unit-level semantic segmentation strategy previously developed by (Schuegraf et al. 2023).

Our method uses a deep convolutional neural network at its core for a 3-class problem: building component, separation line and background. The inputs to the network are the DSM and orthorectified RGB satellite images tiled to patches of size 512×512 px. We employ the well-proven U-Net shape architecture, consisting in our case of two ResNet34 encoders for each input modality and one decoder. In order to maintain detailed spatial information, we aggregate feature maps acquired at four distinct scales from the two encoders through summation. These aggregated feature maps then serve as input for the full-scale skip-connections.

We follow the same training procedure as in (Schuegraf et al. 2023) and employ a combination of segmentation and regularization losses. To minimize the dissimilarity between predicted and true probability distributions, we incorporate the weighted *cross-entropy* loss function:

$$\mathcal{L}_{CE}(x, y, p, w) = -\sum_i y_i w_i \cdot \log(p(x_i)) \qquad (3.1)$$

to achieve accurate and meaningful outcomes in the context of multi-class semantic segmentation. Here, $y$ denotes the ground truth, $x$ is the input tensor, $p(.)$ is the *softmax* output of the neural network model, $i$ is the respective class and $w$ is an array of manually selected loss weighting coefficients, which we set to [1, 1, 4]. Often, the utilization of cross-entropy loss leads to smoothed or indistinct boundaries for objects. To suppress this issue and push the model towards more precise delineation of boundaries, we used the *generalized dice loss* (Sudre et al. 2017):

$$\mathcal{L}_{DICE}(x, y, p) = 1 - 2 \cdot \frac{\sum_i v_i \sum_n y_{in} \cdot p(x_i)_n}{\sum_i v_i \sum_n y_{in} + p(x_i)_n} \qquad (3.2)$$

where $v_i$ is the inverse frequency of the class $i$. $\mathcal{L}_{DICE}$ is developed for precise boundary detection.

To penalizes structural irregularities like curved corners or uneven edges we employed the topological loss (Mosinska et al. 2018):

$$\mathcal{L}_{TOP}(x, y, p)_C = \sum_{n=1}^{N} \sum_{m=1}^{M_n} \|l_n^m(y_C) - l_n^m(p(x)_C)\|_2^2 \qquad (3.3)$$

which minimizes the differences between the VGG19 (Simonyan and Zisserman 2015) descriptors of the ground-truth images and the corresponding predicted delineations, in our case for both the building and separation line classes separately. In **Equation (3.3)**, we denote the class on which to apply the term as $C$, $l_n^m$ describes feature map $m$ of layer $n$ of a pre-trained VGG19.

The final objective function combines three above-described losses:

$$\mathcal{L}_{TOTAL} = \mathcal{L}_{CE} + \mathcal{L}_{DICE} + \lambda_{BM} \cdot (\mathcal{L}_{TOP})_{BM} + \lambda_{TB} \cdot (\mathcal{L}_{TOP})_{TB} \qquad (3.4)$$

where $\lambda$ controls the influence of the topological term on the overall training procedure, abbreviations $BM$ and $TB$ relates to building mask and touching border classes.

Following this, a map representing instances of building sections is created through the application of the watershed transform (Beucher and Meyer 1992; Beucher and Meyer 2018) in a post-processing stage. Essentially, the watershed transform interprets the obtained three class maps, consisting of background, building, and separation line, along with a seed image and a mask, as a topographical surface. The seed map and mask are derived from the predicted information related to buildings and separation lines. Subsequently, the watershed transform simulates a flooding scenario, wherein water begins flooding from the seeds and settles into basins. These basins are delineated by watershed lines, aligning with high image intensities. The mask confines the virtual water flow to specific regions, and the enclosed regions marked by watershed lines are then identified as objects.

### 3.3.2 LoD2 building reconstruction

Upon obtaining unit-level building segments, we apply 2D footprint extraction process and 3D primitives computation process for each building unit to generate rectangular-based 3D building models. The Orthophoto and DSM, derived from very high-resolution satellite imagery, typically have spatial resolutions ranging from 0.3m to 1m. Due to this resolution constraint, accurately detecting small buildings and detailed roof structures from complex buildings remains a challenge. To address this, we employ a model-driven approach for 3D building reconstruction from studies (Shengxi Gui and Qin 2021; Partovi et al. 2019). This approach assumes that a complex building footprint can represented by 2D rectangles, thus it turns the LoD2 modeling problems into a topology fusion problems from 3D primitives buildings (extended from the 2D rectangle footprints).

In order to represent building footprints as regularized 2D shapes, unit-level building segments are vectorized into polygons and subsequently refined into rectangular footprints. This process contains three steps from satellite-derived data: initially, coarse boundary delineation is achieved using the Douglas-Peucker algorithm (Douglas and Peucker 1973), effectively primary vectorizing building segments into initial polylines of building boundaries. This is followed by a polyline adjustment step, where the main orientations of each building unit are calculated, and shorter line segments with similar orientations are merged into more extended line segments. The final step involves polyline regularization with the LSD algorithm (Von Gioi et al. 2008), aligning the orientations of line segments with detected line segments with texture information from Orthophotos. The culmination of this process not only identifies the main orientation of each building segment but also accurately vectorizes the building polygons from raster facilitating the extraction of rectangular building footprints without DSM data and refining building shapes from satellite images with enhanced precision.

The vectorized building footprints may still be in arbitrary polygon with a number of vertices. To facilitate the process of generating 3D primitives, it is necessary to decompose these polygons into multiple simple rectangles. Our approach employs a grid-based decomposition approach (Shengxi Gui and Qin 2021), predicated on the concept that complex building polygons can be fundamentally broken down into multiple, simpler rectangular entities, which then serve as the regularized building 2D model for the subsequent 3D reconstruction stage. The procedural workflow of this decomposition is divided into four distinct parts: First, for each unit, the 2D building polygon is rotated to align its primary local orientation with the x-axis. Second, initial separation of the building mask is performed, using DSM and Orthophoto gradients. Third, a three-tier image pyramid approach is applied to iteratively identify and refine the largest possible inner rectangles, progressing from the coarse to the finer layers. Finally, excessively segmented adjacent rectangles are consolidated, leveraging both Orthophoto and DSM data to ensure accurate and efficient footprint reconstruction. To determine if two neighboring rectangles should be merged, the criteria are adapted as **Equation (2.2)**.

After determining the rectangular footprint of each building unit, 3D roof structure can be fitted based on rectangular models derived from satellite-based DSM. These preliminary model shapes include five types of rectangular building roof models: flat, gable, hip, pyramid, and mansard, and each roof model represents a specific architectural style and primitives. A set of 3D parameters, including ridge height, eave height, and hip structure, is utilized to characterize the detailed roof primitives across all five model types (Shengxi Gui and Qin 2021). These parameters are computed and optimized through an exhaustive search strategy designed to identify the parameters set that minimizes the root mean square error (RMSE) between the fitted roof height and DSM data. The optimization includes iterative parameter updates informed by DSM, starting with the determination of terrain height as the local minimum of the building height. Despite the DSM data noise since resolution or stereo matching limitation, our exhaustive search approach efficiently selects the most accurate roof type and parameter set, maintaining computational accuracy even for buildings only with a few hundred pixels. The final output of this process includes detailed 3D parameters for the building model, which reconstructs buildings into LoD2 levels.

**3.4 Methods for circular building reconstruction**

Most of buildings have rectangular shape, and circular building reconstruction section detects a small number of circular buildings from urban regions. The method described for circular building reconstruction employs the same data as unit-level building reconstruction part including. Similar as rectangular-based building model reconstruction, the method to reconstruct circular-based building initiates with building segmentation and then reconstructing building models in 2D and 3D. Semantic segmentation process aims to detect and segment building masks from Orthophoto and DSM. The circle detection at 2D level aims to identify buildings with circular geometries, ranging from complete circles to partial circular forms such as sectors and rings, from building mask. Subsequently, the computation of circular building roof structures step selects the optimal roof type and calculates the corresponding 3D primitives.

### 3.4.1 Circle detection at 2D level

To express building footprints as regularized 2D circular shapes, individual building segments are transformed into polygons with key-points and then further refined into circular footprints for circular buildings. The circle detection contains 5 main steps shown as workflow in **Figure 3.3**: 1) initial polygon/key-point detection; 2) coarse circle center computation; 3) radius computation and non-candidate circle exclusion; 4) fine circle parameters computation via least squares; 5) radian computation and final shape decision.

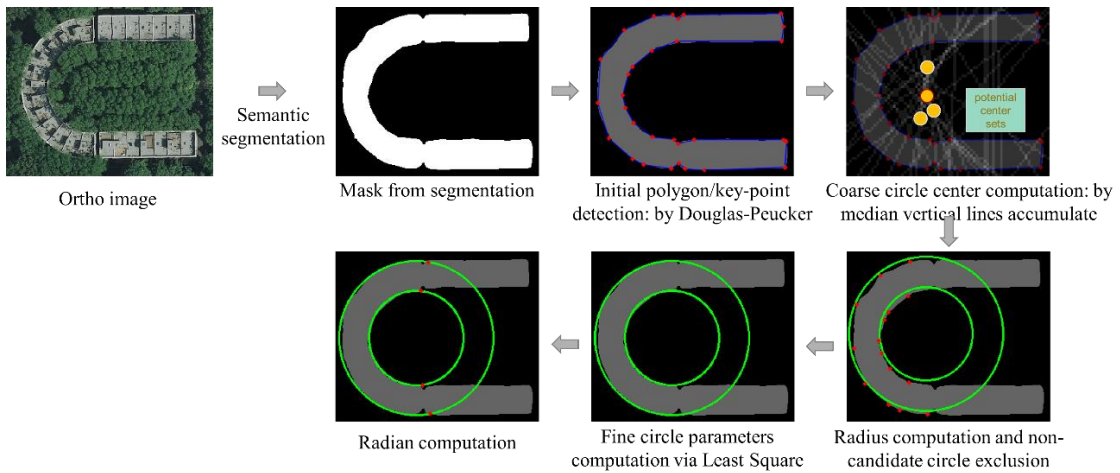

**Figure 3.3** Workflow of detect circular building from building mask, a sample of C-shape building has been shown in this workflow

The process begins with the transformation of the building mask into a simplified polygon utilizing the Douglas-Peucker algorithm, similar as the procedure outlined in the **Section 2.4.2** and **Section 3.3.2**, and this step provides key points that to represent the contour to further determine the circle parameters for the following steps.

Coarse circle center computation stage will compute an initial position of circle center based on key-points detected by using Douglas-Peucker algorithm. This step calculates the median point of each polygonised line segment and constructs perpendicular bisectors at these points. Next, a down-sampled grid space (a quarter of its original size as coarse grid) is adapted to count the intersections of these perpendicular bisectors in a large tolerance. The supe pixel (grid) with large perpendicular bisectors intersections indicates likely circle centers. As shown in **Figure 3.3** top right panel, several candidates of circle center are determined in coarse center computation step.

Radius computation and non-candidate circle exclusion proceeds to determine the final center and radius by measuring distances from the identified coarse center to the key-points. For each candidate circle center, the distance and the gradient of distance between candidate circle center and each key-points are computed. Apposite circle center will have a constant value for both distance and gradient of distance around radian arc, and if there are more than one ranges with gradient of distance values close to zero, the building may contain concentric circles such as C-shape and S-shape, which shows as **Figure 3.4**. Therefore, by analyzing the gradient of these distances and the distribution of key-points, a preliminary center location and radius are established. If the key-points form a continuous group within a certain distance, that distance is designated as the correct center and coarse radius. Conversely, if key-points are sporadically placed or encompass less than a $\pi/2$ radian arc (exclude rectangular buildings and noise patterns), the corresponding circle is deemed a non-candidate and excluded from further consideration.

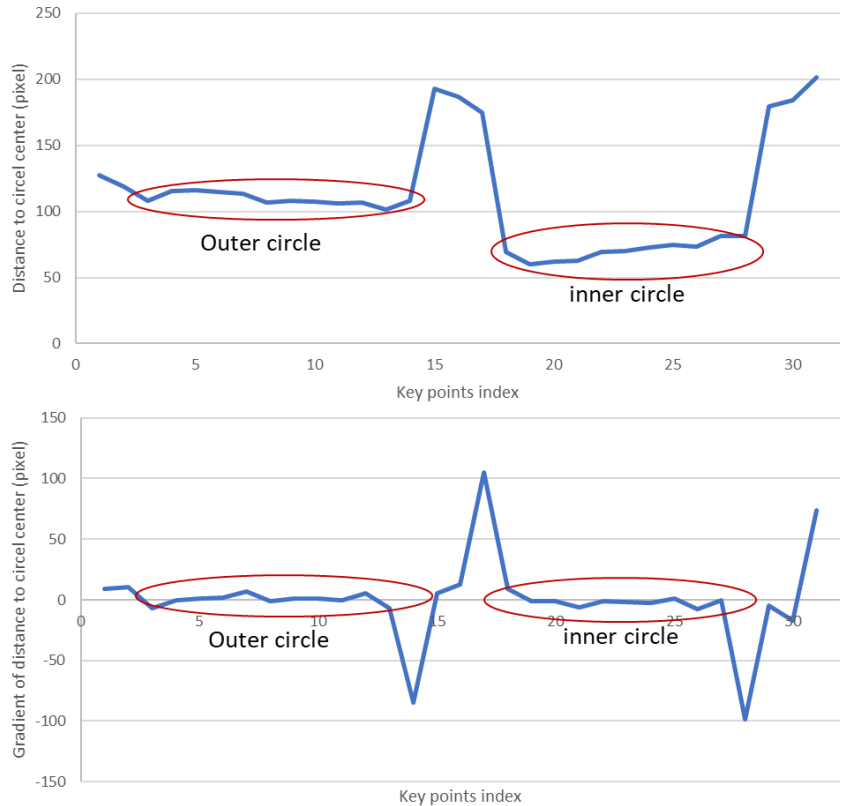

**Figure 3.4** Distance and gradient of distance between candidate circle center and each key-point, if key-points form a continuous group within a certain distance and gradient near to zero, there may be a correct circle.

Fine circle computation step leverages the coarse center and radius determined in the previous steps, along with key-points situated around the potential circular structure, the algorithm employs a Least Squares (LS) method to refine the circle's parameters. This computation enhances the accuracy of the circle center and radius, providing a detailed geometric representation (Chernov and Lesort 2005). The LS fitting for circle follows the formula:

$$F(x_c, y_c, r) = \sum_{i=1}^{N} [(x_i - x_c)^2 + (y_i - y_c)^2 - r^2]^2 \tag{3.5}$$

where $N$ is the number of key-points locate on circle, and $x_i, y_i$ are the coordinates of these points, and $x_c, y_c, r$ are center of circle and radius. This formulation directly applies the least squares principle by minimizing the squared differences between the squared distances from the points to the center of the circle and the square of the radius, and the initial values for center location $(x_{c0}, y_{c0})$ and radius $r_0$ are defined by coarse center and radius from previous step. By minimizing $F(x_c, y_c, r)$, we will get a best fit of fine center location and radius as the fine circle parameters.

The final step focuses on the determination of the circle arc range and length. By identifying the start and end key-points of the detected fine circle, the range and length of circular building are determined. Oil tank and circular tower cover full circle, while C-shape and S-shape buildings only consist of part of circle, or the combination of sector and rectangular building. Thus, in this step, the detail radian parameters are computed to better classify the circular building shape, whether it be a complete circle, a semicircle, or another variant, thereby enabling the decomposition of circular shapes from complex building masks.

### 3.4.2 Roof structure computation for circular building

This subsection elaborates on the methodology employed to determine the type and parameters of roofs on circular buildings, enhancing a specific model-driven method for circular building 3D reconstruction. The process is methodically broken down into distinct steps aimed at converting the two-dimensional detection problem into a one-dimensional analysis, thus facilitating a more nuanced understanding of roof structures.

Building height data derived from Digital Surface Models (DSMs) serve as the primary foundation for roof classification and primitive computation. Unlike rectangular buildings, circular structures, particularly those comprising only partial radial arcs, necessitate specific data preprocessing and selection to accurately model circular roofs. To address this, we exploit the inherent symmetry of circular forms, adapting height information for eight directions from circle center, each segmented by 45 degrees. This approach allows us to analyze height and its gradient, thereby converting the 2D spatial challenge into a linear (1D) problem, similar to the cost propagation step in Semi-Global Matching (SGM) (Hirschmuller 2007). This step utilizes only those directions encompassed by the building's radian as depicted in **Figure 3.5**, focusing on the height at points where the building mask intersects with the circular coverage.

The model library for circular buildings encompasses three roof types: flat, conical, and spherical, each embodying distinct architectural styles and primitives. As illustrated in the right part of **Figure 3.5**, flat roofs maintain a uniform height across the roof area, conical roofs exhibit the highest point at the center with a linear height decrease towards the edges, and spherical roofs also peak at the center but with a more rapid height decline near the edges. The classification method for circular roofs employs height and its gradients from the center to the edges across eight directions, analyzing both the first derivative (gradient) and the second derivative (quadratic gradient). The gradient of height serves to distinguish flat roofs from conical and spherical ones, while the quadratic gradient is specifically designed to differentiate between conical and spherical roofs. This gradient-based approach allows for a clear classification of roof types based on their geometric characteristics.

For each direction that provides a distance to the center and corresponding height information, the algorithm proceeds with parameter estimation for the detected roof types by combining roof height and distance to center for all directions. Parameters for flat (height is constant) and conical (linear relationship between height and distance) roofs are determined using a combination of Random Sample Consensus (RANSAC) (Schnabel, Wahl, and Klein 2007) and Least Squares fitting (O'Leary, Harker, and Zsombor-Murray 2005) to ensure robustness against outliers and achieve precise parameterization. Conversely, the parameters for spherical roofs are calculated employing RANSAC in conjunction with a circle-specific least squares method, tailoring the computation to accurately reflect the unique properties of spherical geometries.

This algorithm streamlines the intricate process of 3D reconstruction into a series of simplified tasks within a mathematical model framework. Unlike the exhaustive search methods employed for rectangular buildings, this approach significantly reduces computational demands to a minimal level, thereby accelerating the entire processing workflow.

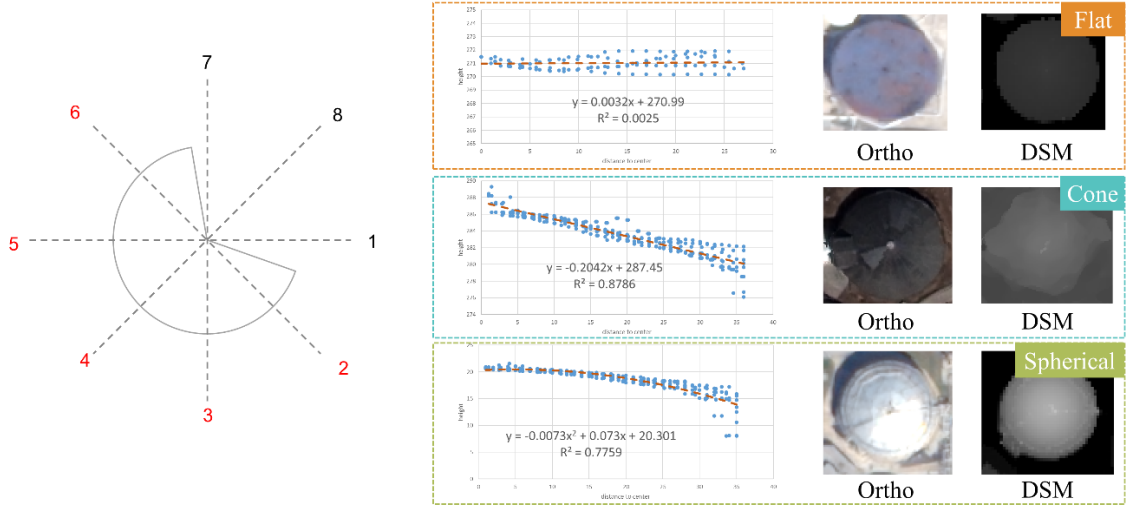

**Figure 3.5** Left: detected 2D circle with radian, and propagation from 8 direction, only the direction covered by radian is available (red direction). Right: typical height distribution for flat roof (upper) and cone roof (lower), from the figure, we can calculate the roof parameter by listing the relationship between distance to center and the height, flat roof has a consistent value of height, cone roof best fit a linear function, and spherical roof best fit a polynomial function.

## 3.5 Experiments

### 3.5.1 Study area

Our experiments include four cities, each exemplifying unique geographical locations and distinctive urban landscapes, including 1) Columbus, Ohio, a typical U.S. city characterized by low-density residential and industrial areas; 2) Buenos Aires, Argentina, a South American megacity, that contains a mix of sparsely populated residential areas and densely inhabited slums; 3) London, UK, a European megacity with a compact urban structure and high-density development; 4) Trento, Italy, a medium-sized European city with numerous adjacent buildings.

The accuracy of model-driven 3D building reconstruction, which deduces a set of 3D primitives based on texture and height data from individual building sections, depends entirely on the accuracy and comprehensiveness of the regularized building footprint in producing the final LoD2 model. Furthermore, the density or urban complexity represents the difficulty of 3D building reconstruction. In areas characterized by a high concentration of buildings, accurately delineating individual building perimeters becomes particularly challenging. This challenge is compounded in scenarios where adjacent buildings feature roofs with low texture contrast, often leading to the aggregation of multiple structures into a single reconstruction section, thereby adversely affecting the accuracy of the LoD2 model. Hence, the success of building reconstruction in densely populated regions is heavily dependent on the segmentation effectiveness for Orthophotos with weak textures. The 3-band (RGB) Orthophotos and DSMs for all study areas are generated using a multi-view stereo matching approach (RSP, Qin 2016, 2019) from multiple World-view-2 stereo pairs for the Columbus, London, and Trento dataset, and Worldview-3 for Buenos Aires dataset.

**Table 3.1** Study areas basic information and building density

| Region | Image size (pixel) | GSD (pixel size) | Location | Building area proportion | Building instances |
|---|---|---|---|---|---|
| **Columbus 1** | 1003 × 890 | 0.5 m | USA, North America | 0.2305 | 224 |
| **Columbus 2** | 1646 × 1118 | 0.5 m | USA, North America | 0.2866 | 151 |
| **Buenos Aires 1** | 3000 × 3000 | 0.3 m | Argentina, South America | 0.2704 | 352 |
| **Buenos Aires 2** | 3000 × 3000 | 0.3 m | Argentina, South America | 0.0587 | 111 |
| **London 1** | 3000 × 3000 | 0.5 m | UK, Europe | 0.5985 | 676 |
| **London 2** | 3000 × 3000 | 0.5 m | UK, Europe | 0.2968 | 910 |
| **Trento** | 3680 × 3309 | 0.5 m | Italy, Europe | 0.3114 | 1556 |

**Table 3.1** shows the information of each study area. In total, there are three low building density regions with small numbers of buildings, and most buildings are isolated, and four high building density regions have dense building distribution with large numbers of buildings, and many buildings are located immediately adjacent to neighborhoods.

### 3.5.2 Evaluation in 2D and 3D level

The evaluation of 2D segmentation and 3D geometry are evaluated separately using both a 2D intersection over union $IOU_{2D}$ and 3D intersection over union $IOU_{3D}$ based on manually created reference data for building footprint and light detection and ranging (LiDAR)-based DSM for 3D geometry (Kunwar et al. 2020). $IOU_{2D}$ assesses the accuracy of 2D building footprint extraction, while $IOU_{3D}$ evaluates the accuracy of 3D model fitting. The IOU2D and IOU3D are defined following as **Equation (2.13)** and **(2.14)**.

Other than unit-level building segmentation, two publicly available semantic segmentation methods for building footprint detection are compared to evaluate the performance at 2D and 3D levels. The first one is based on HRNetV2 (Y. Wang, Zorzi, and Bittner 2021) to get building segments by using Orthophoto with RGB bands. The training and validation datasets were combined with satellite and aerial imagery (S. Gui, Qin, and Tang 2022). The second one is High Resolution Land Cover Classification – USA (Ronneberger, Fischer, and Brox 2015; Robinson et al. 2019), developed by ESRI for ArcGIS multi-classes semantic segmentation. This approach uses the UNet model architecture and is trained based on aerial imagery with 0.8m-1m resolution, which can also be used to segment buildings.

**Figure 3.6** displays the segmentation results from three semantic-based segmentation methods and ground truth building mask. The visual comparison indicates that compared to normal semantic segmentation methods, our unit-level segmentation method can extract building sections from very dense urban and complex structure buildings. Besides, since the training dataset for ESRI's segmentation method is aerial imagery, the segmentation result in Buenos Aires regions (from Worldview-3) is not as good as other study regions (from Worldview-2).

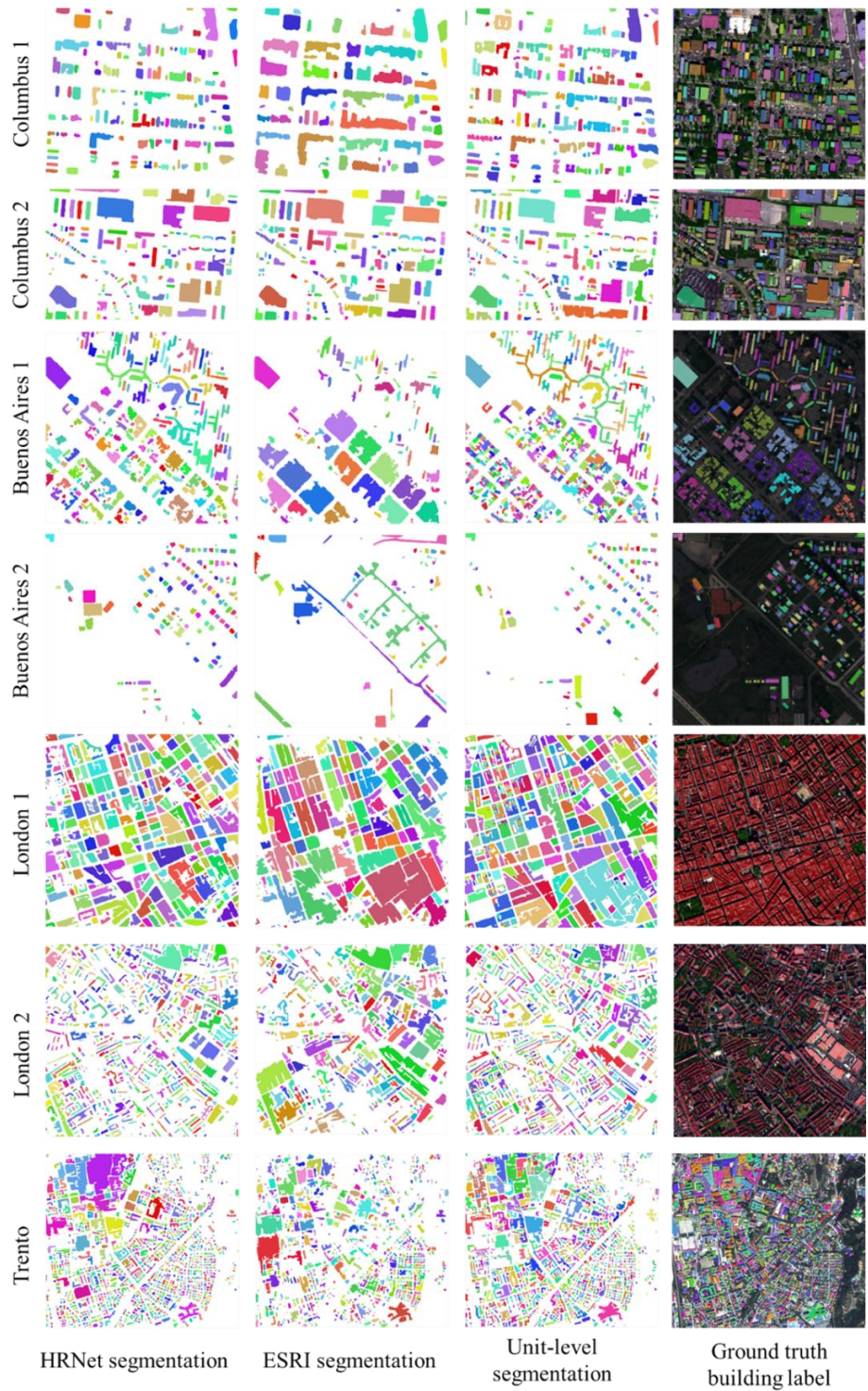

**Figure 3.6** Building semantic segmentation results for each study region and the ground truth

The numerical results comparing 2D and 3D levels are presented in **Table 3.2**. These findings reveal that unit-level building segmentation performs best in three regions for 2D building masks and in five regions for 3D building models. The overall accuracy indicates that the performance of building model reconstruction is largely contingent upon the initial building segmentation accuracy. Nonetheless, unit-level segmentation significantly enhances the accuracy of 3D primitives for each building section.

**Table 3.2** Accuracy comparison in 2D label (semantic segmentation), 2D footprint, and 3D model (reconstruction) for all regions. The difference in the method is the input of building mask.

| IoU | HRNet 2D label | ESRI 2D label | Unit-level 2D label | HRNet 2D footprint | ESRI 2D footprint | Unit-level 2D footprint | HRNet 3D model | ESRI 3D model | Unit-level 3D model |
|---|---|---|---|---|---|---|---|---|---|
| Columbus 1 | 0.7140 | 0.6647 | **0.7360** | 0.5389 | 0.6151 | **0.6248** | 0.4881 | 0.5342 | **0.5801** |
| Columbus 2 | **0.8492** | 0.8306 | 0.7980 | 0.7526 | **0.7815** | 0.7309 | 0.7403 | **0.7649** | 0.7196 |
| Buenos Aires 1 | **0.6706** | 0.4434 | 0.6324 | **0.6149** | 0.4043 | 0.5662 | **0.5610** | 0.3296 | 0.5314 |
| Buenos Aires 2 | 0.5635 | 0.1534 | **0.6010** | 0.4965 | 0.1233 | **0.5255** | 0.4480 | 0.0932 | **0.4970** |
| London 1 | 0.6826 | 0.5993 | **0.7471** | 0.5668 | 0.5222 | **0.6265** | 0.3857 | 0.3067 | **0.4382** |
| London 2 | 0.5974 | 0.4846 | **0.6115** | 0.4882 | 0.4404 | **0.5377** | 0.4154 | 0.3348 | **0.4728** |
| Trento | **0.6578** | 0.4071 | 0.6400 | 0.5808 | 0.3573 | **0.6021** | 0.3010 | 0.1418 | **0.3311** |

**Figure 3.7** and **Figure 3.8** display building 3D models in two high building density regions, London area 1 and Trento area. From instance-level comparison, it indicates that unit-level segmentation of building masks effectively divides complex or densely packed buildings in areas with weak textures into individually segmented units, maintaining consistency along their shared boundaries, and then computing fine 3D roof parameters. In contrast, building models generated by the other two methods often treat complex structures or buildings in close proximity as a single and large building section.

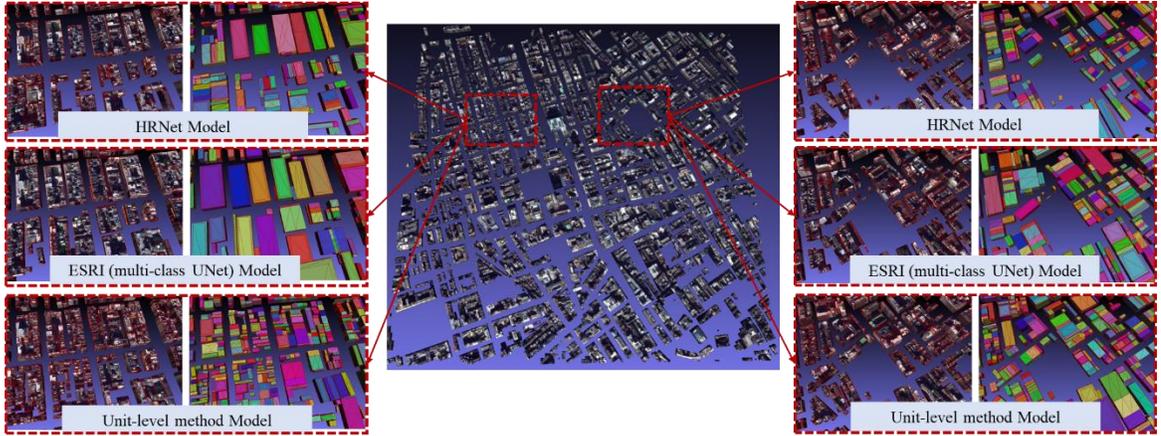

**Figure 3.7** Building 3D models in London 1 region with dense urban structures

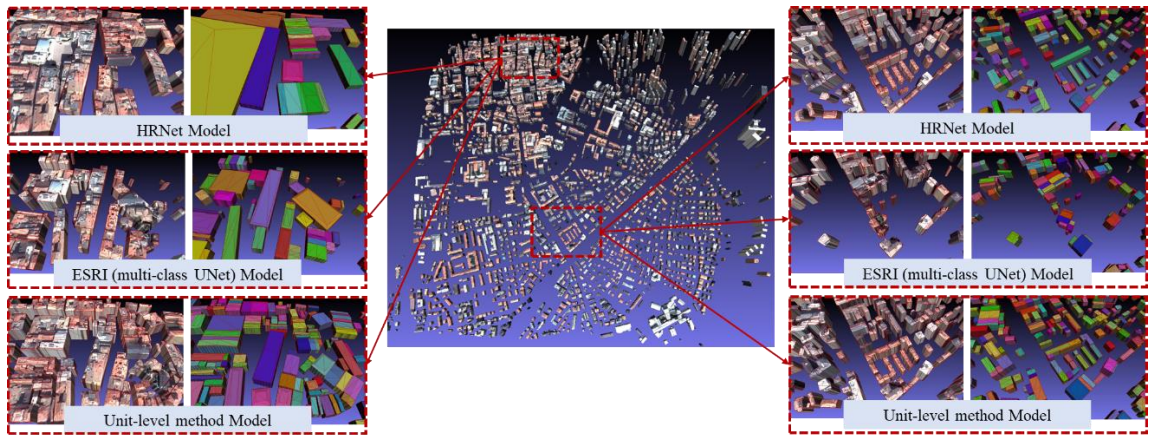

**Figure 3.8** Building 3D models in Trento region with both dense and sparse buildings

**3.5.3 Qualitative analysis of circular building reconstruction**

For circular building reconstruction part, since there are limited datasets including ground truth height information for circular buildings and most circular buildings are simple oil tanks, we only do qualitative analysis at this section.

The visual results for 3D models of circular-related buildings demonstrate successful generation capabilities. However, the current performance heavily relies on the accuracy of semantic segmentation. Future efforts will aim at enhancing the precision of building masks and developing post-processing techniques for these masks to improve key-point detection for circle identification, thereby refining the overall model generation process.

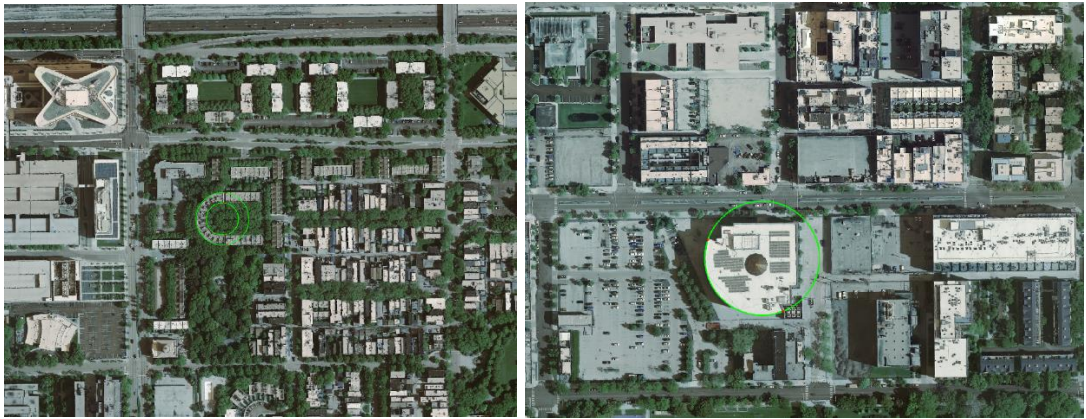

**Figure 3.9** 2D detection for circular building with complex structure, left: C-shape building, right: sector shape building

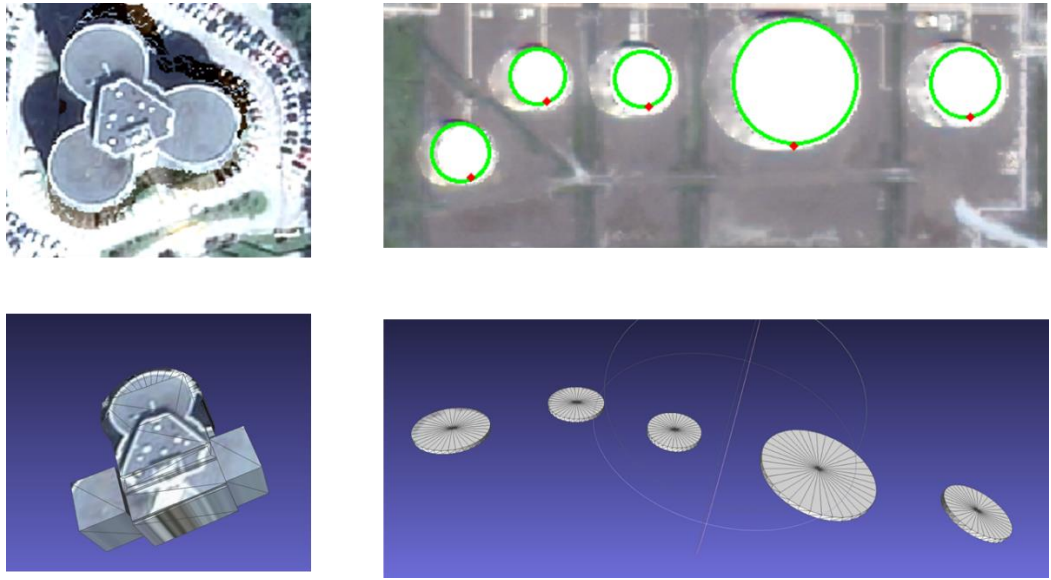

**Figure 3.10** 3D detection for circular building with complex structure, left: complex building but not well reconstructed, right: well reconstructed circular buildings

Moreover, we updated the circular reconstruction module to SAT2LoD2 workflow, and our open-source software is able to generate both rectangular and circular buildings.

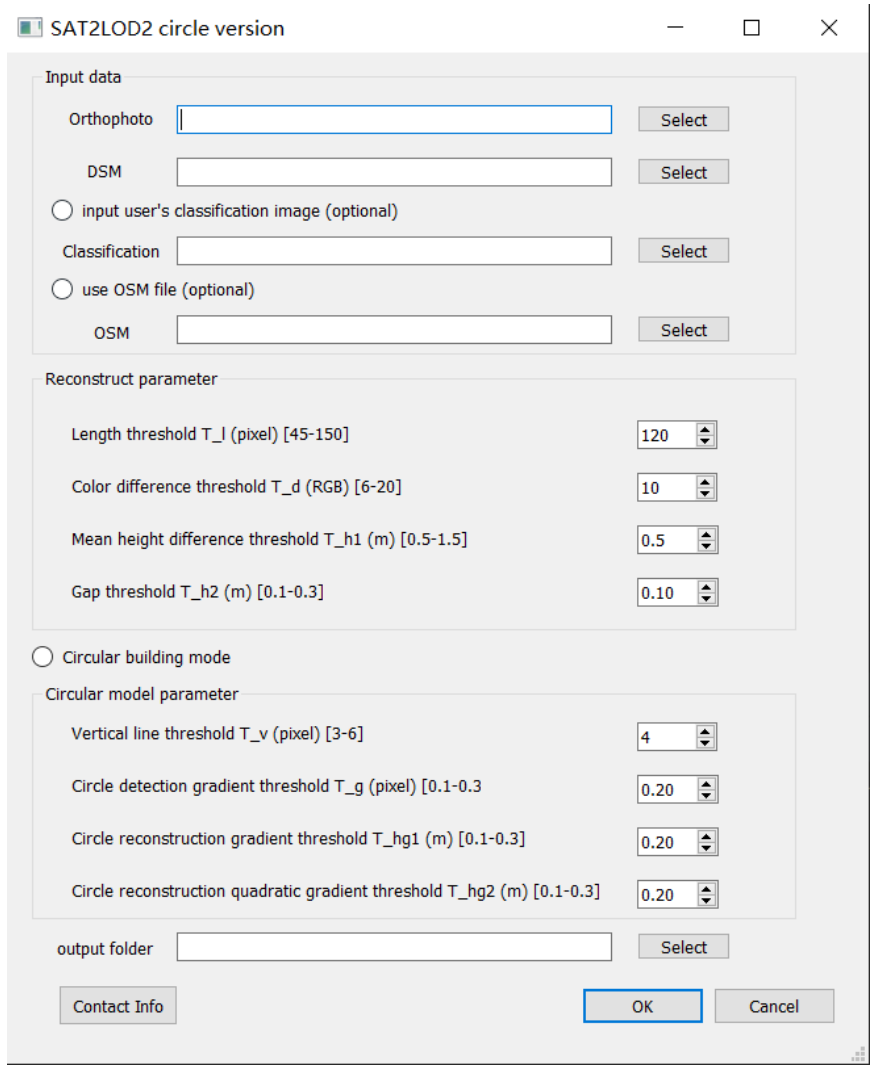

**Figure 3.11** Circular building version for SAT2LoD2

## 3.6 Discussion

The experimental results indicate that unit-level building reconstruction method obviously improves the granularity and accuracy of the building LoD2 model, particularly in densely populated urban environments. This approach significantly improves the delineation of building boundaries and the calculation of 3D roof primitives by segmenting complex and adjacent buildings into distinct sections. Such precise urban modeling is crucial for creating more accurate and reliable representations of building structures.

The enhanced detail is particularly advantageous for planning and analysis in urban development, notably in areas characterized by dense, irregularly shaped buildings with non-distinct textures, such as slums and poorly maintained neighborhoods—areas that previous reconstruction methods struggled to accurately model. The unit-level method has considerable capacity for simulating highly intricate and business-oriented structures. Proficiently analyzing and precisely depicting the complex formations of these edifices can significantly assist in diverse urban planning and architectural implementations. This approach can offer a more intricate comprehension of the urban environment, particularly in the case of commercial structures that frequently showcase distinctive and intricate architectural styles.

The reconstruction of circular buildings presents unique challenges due to their atypical shapes and the less distinct edge features in satellite imagery, complicating their inclusion in accurate 3D urban models. Despite these challenges, initial qualitative analyses have shown promising capabilities in 3D building model generation, mainly relying on the precision of semantic segmentation. To advance this field, future efforts must focus on enhancing the accuracy of building masks and developing sophisticated post-processing techniques that improve key-point detection for better circle identification. Such improvements are essential for capturing the architectural diversity of urban landscapes more effectively and can significantly contribute to urban planning and development.

**3.7 Conclusion**

This section introduces an effective level of details LoD2 building reconstruction approach at the unit-level, leveraging unit-level building segmentation results from satellite-derived Orthophoto and digital surface model (DSM) and a model-derive approach for 3D modeling. This method initiates with the segmentation to get unit-level building segments, followed by a polygon composition strategy designed to distinguish duplex or dense buildings as separate entities equipped with 3D primitives. Our technique effectively segments complex buildings and immediately adjacent buildings in densely populated urban areas with low-texture quality and detects circular parts from a complex building structure, subsequently reconstructing 3D building models utilizing a comprehensive library of predefined models. Besides, it adeptly detects and reconstructs circular parts from a complex building structure, facilitating the accurate reconstruction of circular buildings and complex buildings including circular structure. The empirical evaluation of experiments shows that our unit-level LoD2 modeling surpasses the construction result from publicly available building segmentation methods, especially in dense and complex urban environments.

# Chapter 4. Glacier Dynamics Tracking at 3D level from Time-series PlanetScope-derived Elevation Models and Climate Indicators

This chapter mainly introduces glacier monitoring by using satellite-derived 3D geospatial data, which has been presented as "Using PlanetScope-Derived Time-Series Elevation Models to Track Surging Glacier 3D Dynamics in Mid-Latitude Mountain Regions" on "AGU Fall Meeting 2023" by Shengxi Gui and Rongjun Qin, and this part of work is in preparation for a journal article may be submitted to "Remote Sensing of Environment".

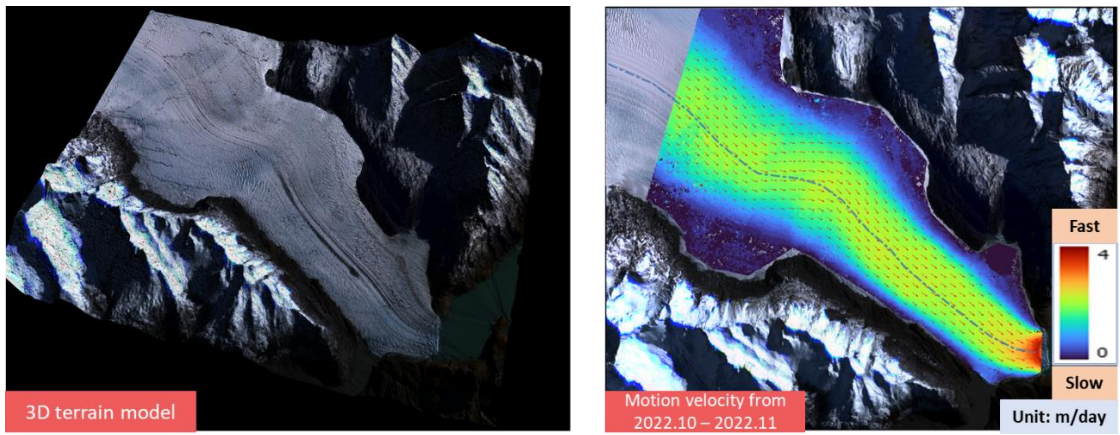

**Figure 4.1** 3D tracking on Viedma Glacier, left: 3D terrain model for Viedma Glacier at 2023.01, and right: glacier vertical motion velocity from 2022.10-2022.11

## 4.1 Chapter Abstract

Glacier retreat reflects how climate changes. Remote sensing images play a vital role in providing data points and monitoring them. Oftentimes the observed glacier retreats are sparse and binary (on and off), yet it is unclear that these binary observations are within seasonal variations. Thus, data with high-temporal resolution are necessary to 1): establish sufficiently dense observations to achieve enhanced conclusions; 2): derive higher-level data, such as motion velocity, to assess the level of global warming and climate change. PlanetScope satellite constellations consist of over 400 satellites and can provide global covered daily/weekly observation with a 4m Ground Sample Distance (GSD). Moreover, it was shown in our prior work that the data is sufficiently rich to establish photogrammetric 3D measurements at its coverage. In this work, we report a study on monitoring surge-type glaciers in middle-latitude mountain regions in America and Asia using the derived time-series 3D elevation models from PlanetScope. The study includes three sites: La Perouse Glacier (North America), Viedma Glacier (South America), and Skamri Glacier (Central Asia). Based on PlanetScope data, we derived near bi-monthly 3D elevation models for the year 2019-2023 using RPC stereo processor (RSP), to track the ice flow both in 2D and 3D. The results can be used to decorrelate the factors from seasonable variation: the Viedma Glacier is observed thinner through time with a slower flow rate, while the velocity of La Perouse Glacier is accelerating, while the two other sites did not show obvious glacier retreat other than those caused by seasonable variations.

## 4.2 Introduction

With climate change since human beings come to industrial society, glacier retreat is accelerating worldwide. Glaciers play a critical role in global environmental dynamics, especially as major contributors to sea-level rise in recent years (Kulp and Strauss 2019), which sensitivity deeply influencing hydrological cycles and ecosystems (Hugonnet et al. 2021; Huss and Hock 2018; Change and others 2014). Currently, the retreat of glaciers serves a complex role by temporarily mitigating water stress for populations dependent on ice reserves through increased river runoff, however, this benefit is temporary. Adding complexity to glacier retreatment, glacier fast motion and surging, exhibit exceptional flow rate of surface ice, further complicates predictions about water availability and glacier dynamics (Quincey et al. 2011). For example, the Jakobshavn Isbrae in Greenland boasts the distinction of having the world's fastest glacial flow, with speeds exceeding 40 meters per day (Joughin, Abdalati, and Fahnestock 2004). Such rapid advancement stands in sharp contrast to the typical glacial flow rates, which average just a few meters annually (Chandler et al. 2018). It is crucial for assessing the broader impacts on ecosystems and addressing cryosphere-related hazards, emphasizing the need for an integrated approach to accurately monitor and predict the glacier fast dynamics.

Remotely sensed data, especially satellite imagery, acts as a unique approach to collecting ground information in inaccessible regions like glaciers, mountains, and jungles. Therefore, satellite imagery is an effective source to monitor glacier dynamics for such a large-scale range. The PlanetScope satellite constellation has over 150 satellites in orbit, which creates an unmatched capability to collect almost daily images worldwide (Roy et al. 2021) with a Ground Sample Distance (GSD) of 4 m per pixel, which has a higher spatial resolution than most commonly used global-cover optical satellite imagery as Landsat-8 (30 m) and Sentinel-2 (10 m). The typical multi-spectral images with extensive coverage mainly provide 2D information. Moreover, for high-spatial-resolution images such as PlanetScope, 3D information such as Digital Surface Model (DSM) can be generated using multi-view stereo satellite images with appropriate intersection angles( Huang, Tang, and Qin 2022). **Figure 4.2** provides the current dynamics from the PlanetScope image in the past half a year (upper row), about 400 m at the red rectangle and 150 m at the blue rectangle, and the historical changes at the terminus part (yellow rectangle in **Figure 4.2**) in the past 15 years from Google Earth (second row). **Figure 4.3** shows both 2D and 3D change for Viedma Glacier, which indicates that there is a combination of fast moving and retreatment.

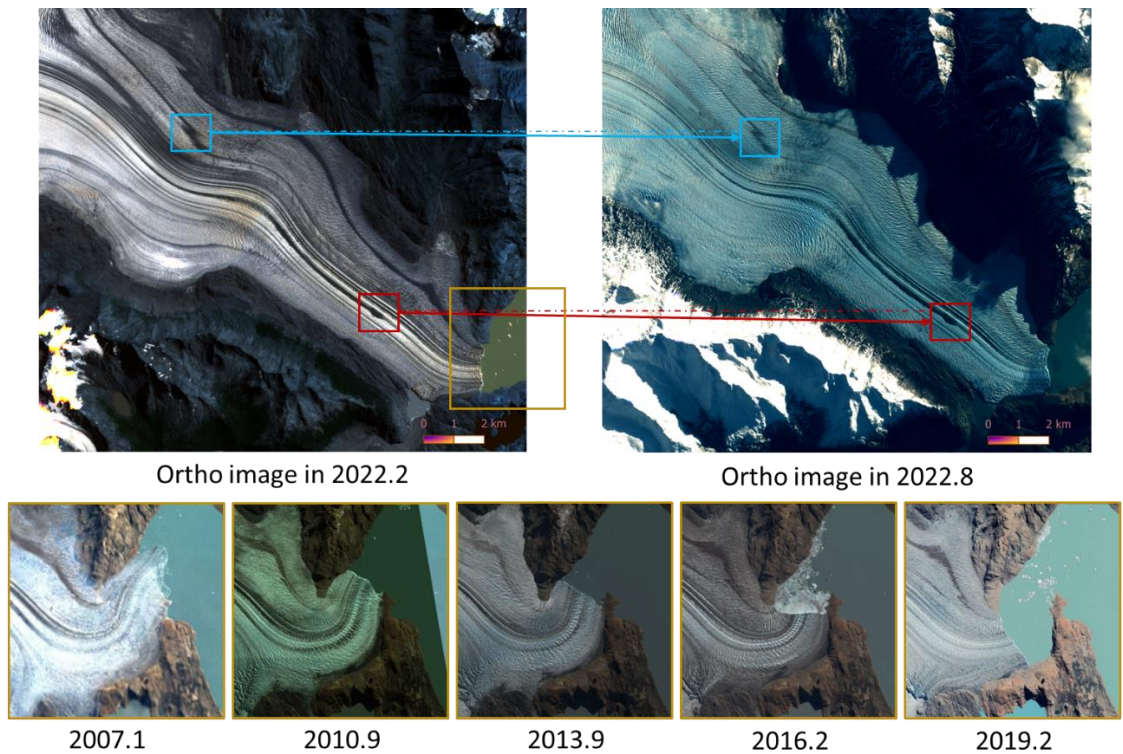

**Figure 4.2** Satellite view (Ortho image) of surging type glacier (Viedma Glacier), the first row uses Planet Scope Ortho images to show the glacier motion between 2022.02 and 2022.08. The red and blue rectangles in these two images represent the same area. The second row shows the series change at the terminus (yellow rectangle) from 2007 to 2019 with Google Earth images.

Quantitively evaluating the dynamics of large-scale glaciers has long been a significant topic in the global climate change field. Satellite data are the only available earth observation method for those inaccessible glaciers between mountains to monitor change comprehensively. However, for those fast-moving glaciers, there is still a lack of sufficient satellite imagery to analyze the mechanism of dynamics with climate for years. First, previous studies that used optical satellite imagery explored the dynamics model at 2D level (Derkacheva et al. 2020; Vecchio et al. 2018), and Synthetic Aperture Radar (SAR) based studies (J. Li et al. 2018) can generate 3D surface models while spectral information is not as superb as optical imagery. In addition, Lenzano et al. (2018) utilized a close-view camera to estimate flow in the terminus while the integrated motion was inaccessible. Second, numerous current research on the fast-flowing glacier areas studied only for a few certain periods (Dematteis et al. 2022; Euillades et al. 2016; Vecchio et al. 2018). It is still challenging to study dynamics with time series for years. Third, it is still a question of whether the volume change in the fast-moving glaciers is periodical (ice recovers annually) or the retreatment is permanent or has a fluctuation with superimposed seasonal change and constant retreat. Finally, for a specific glacier, the precise climate indicators that quantitatively control the glacier's motion rate over different time periods remain unclear (Lovell, Carr, and Stokes 2018), as does the manner in which these indicators influence the glacier's mode of movement.

The objective of this research is to quantitatively evaluate the motion in the several fast-moving glaciers in mid-latitude mountain regions worldwide (include South America, North America, and High Mountain Asia) at the 3D level, and analyze the seasonal and annual dynamic mode in this area, to explore the global warming level and local & global climate change in these regions. High spatial resolution (4 m) and high temporal resolution (daily) Planet Scope images from 2019 to 2023 are adapted to generate DSM and Ortho images bi-month in study areas. Furthermore, a novel glacier motion tracking approach is proposed by adapting multi-direction stereo matching from Orthophoto between two periods. Moreover, the horizontal motion and volume change with time are analyzed with local climate data and global ocean dynamics, to discover the principle of retreat in this area is periodical glacier change or permanent and the quantitative model for climate-driven glacier dynamics.

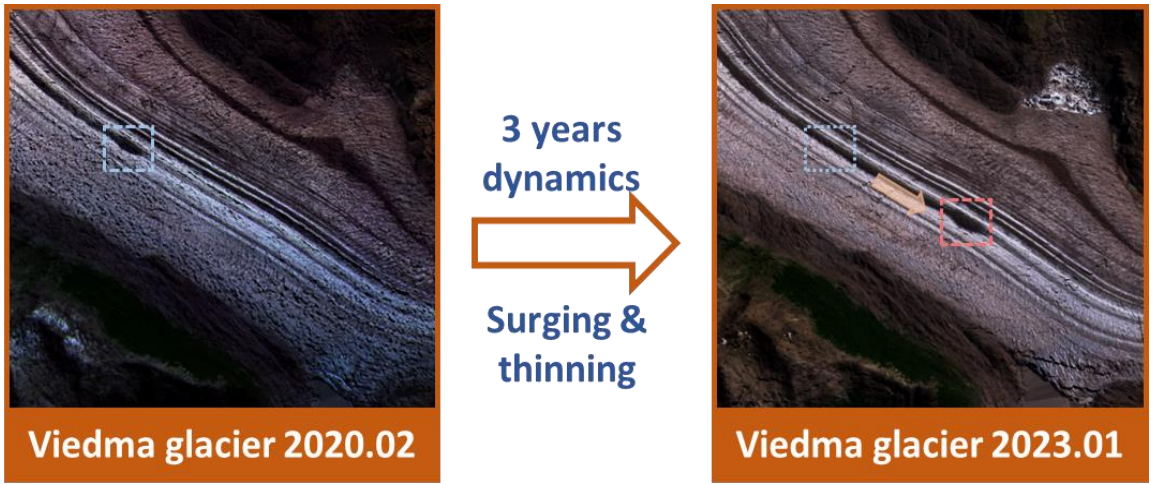

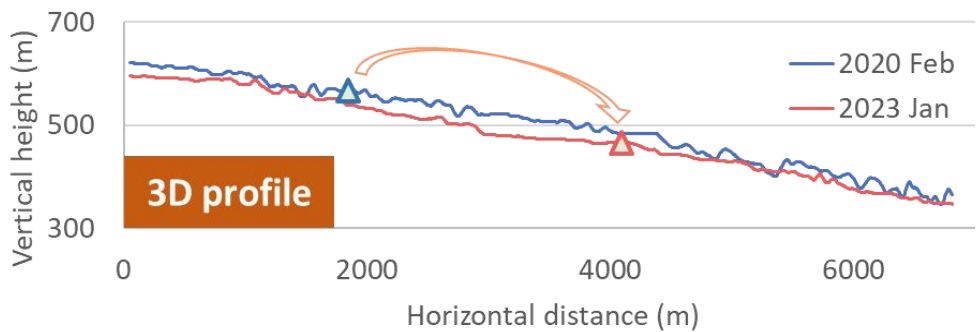

**Figure 4.3** Example of glacier 3D dynamics in Viedma glacier within 3 years at same season. There is both fast glacier motion and glacier retreatment.

## 4.3 Study areas and datasets

### 4.3.1 Study areas

In the study of glacial dynamics, attention often gravitates towards the polar regions, for which glaciers locate at Arctic and Antarctic, due to their extensive ice coverage. However, it is crucial to acknowledge that rapid glacial movements are not solely confined to these high-latitude areas. Mid-latitude mountain regions around the world are significant sources to several fast-dynamic glaciers, demonstrating substantial activity and variability (Hugonnet et al. 2021). Their existence highlights the varied conditions under which glaciers can evolve and the intricate relationship between climate and glacier dynamics beyond the polar regions. In our study, we selected three representative glaciers locates worldwide (**Figure 4.4**): Viedma Glacier (South America), La Perouse Glacier (North America), and Skamri Glacier (Asia).

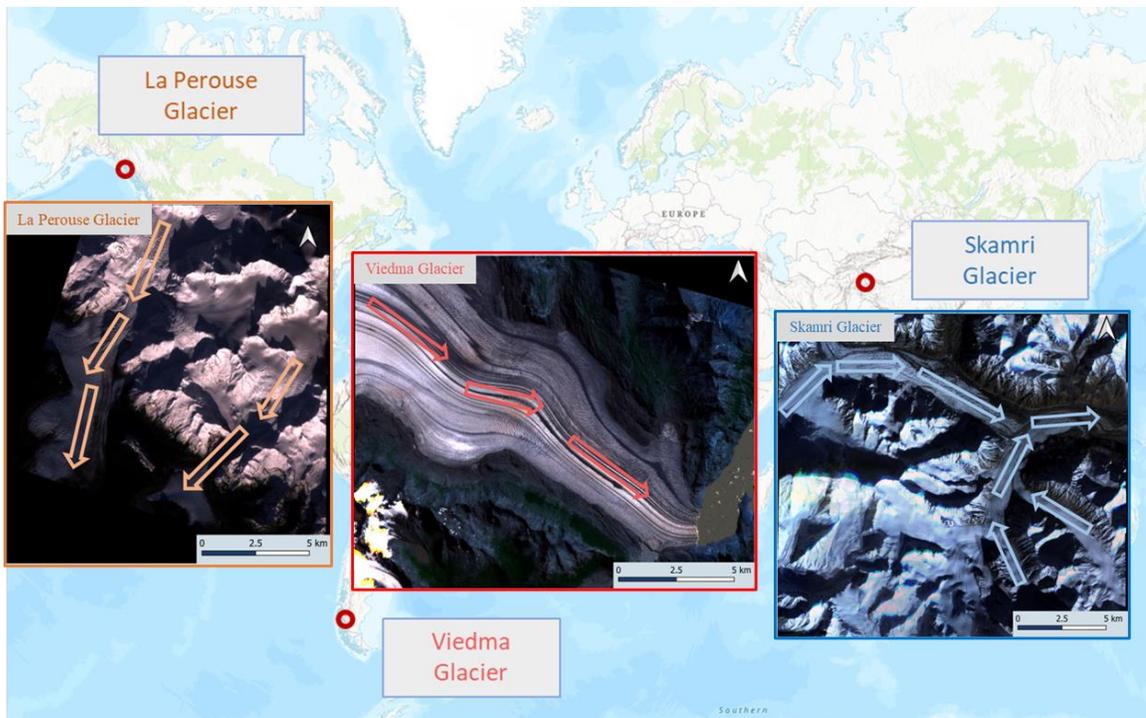

**Figure 4.4** Three study areas, include Viedma Glacier (South America), La Perouse Glacier (North America), and Skamri Glacier (High Mountain Asia)

As one of the largest glaciers outside polar regions, Southern Patagonian Ice Field (SPI), which is located in South America in the Austral Andes Mountain range, has been providing a significant retreat for several decades since 1944 (Aniya, 1995). The Viedma glacier (49°27′36″S, 73°11′42″W) is one of the fast-changing areas in SPI, and the ice flow field can reach a maximum surface velocity value of 3.5 m/d at the terminus (Lenzano et al., 2018) and 5.5 m/d at the middle basin (Vecchio et al., 2018) from 2014 to 2016. Since the dynamic of the Viedma glacier sustains decades, the motion speed of the glacier reflects the level of global warming and climate change, and both El Nino and La Nina in recent years will greatly affect the dynamic progress. In our study area in Viedma Glacier, we selected subregion with an area size around 17000 m by 14500 m, and height of ice surface from 200 m to 1000 m.

Similar to rapid retreat observed in the Viedma Glacier locates at South America, other mountain regions worldwide are experiencing dynamic changes in their glacial landscapes. In North America, the La Perouse Glacier (58°33′46″N, 137°04′59″W) presents a compelling case study for another region nearby the Pacific Ocean and highly impacted by El Nino and La Nina. Located within Glacier Bay area at Alaska, USA, this glacier exhibits signs of fast-flowing with a flowing velocity of 100–400 m/year (Van Wychen et al. 2018). Understanding the factors driving this acceleration in La Perouse Glacier is crucial for assessing the broader impacts of climate change on North American glacial systems. In our study area in La Perouse Glacier, we selected subregion with an area size around 18300 m by 17000 m, and height of ice surface from 0 m to 1500 m.

Similarly, in Central Asia, the Skamri Glacier (36°03′N, 76°15′E) locates within the Karakoram Mountain offers another case into the response of glaciers to a warming climate but is much less impacted by ocean climate system. Investigating the dynamics of Skamri Glacier can provide valuable insights into the response of Central Asian glaciers to climate change and inform regional projections of glacial retreat. In our study area in La Perouse Glacier, we selected subregion with an area size around 19800 m by 18300 m, and height of ice and debris surface from 4000 m to 5500 m.

**4.3.2 Datasets for remote sensing imagery and climate**

For our investigation across the three study areas, we acquired PlanetScope satellite imagery spanning from beginning of 2019 to August 2023, which includes imagery from Dove (4 bands), Dove R (4 bands), and Super Dove (8 bands) sensors. To facilitate the creation of 3D reconstructions of glacier surfaces in rapidly flowing regions, we meticulously selected PlanetScope images captured within a 10-day window for each period, ensuring the generation of dependable 3D geospatial data. This strategy enabled us to obtain monthly or bi-monthly 3D geospatial observations for the study areas. Over the span from 2019 to 2023, we have employed 30 periods based on 266 PlanetScope images for the Viedma Glacier region, 24 periods based on 267 PlanetScope images for the La Perouse Glacier region, and 34 periods based on 490 PlanetScope images for the Skamri Glacier region.

In addition to remotely sensed imagery, we incorporated both local and global climate data to explore the correlation between changes in glacier flow velocity and climate indicators. For each of the study areas, we gathered local climate data spanning from 2019 to 2023, which includes hourly measurements of 1) air temperature, 2) precipitation, and 3) evaporation, sourced from the ECMWF Reanalysis v5 (ERA5) climate datasets (Muñoz-Sabater et al. 2021). The climate data boasts a spatial resolution of 0.25 degrees in both longitude and latitude, with each glacier falling within a single pixel's coverage. To assess the impact of global ocean climate dynamics, we employed the Oceanic Niño Index (ONI) to investigate the relationship between global climate change and the glaciers significantly influenced by the Pacific Ocean (Viedma Glacier and La Perouse Glacier). Conversely, the North Atlantic Oscillation (NAO) was used to explore the connection between global climate variations and the Skamri Glacier, which is marginally affected by the Atlantic Ocean.

**4.4 Methodology**

Our methodology commences with a series of time-lapse PlanetScope images, leveraging an intricate image processing pipeline to analyze glacier dynamics in three dimensions. Furthermore, it integrates climate indicators to discern the primary factors influencing changes in glacier flow velocity. Additionally, for each study area, the processing is conducted independently, enabling the generation of detailed glacier dynamics and the analysis of time-series results. As shown in **Figure 4.5**, the proposed workflow initiates with 3D reconstruction (RSP) from stereo PlanetScope imagery for each period. Next a registration of DSM and Orthophoto are applied to reduce the misalignment of different periods. Combining adjacent periods of Orthophoto, motion between these two periods has been tracked by using multi-direction stereo matching approach. Finally, both glacier 3D dynamics and climate impacts are analyzed to quantitatively describe the glacier flowing.

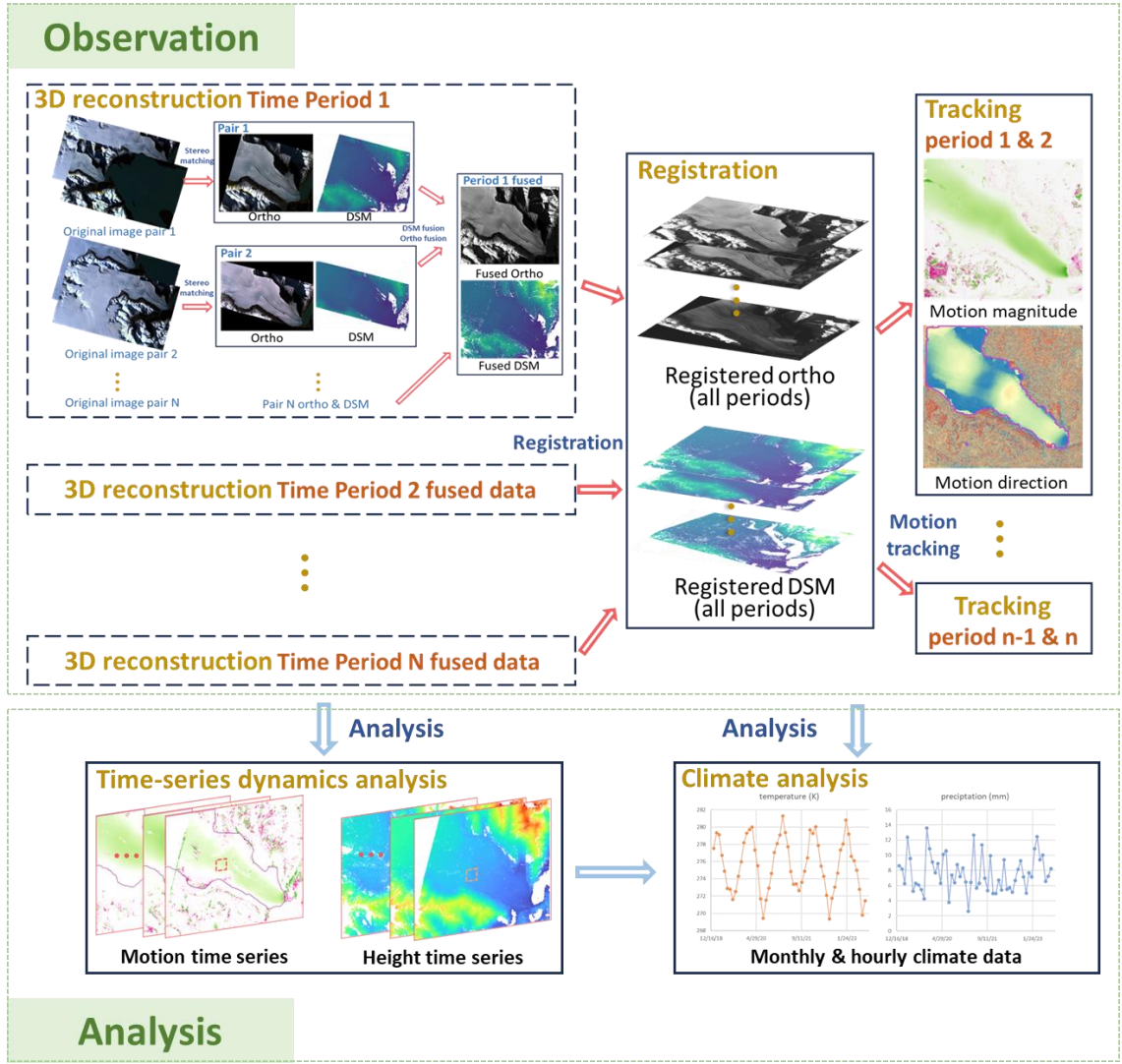

**Figure 4.5** The workflow of our processing & analysis for glacier dynamics

**4.4.1 3D glacier surface reconstruction**

Dense stereo matching is a typical multi-view optical 3D recovery solution. The multi-view image 3D reconstruction processing begins with feature detection to record the key points of each image (Lowe 2004b). Next, Structure-from-Motion (SfM) performs aerial triangulation and sparse point cloud generation. Then dense matching step helps reconstruct the dense point cloud and DSM. RPC stereo processor (RSP) provides sequentially a level 2 rectification, geo-referencing, point cloud generation, DSM resampling, and orthophoto rectification from satellite stereo images (Rongjun Qin 2016b). Huang, Tang, and Qin (2022) indicated that PlanetScope images have the capability of reconstructing DSM for the ice-covered region, and **Figure 4.6** shows a sample result of DSM generation in the textureless glacier for different periods. Specifically, for each period, we adapted PlanetScope images captured within a 10-day window for each period, ensuring the generation of dependable 3D geospatial data. Moreover, most of periods have more than one pair of reconstructed DSM and Orthophoto, a median filter is applied to fuse and stitch multiple pairs to generate geospatial data cover whole study area (Rongjun Qin 2016b). Following the 3D reconstruction process, a DSM and an Orthophoto for each period are produced, accurately depicting the 3D context for that specific timeframe. These outputs serve as foundational data sources for subsequent processing steps, providing a detailed and comprehensive representation of the glacier spatial condition during each period.

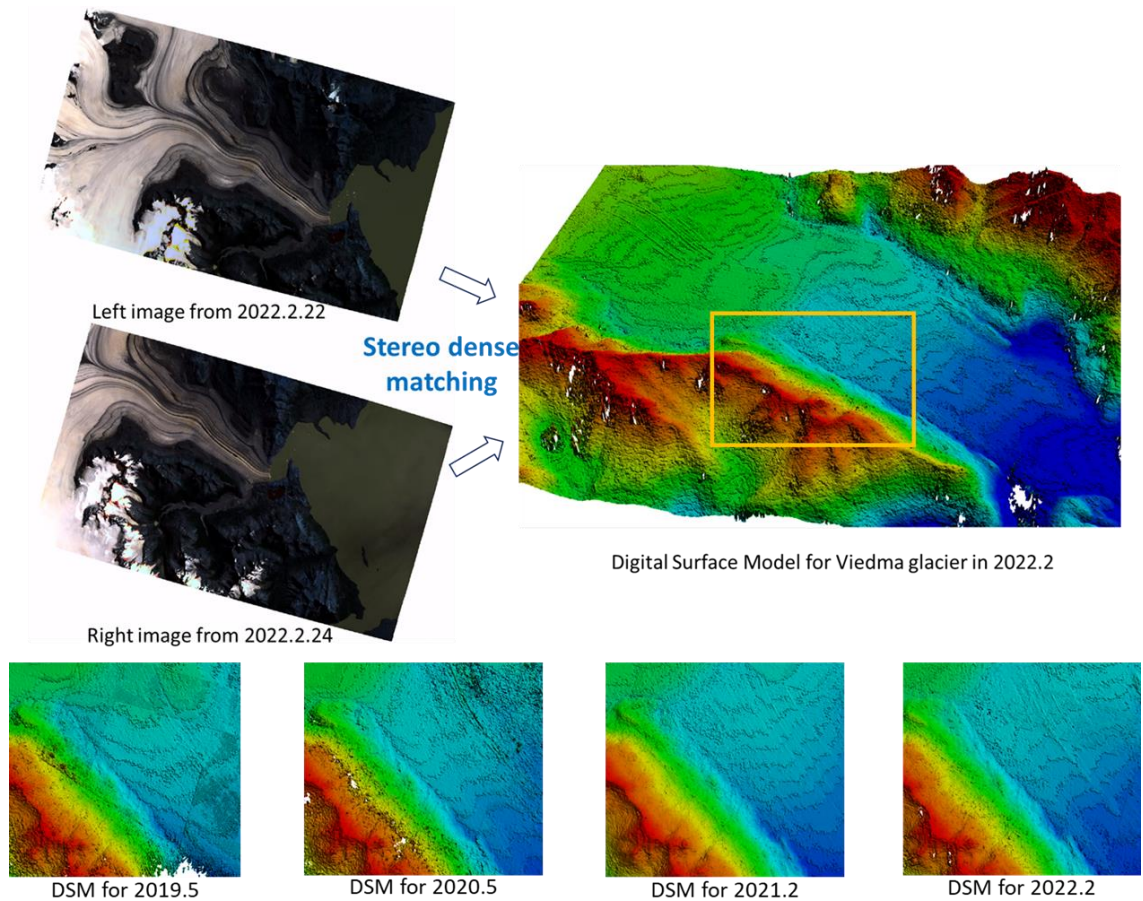

**Figure 4.6** Stereo matching process and generated DSM, a sample for the Viedma area, the first row shows the principle of stereo dense matching to generate DSM from stereo images, and the second row shows the DSM from 2019 to 2022 (yellow area)

**4.4.2 DSM and Orthophoto registration with Iterative Closest Point (ICP)**

DSM registration deals with the inconsistent issue for DSM in different periods, since the precision of RPC parameters and dense matching. Feature-based methods establish registration between 2D imagery and 3D point clouds through feature correspondences (Parmehr et al. 2014), to match several DSMs to have a minimum difference in X, Y, and Z directions with the help of Ground Control Points (GCP). In this study, due to the challenging nature of placing GCPs around rapidly flowing glaciers or in the adjacent mountainous terrain, we have adapted to utilize the features of DSM and Orthophotos. This approach allows us to conduct image-based relative registration, aligning geospatial data across different periods effectively. This method ensures accurate alignment of data sets without the need for physical GCPs, which is particularly advantageous in inaccessible glacier and mountain regions.

There is a two steps pipeline for DSM & Orthophoto misalignment removal for different periods: 1) registration for DSMs to get transformation (rotation matrices), and 2) apply computed transformation for both DSM and Orthophoto. The registration of DSM is specifically focused on aligning the unchanging regions of each glacier, indicating that the process is based primarily on the surrounding mountainous terrain rather than the ice-covered areas of the glacier (mask out glacier during registration). The foundational method employed for DSM registration is the Iterative Closest Point (ICP) algorithm (Besl and McKay in 1992). This technique iteratively reduces the distance between two sets of 3D data points, effectively ensuring accurate alignment of the stable, non-glacial features within the landscape. For glacier registration for different periods of time adapt an ICP-based pair-wise registration method (Xu, Qin, and Song 2023; Xu et al. 2021), to reduce the geometrical distortion and height misalignment. We selected one period as target DSM for each glacier, and others as source DSMs. The registration process involves selecting corresponding points between the source DSM and target DSM, estimating the transformation (rotation matrix) including rotation and translation needed to align these points, applying the transformation to the source dataset, and repeating this process until the alignment converges to a satisfactory level of accuracy. Following the registration of the DSM, a rotation matrix is generated for each period, which is subsequently applied to the entire region, encompassing both the DSM and orthophotos for each respective period. This procedure facilitates the alignment of glacier surfaces from different periods, accommodating even in the face of rapid ice volume retreat or swift ice surface flow. This ensures consistent and accurate comparison of glacier changes over time, regardless of the dynamic nature of glacial movements.

**4.4.3 Glacier 2D motion tracking based on multi-directional stereo matching**

From aligned DSM and Orthophoto, image feature track aims to quantitatively evaluate the ice surface motion by using optical satellite images. Spectral information, as well as the relative height information, distributes as a unique pattern, and by tracking with image matching, the ground motion can also be calculated (Baird, Bristow, and Vermeesch 2019; Dematteis et al. 2022). We utilized these spectral features on glacier surface, proposed a novel tracking approach based on multi-directional stereo matching, to track the 2D ice flow between two neighbor periods (time difference is usually 1~4 months). The workflow of our glacier tracking methodology is depicted in **Figure 4.7**. This process begins with two Orthophotos from distinct periods, followed by the generation of a series of rotated images for each, at intervals of either 3° or 5°. Subsequently, pairs of images with the same rotation angle are employed for dense stereo matching. Next an initial map is created for velocity and direction of glacier movement, which is based on minimum matching energy.

The final stage involves refining this map to eliminate outliers, ensuring the accuracy and reliability of the velocity and direction data captured for the glacier's dynamics.

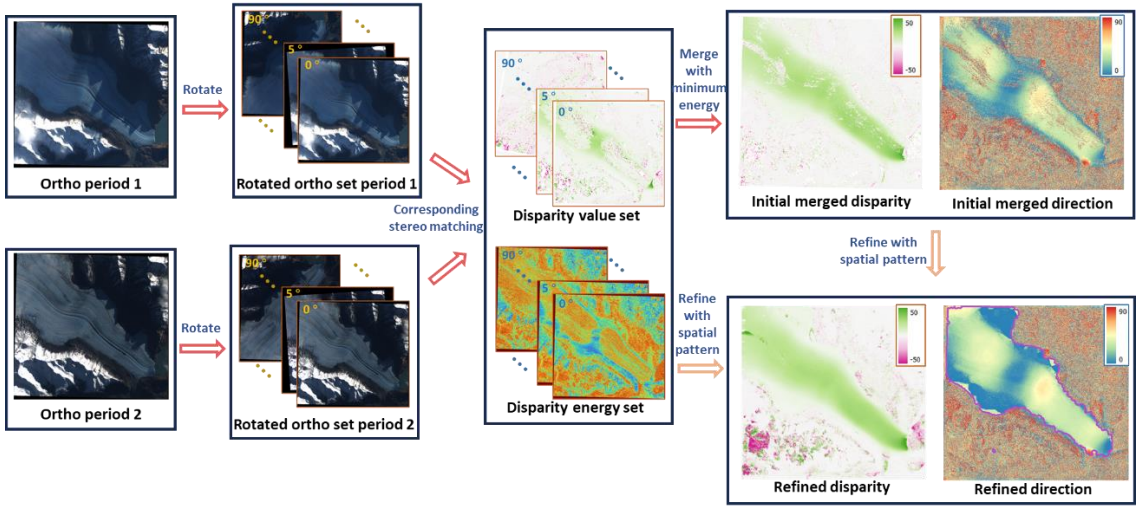

**Figure 4.7** The pipeline of 2D motion tracking based on multi-directional stereo matching, input data are two Orthophotos from different periods, output is a motion velocity map and a motion direction map.

Given the pixel size of the image and the time interval between two periods, the computation of ice flow velocity is converted to image matching, which identifies the displacement of the same object across two images (Dematteis et al. 2022). Furthermore, we further simplify 2D image matching to 1D stereo matching with multiple directions. We adapt Semi-Global Matching (SGM) as 1D stereo matching method to track the glacier motion. SGM is a popular method for dense stereo matching that is extensively employed in photogrammetry and computer vision. It is utilized to generate dense disparity maps from stereo images, facilitating the reconstruction of 3D models from 2D images. The fundamental principle of SGM is to combine matching costs not only from the local vicinity of each pixel, but also along several paths covering different directions across the entire image. The key formula underlying SGM involves the minimization of a cost function that combines both data and smoothness constraints:

$$E(D) = \sum_{p} C(p, D_p) + \sum_{q \in N_p} P_1 T[|D_p - D_q| = 1] + \sum_{q \in N_p} P_2 T[|D_p - D_q| > 1] \quad (4.1)$$

where, $q$ is a pixel in the image patch centered at pixel $p$. The pixelwise cost and the smoothness constraints are expressed by defining the energy $E(D)$ that depends on the disparity image $D$. The problem of stereo matching can now be formulated as finding the disparity image $D$ that minimizes the energy $E(D)$. Unfortunately, such a global minimization (2D) is NP-complete for many discontinuity preserving energies. In contrast, the minimization along individual image rows (1D) can be performed efficiently in polynomial time using Dynamic Programming. Thus, a new idea of equalizing the matching cost from each one-dimensional direction. By aggregating (smoothing) the cost $C(\mathbf{p}, d)$ for the pixel $\mathbf{p}$ and the disparity $d$, by calculating the sum of all the one-dimensional minimum cost paths, the disparity d of the pixel p is finally obtained.

After generating disparity maps for multiple directions, these 1D disparity maps are merged into an initial 2D motion set including a merged disparity map and a direction map, by adapting the direction with minimum matching energy for each pixel:

$$D_{initial}(p) = D_{di}(p), with\ di = \arg\min\{j|E_{di}(p), j \in \{1,2,\dots,n\}\} \quad (4.2)$$

where $D_{initial}(p)$ is the initially merged disparity value for pixel $p$, and $di$ is the index of direction $i$ with minimum matching energy computed by SGM.

Since there remains noise in matching result, a refinement step is adapted to further post-process the initial 2D motion maps. The proposed solution for refining the disparity in multi-directional merging involves utilizing a direction histogram for each point. This approach draws inspiration from the computation of the main direction in the SIFT descriptor algorithm. Here, the window size (σ) is determined by the difference between a point's direction and the average direction of the surrounding 50 by 50 pixels.

Weights for pixels in this localized window are assigned using a Gaussian distribution and confidence level based on similarity of nearby disparity, with the predominant direction for each point being identified as that with the maximum weighted value. The confidence level of pixel $p$ is computed as:

$$Confidence(p) = \begin{cases} 4, & if \ |dir(p) - \overline{dir(p)}| < 6° \\ 3, & if \ 6° \leq |dir(p) - \overline{dir(p)}| < 12° \\ 2, & if \ 12° \leq |dir(p) - \overline{dir(p)}| \leq 18° \\ 1, & if \ |dir(p) - \overline{dir(p)}| > 18° \end{cases} \quad (4.3)$$

where $dir(p)$ is the initial direction of pixel $p$, $\overline{dir(p)}$ is the mean direction value around 50 by 50 pixel around pixel $p$. For each pixel, there should be a searching window size base on its difference from the nearby pixels. While, to accelerate the processing speed, I use the convolutional function of four dispersed window-size kernels to compute the nearby energy, and then use the difference of nearby direction to set a confidence value for each pixel. The confidence level help to assign the processing nearby window size and the weight to refine nearby pixel:

$$Window\_size(p) = \begin{cases} 21, & if\ Confidence(p) = 4 \\ 51, & if\ Confidence(p) = 3 \\ 101, & if\ Confidence(p) = 2 \\ 201, & if\ Confidence(p) = 1 \end{cases} \quad (4.4)$$

$$G(x,y) = \begin{cases} \dfrac{1}{2\pi\sigma^2} e^{-(x^2+y^2)/2\sigma^2}, & if\ Confidence(p) = 2, 3, 4 \\ \dfrac{1}{Confidence(x,y)+4} \times \dfrac{1}{(Window\_size(p))^2}, & if\ Confidence(p) = 1 \end{cases} \quad (4.5)$$

where $Window\_size(p)$ is the window side selected to refine the direction for pixel $p$, if the confidence level is low, it prefers to use the direction statistics within a large region to refine this pixel. $G(x,y)$ is the gaussian weight with the mean direction around pixel (x, y) with an initial window 50 by 50 to differentiate single direction value in pixel $p$, in order to search for the most reliable direction, the convolutional function is also adapted, and the kernel function is based on the confidence value, for confidence value from 2 to 4, we adapt Gaussian kernel, and for confidence value for 1, we use a mean function to deal with probably incorrect direction. Next, give a loose condition function to update the direction in next iteration. To mitigate issues such as voids or inaccurately extensive areas, the method incorporates several iterations of correction. While challenges remain within large gaps or super noisy regions, these are significantly mitigated through the iterative process.

### 4.4.4 Time series 3D dynamic estimation

We utilize DSM from different periods to study the time-series 3D dynamics of glaciers, enabling a detailed analysis of their morphological changes over time. Seasonal variations significantly impact glacier height, with changes occurring due to factors such as snow accumulation in winter and melting in summer. These cyclical effects can mask or exaggerate the signals of long-term glacial retreat caused by climate change. To accurately distinguish between temporary seasonal fluctuations and irreversible glacier retreat, we employ a season change function. This mathematical model is designed to account for the seasonal cycle's impact on glacier height, allowing us to isolate and analyze the underlying trends of permanent glacial retreat. By applying this function, we can more reliably differentiate between short-term seasonal changes and the long-term, potentially permanent retreat of glaciers, providing a clearer understanding of glacier dynamics in the context of global warming. An advance mathematical function with time is used to represent the height dynamics of studied glaciers, which can distinguish seasonal change of glacier height and the long-term retreatment or accumulation. The function is the combination of sine function and linear function (Q. Wang and Sun 2022):

$$H(t) = a_0 \sin(a_1 t + a_2) + a_3 t + a_4 \tag{4.6}$$

Where $H(t)$ is the mean glacier height in a selected small area at time $t$, and $a_0 \sim a_4$ are coefficients that represent the combination of linear function and sine function, with $a_1 = 2\pi/365$ assigned as a constant to represent the phase equal to one year. The coefficients $a_0 \sim a_4$ are fitted by using the time series data of registered DSM.

### 4.4.5 Climate factors impact for surging velocity

In exploring the influence of climate on glacier surging velocity, we have analyzed seven monthly-based climate factors derived from ERA5 data for each study region. These factors include: 1) air temperature, 2) precipitation, 3) global climate phenomena including ENSO (El Niño Southern Oscillation) and NAO (North Atlantic Oscillation), 4) evaporation, 5) the number of days with maximum air temperature exceeding 0°C, 6) the number of days with mean air temperature above 0°C, and 7) the accumulated melting heat. The last three indicators from above are computed based on the hourly air temperature, and them mainly consider the melting effect.

To understand the relationship between these climate variables and glacier surging velocity, we have applied a multivariate linear regression model to the selected areas. Furthermore, we consider the possibility of a lag-effect (Ting Zhang et al. 2023) suggesting that the impact of climate factors on glacier velocity might not be immediate, but rather delayed, acknowledging the complex interaction between climate dynamics and glacial movements.

**4.5 Result and discussion**

**4.5.1 Time-series dynamics for three regions**

For all three regions, we compute the 3D terrain models for each period, and compute the motion between each nearby period. **Figure 4.8~4.10** show the time-series glacier dynamics result for three study areas.

In the Viedma region, characterized by its rapid glacial movement, the velocity map depicted in **Figure 4.8 (e)** reveals that during the summer season, velocities reach approximately 2m/day at the glacier's upper parts and about 4m/day near the terminus, with the fast movement around middle profile. The velocity profile shown in **Figure 4.8 (d)** aligns with seasonal variations, indicating a notable acceleration in the summer. Additionally, **Figures 4.8 (f)** and **(g)** highlight a significant retreat of the ice surface for the Viedma Glacier, observed after accounting for seasonal effects, further emphasizing the obviously volume loss with a changing climate.

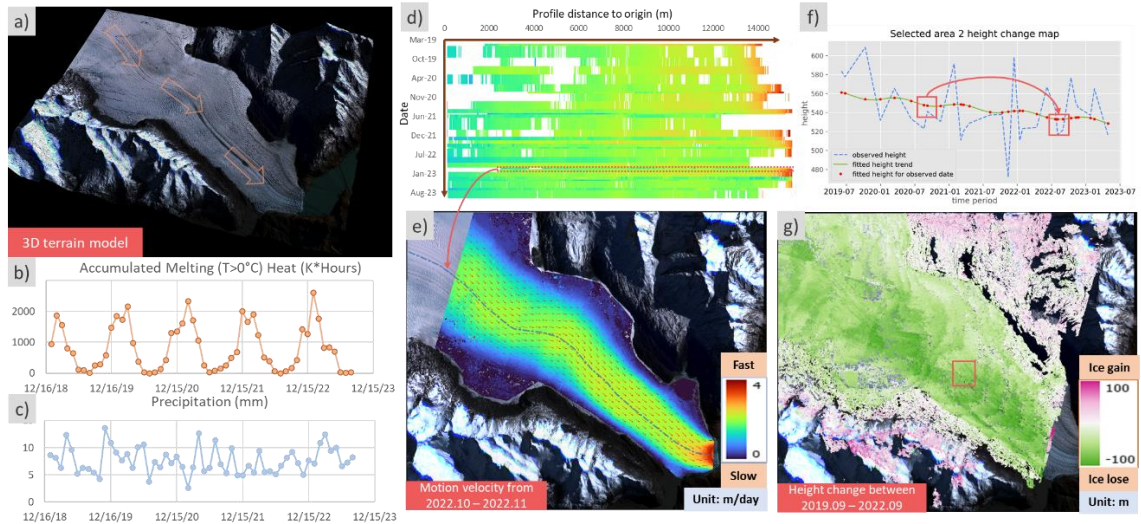

**Figure 4.8** Time-series glacier dynamics for **Viedma Glacier**. For each panel: **a)** reconstructed 3D terrain model by using PlanetScope imagery and ice flow main direction, **b)** accumulated melt heat map, **c)** precipitation map from 2019 to 2023, **d)** ice flow velocity profile for dash blue line at panel (e), **e)** ice flow velocity map with direction for period 2022.10-2022.11, **f)** selected area (red window in panel (g)) height change with time, **g)** height change map between two periods from 2019.09-2022.09

The La Perouse Glacier exhibits substantial flow, with a discernible spatial pattern in velocity that varies significantly across different parts of the glacier. According to the velocity map in **Figure 4.9 (e)**, during the summer season, the glacier's upper sections experience velocities of approximately 5m/day, whereas areas near the terminus, closer to the sea, exhibit lower velocities around 1m/day. This gradient in velocity from the upper parts to the terminus is illustrated in the velocity profile shown in **Figure 4.9 (d)**, which corroborates with seasonal changes, showcasing an acceleration during the warmer months. Furthermore, **Figures 4.9 (f)** and **(g)** reveal a distinct dynamic at play; there is evident accumulation in the glacier's upper regions while melting is more pronounced around the terminus. This pattern underscores the glacier's response to seasonal temperature variations, highlighting the interplay between accumulation and melting processes.

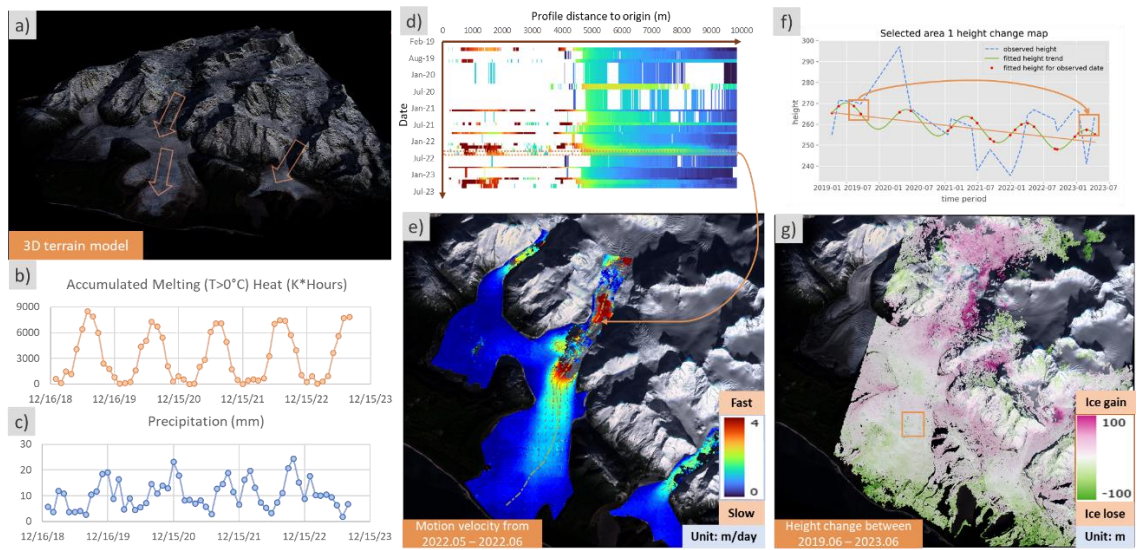

**Figure 4.9** Time-series glacier dynamics for **La Perouse Glacier**. For each panel: **a)** reconstructed 3D terrain model by using PlanetScope imagery and ice flow main direction, **b)** accumulated melt heat map from 2019 to 2023, **c)** precipitation map from 2019 to 2023, **d)** ice flow velocity profile for dash yellow line at panel (e), **e)** ice flow velocity map with direction for period 2022.05-2022.06, **f)** selected area (orange window in panel (g)) height change with time, **g)** height change map between two periods from 2019.06-2023.06

In the Skamri Glacier region, the flow velocity during summer is notably slower compared to the other two regions, accompanied by colder weather and minimal precipitation. The velocity map in **Figure 4.10 (e)** indicates that velocities peak at approximately 1m/day in the glacier's upper sections and decrease to about 0.5m/day near the merged glacier. Interestingly, the velocity profile presented in **Figure 4.10 (d)** does not exhibit a distinct seasonal trend, suggesting that Skamri Glacier's movements may be less influenced by seasonal temperature variations. Moreover, **Figures 4.10 (f)** and **(g)** detail the spatial changes in ice volume; the glacier's left portion is undergoing retreat, while the right side shows material accumulation. This differential behavior highlights the complex dynamics governing ice flow and volume changes within the Skamri Glacier region.

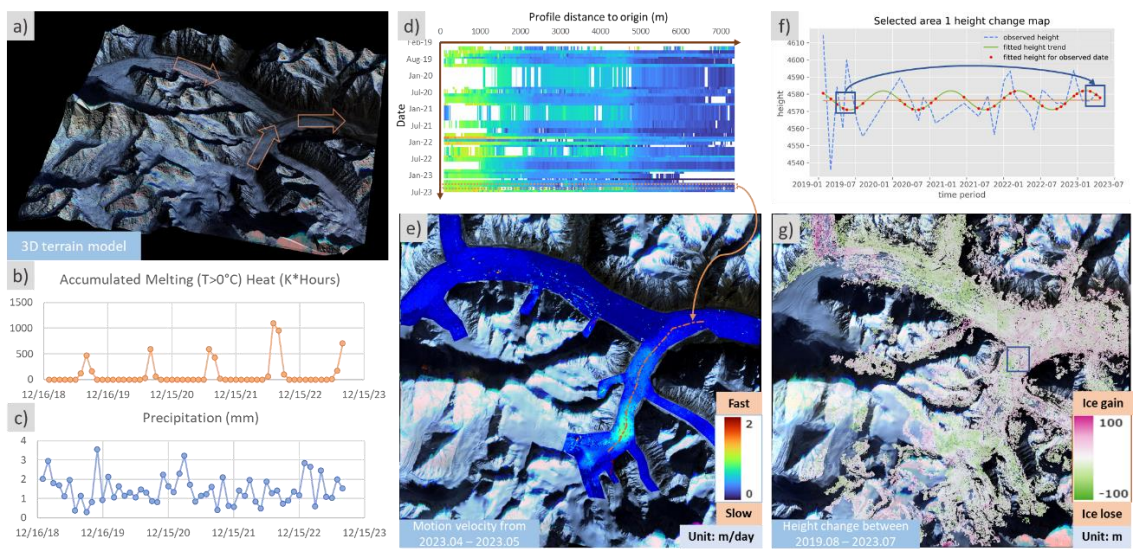

**Figure 4.10** Time-series glacier dynamics for **Skamri Glacier**. For each panel: **a)** reconstructed 3D terrain model by using PlanetScope imagery and ice flow main direction, **b)** accumulated melt heat map from 2019 to 2023, **c)** precipitation map from 2019 to 2023, **d)** ice flow velocity profile for dash yellow line at panel (e), **e)** ice flow velocity map with direction for period 2023.04-2023.05, **f)** selected area (blue window in panel (g)) height change with time, **g)** height change map between two periods from 2019.08-2023.07

**4.5.2 Climate-driven effect to glacier surging**

To examine the climate-driven effects on glacier surging, we employed a multivariate linear regression model to analyze the relationship between the surging velocity in selected areas and seven key climate factors. Additionally, we considered the possibility of a lag-effect, suggesting that the impact of these climate factors on glacier velocity might not be immediate. Analysis of **Figure 4.11** indicates that both the Viedma and Skamri Glaciers exhibit an optimal regression fit with a lag-time of approximately 40-45 days, whereas the La Perouse Glacier demonstrates its best fit with no lag-time. This variation in lag-time among the glaciers can be attributed to the differences in climate and geographical locations affecting each glacier. Specifically, the La Perouse Glacier, characterized by significantly higher precipitation levels compared to the other two regions, may be a hydrologically controlled surges (Miles et al. 2020) region. In this case, increases in flow and sliding velocities are directly linked to changes in the efficiency and water pressure of the subglacial drainage system. In contrast, the Viedma Glacier's proximity to the sea and its susceptibility to global climate influences, such as ENSO, differentiate its response to climate factors compared to the Skamri Glacier. This analysis indicates the complexity of glacier dynamics and the importance of considering both local and global climate variables in analyzing glacier flowing.

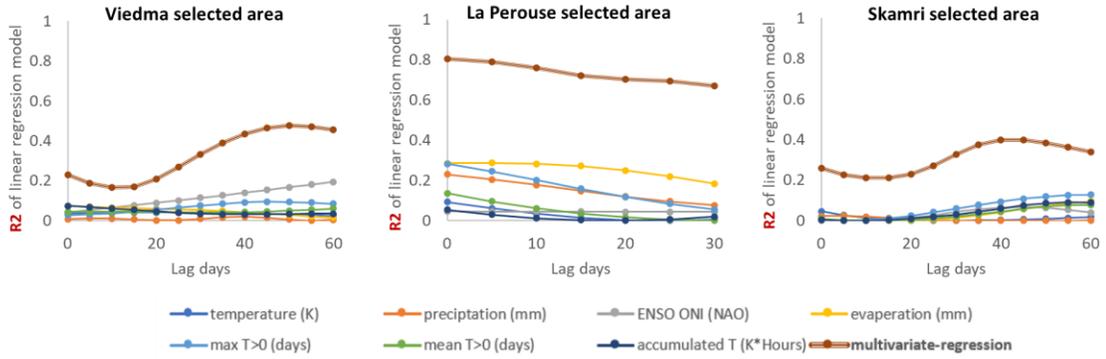

**Figure 4.11** Quantitatively analysis of lag effect of climate factors for each region, this is a plot between R2 of regression model and lag days, the highest R2 represents best lag days of one region

**4.5.3 Discussion of motion tracking method**

Our glacier tracking method offers a distinct advantage over traditional correlation-based approaches (Leprince et al. 2007; Lei, Gardner, and Agram 2021; Baird, Bristow, and Vermeesch 2019), which typically rely on the correlation of search windows between a reference image and a search image across two dimensions, X and Y, within a search space of $[-d, d] \times [-d, d]$, where $d$ represents the search range. In contrast, our approach is rooted in stereo matching, essentially a form of image correlation, but with a crucial innovation: by incorporating image rotation, we effectively transform the search space into a 1.5-D space, represented as $n_{dir} \times [-d, d]$, with $n_{dir}$ is the number of directions. This adaptation significantly narrows the search space compared to conventional methods. Ideally, we would simultaneously optimize direction and disparity using the Semi-Global Matching (SGM) method, which includes a smoothness term for enhanced precision. However, due to the substantial memory requirements of this approach, which may exceed the capabilities of our computing resources, we instead utilize a limited disparity searching range. These maps allow us to effectively optimize both direction and disparity, offering a practical and efficient solution that maintains the robustness and accuracy of our glacier tracking method.

## 4.6 Conclusion

Through the development of a comprehensive photogrammetry pipeline of 3D reconstruction and flow tracking, we have successfully monitored the 3D dynamics of glaciers across different study regions, revealing distinct trends and responses to climate variables. Specifically, the Viedma Glacier demonstrated a consistent volume retreat trend after 2019, decreasing by 9.7m/year, with its velocity most affected by climate conditions after a lag of 45-50 days. Conversely, the La Perouse Glacier exhibited a rapid volume retreat from 2019 to 2021 at a rate of 11.2m/year, followed by a period of accumulation at 8.4m/year post-2022, with climate factors influencing surging velocities almost instantaneously. The Skamri Glacier showed a more stable profile, with no significant volume retreat observed during the study period, and climate impacts on velocity manifesting after a 40-45 day delay. These results emphasize the varied behaviors of glaciers in reaction to changes in the environment, emphasizing the complex connection between glacier dynamics and climate. Such insights are essential for forecasting future glacier changes and developing ways to reduce the effects of global warming on these important natural environments.

# Chapter 5. Enhanced Remote Sensing of Surface Water Chlorophyl-a: Coupling Dynamic Algae Vertical Movement Modeling with Multispectral PlanetScope Satellite Images

This chapter mainly introduces the algae bloom monitoring by using PlanetScope imagery for lakes in Ohio, which started from 2023 Fall and collaborate with Dr. Rongjun Qin, Dr. Linda Weavers, Kaiden Murphy, and Mark Tischer, and this part of work is in preparation for a journal article may be submitted to "Remote Sensing of Environment".

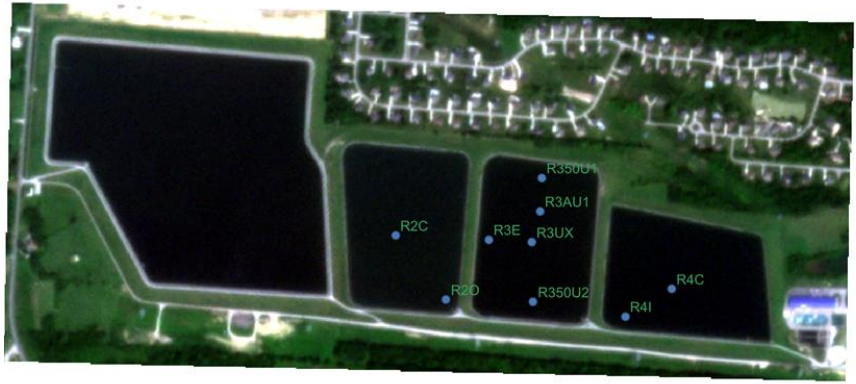

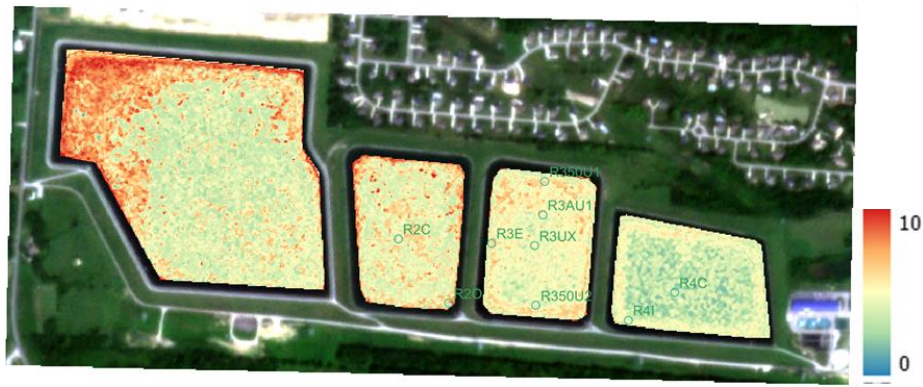

**Figure 5.1** Graphical abstract of harmful algal blooms monitoring by using PlanetScope, top: PlanetScope image (RGB) for Del-Co Reservoir, bottom: predicted Chl-a map

## 5.1 Chapter Abstract

In an era where precision in environmental monitoring is paramount, accurately quantifying chlorophyll-a (Chl-a) concentrations from satellite imagery remains a significant challenge, especially during non-bloom seasons and the study focuses on small-scale with very high resolution. This study introduces an innovative approach to this challenge, leveraging a behavior function tailored to enhance prediction accuracy for imagery with weak algal signals and the time difference between remote sensing imagery and filed measurement. By developing a behavior function that adjusts for algae fluctuation and satellite observation times during the day, the research introduces a novel approach to enhance accuracy by considering time effect for the satellite-algae model. This behavior function further enables the computation of underwater algae volume—previously challenging—providing precise estimates of regional algal quantities and seasonal variations. Focusing on the Delco and GLSM regions, the study utilizes sonde measurements and time-series PlanetScope imagery for 2022 and 2023. The analysis employs field measurements to model the diurnal vertical movement of algae, integrating this with satellite capture times to accurately determine surface Chl-a at specific moments. Key findings demonstrate the behavior function's significant role in improving regression model accuracy and underwater algae estimation.

## 5.2 Introduction

Lakes stand as vital freshwater resources, offering habitats for a diverse array of species, bolstering local economies through tourism and fisheries, and acting as essential sources of water for irrigation, industrial activities, and domestic needs (Adrian et al. 2009; Mooij et al. 2010). However, lake environment is fragile and easily disrupted by environmental changes, leading to the rapid proliferation of algae, commonly known as algal blooms (Carmichael and Boyer 2016). The proliferation of these blooms is often triggered by eutrophication, a process fueled by the influx of excessive nutrients such as nitrogen and phosphorus (Park et al. 2017). Cyanobacteria harmful algal blooms (cHABs) are significant increases in toxin-producing algae. These blooms, often consisting of cyanobacteria or blue-green algae, pose serious health risks, particularly to children, and create challenges in water treatment by affecting taste, odor, and causing clogging issues (Weirich and Miller 2014; Rajasekhar et al. 2012). However, the monitoring of harmful algal blooms within small and mid-sized lacustrine environments is hampered by insufficient spatial and temporal coverage, posing challenges to the analysis of algae's seasonal and daily fluctuations.

Utilizing satellite remote sensing for lake biomass monitoring offers an expansive and long-term observation capability. This approach leverages the distinct reflectance characteristics of satellite images to differentiate between algae and water, utilizing the unique spectral features of ground objects. The deployment of multi-spectral and hyper-spectral satellite sensors, including the Landsat series, MODIS, and Sentinel-2, has advanced the precise detection of cyanobacteria pigments in a large scale. The PlanetScope satellite constellation, comprising over 400 satellites, offers global coverage with daily or weekly observations and a 4m Ground Sample Distance (GSD), facilitating the monitoring of small to medium-sized algal dynamics in lakes through dense, time-series observations. Specifically, the concentration of chlorophyll-a (Chl-a), indicative of algal biomass, alters the spectral reflectance captured in satellite imagery. Therefore, spectral features and their combinations have become the predominant method for quantifying Chl-a concentrations and other cyanobacteria pigments, such as phycocyanin (PC). A number of spectral indices have been developed to quantitatively evaluate algae via remote sensing, such as the Normalized Difference Chlorophyll Index (NDCI, (Mishra and Mishra 2012) and the Specific Absorption Band Index (SABI, (Alawadi 2010). These indices capitalize on the unique spectral signatures of algae-related pigments, like chlorophyll-a and phycocyanin, to identify and quantify algal blooms from satellite imagery. NDCI helps in assessing the concentration of chlorophyll, a direct indicator of algal biomass, while SABI is tailored to detect specific harmful pigments associated with toxic cyanobacteria. With the advent of machine learning (ML) and deep learning (DL) technologies, which have become standard for image recognition and regression challenges, the conventional spectral indices-based monitoring methods are increasingly being enhanced or replaced by ML and DL models capable of learning features autonomously. In recent years, many ML methods have been developed to build up the regression model between spectrum features and Chl-a, including Support Vector Machines (SVM) (Cortes and Vapnik 1995; Teng Zhang, Huang, and Wang 2020), Random Forest (Ho 1995; Belgiu and Drăguţ 2016; Shen et al. 2022), and Convolutional Neural Networks (CNNs) (Pyo et al. 2019; Aptoula and Ariman 2021).

Employing the fundamentals of optical remote sensing, sensors are predominantly limited to capturing the reflectance from surface water, typically restricted to depths of less than 1 meter. Consequently, most of research for lake algae monitoring are concentrated on estimating surface concentrations of Chl-a, which represents for algal biomass. This limitation inherently restricts direct observations of algae that reside underwater or within deeper aquatic zones through satellite imagery. Nonetheless, several studies have proposed innovative techniques designed to extend the monitoring capabilities of remote sensing technologies to detect algal presence and concentrations beneath the water's surface (Cook et al. 2023; Cao et al. 2024; Pyo et al. 2019). These methodologies leverage advanced algorithms and modeling techniques, integrating them with the reflective properties captured by remote sensing instruments to infer the presence and density of subsurface algae (Lai et al. 2023; D. Liu et al. 2021).

However, algae vertical distribution is not consistent even in one day (Ganf and Oliver 1982; Hampton et al. 2014). Algae monitoring studies by using Geostationary satellite (Lou and Hu 2014; Pahlevan et al. 2020) indicate that surface Chl-a levels can vary significantly within a single day at the same location. Therefore, a diurnal vertical movement of algae, with significant shifts occurring related to time, weather, and algae type (D. Liu et al. 2021), necessitates a significant approach to accurately monitor algae via remote sensing analysis.

This movement implies that the surface Chl-a levels at the time of field measurements may differ from those at the moment of satellite image capture, due to the time discrepancy, but previous remote sensing water monitoring research did not systematically study the issue for algae diurnal vertical movement. Therefore, it's crucial to account for this movement when 1) validating regression models that correlate satellite imagery with field measurements, ensuring the surface Chl-a data align temporally, and 2) predicting Chl-a concentrations using satellite data, where the timing of image capture must be factored in to accurately estimate Chl-a levels both at the surface and underwater. Hence, conducting a quantitative analysis of the diurnal vertical movement of algae in lakes is imperative. Through a quantitative analysis of algal behavior, the spectral regression model based on remote sensing data can successfully address the issues arising from the temporal mismatch between satellite observations and field measurements within the same day. This approach enables more precise estimation of algal biomass using remote sensing imagery, enhancing the accuracy of predictions and the reliability of monitoring efforts.

Another challenge is monitoring non-bloom periods in lakes, where weak textural signatures in satellite imagery hinder effective algae detection, requires a nuanced approach. While current research predominantly focuses on algal bloom seasons to build up estimation model (Luo et al. 2023; Aptoula and Ariman 2021; Pyo et al. 2019; Teng Zhang, Huang, and Wang 2020), recognizing the significant changes and ecological impacts during non-bloom periods is equally vital to obtain a whole growing period. These conditions necessitate the development and application of sophisticated remote sensing techniques and analytical frameworks to capture those weak signals of algae dynamics from satellite imagery.

In this Chapter, we concentrate on the temporal impact of algal vertical movement on Chl-a monitoring models and introduce a novel method for quantitatively characterizing daytime algal movements, termed the 'Algal Behavior Function'. This function is instrumental in minimizing the discrepancies between field measurements and satellite observations due to time misalignment, thereby enhancing the precision of satellite-derived Chl-a regression models and facilitating more accurate estimations of total algal biomass underwater, especially for weak texture lakes. Additionally, we explore various factors influencing the Algal Behavior Function, including time of day, types of algae, and climate variables, to further understand and predict algal behavior and its implications on monitoring efforts. Our study starts by briefly introducing study area and dataset for filed measurement and remote sensing (**Section 5.3**). Then **Section 5.4** describes our approach for satellite-Chl-a model and algal behavior function in detail, includes general regression model, Algae diurnal vertical behavior, regression model with Algal Behavior Function, and accuracy assessment. **Section 5.5** summarizes quantitative experiment results adapting algal behavior function to modify satellite-Chl-a model, and the underwater Chl-a prediction based on time. **Section 5.6** discuss the indicators corresponding to algal behavior function and its benefit to algae monitoring, and a case study to evaluate in our study areas. Finally, Section 6 concludes this paper.

## 5.3 Study area and dataset

### 5.3.1 Study area

Two lakes are located in Ohio, the United State of America, have been selected as the study areas, include a big lake named Grand Lake St Marys (as **Figure 5.2** top right panel) and several small lakes named Del-Co Reservoirs (as **Figure 5.2** bottom right panel). Harmful algal blooms (HABs) in Ohio, have become a significant environmental concern. These blooms, primarily caused by excess nutrients from agricultural runoff, not only degrade water quality but also produce toxins harmful to human health and aquatic life.

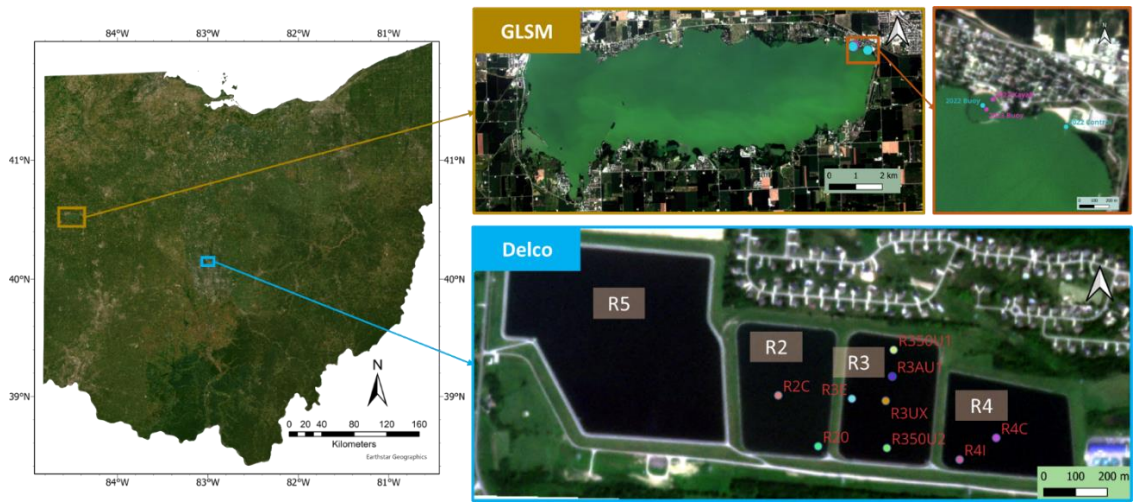

**Figure 5.2** Study areas including Grand Lake St Marys (top) and Del-Co Reservoirs (bottom), both areas locate at Ohio. GLSM area have two 24/7 sample points from April to October for both 2022 and 2023, and Del-Co area have nine sample points per week from April to October 2023

Grand Lake St Marys(GLSM), Ohio's largest man-made lake locates in the northwest part of Ohio (40°31′36″N, 84°30′03″W), as shown in **Figure 5.2** top right panel. The lake region has water surface area as $5.2 \times 10^8 m^2$ m2. The water quality in GLSM has deteriorated due to the excessive nutrient inputs from various sources in the surrounding area. These inputs have likely led to the occurrence of algal blooms, specifically cyanobacteria blooms, which are also known as blue-green algae due to their ability to photosynthesize and contain chlorophyll.

Del-Co Reservoirs locate in the central part of Ohio (40°12'09"N, 83°04'15.5"W), as shown in **Figure 5.2** bottom right panel. The lake region has water surface area as $1.05 \times 10^5 m^2$ and total volume of water as $9.5 \times 10^5 m^3$. The reservoirs are connected, with water flowing in the direction from R5 to R4, as shown in **Figure 5.2**.

### 5.3.2 Field measurement

In the GLSM region, a buoy and anchor system were utilized to secure the hydrophone for accurate sound measurements. Additionally, another anchor kept the boat stationary to avoid drifting. To ensure the signal's reliability, a reference sound field measurement was conducted adjacent to each transducer at the onset of the experiment. This meticulous setup guaranteed the precision and consistency of the sound measurements over time. Besides, water quality was conducted using an EXO2 sonde by Xylem at 1m increments, equipped with sensors for turbidity, pH, temperature/conductivity, total algae (chlorophyll-a/phycocyanin), DO, nitrate, and a central wiper. During the 2022 field season, the sonde, attached to an ultrasonic treatment unit at 1m depth, recorded data every 15 minutes from May 19 to October 3. Location tracking was performed with a fisher finder with GPS (Garmin). Calibration of the sonde varied, being annual in 2022 and monthly in 2023, ensuring accuracy across seasons.

Within the Del-Co region, weekly sonde assessments and biweekly hydrophone evaluations were conducted at designated sites. The EXO2 sonde by Xylem was employed to measure depth profiles ranging from 1m to 6m at each sampling point, with a consistent depth interval of 1m. The measurements for each sampling point were completed within a 10-minute window, ensuring that they could be considered as occurring simultaneously for analytical consistency.

In summary, the GLSM region offers continuous Chl-a rfu observations every 15 minutes, both day and night, at a depth of 1m for each sampling point. Conversely, the Del-Co region provides weekly Chl-a rfu observations at specific times, with depth profiles ranging from 1m to 6m underwater. This distinct approach in data collection between the two regions provides a comprehensive view of Chl-a concentrations over time and depth. Data from GLSM provide insights into the basic temporal patterns of algae vertical movement, while information from Del-Co reveals how the proportion of algae changes across different depths, offering a spatial perspective. This dual approach enriches our understanding of a spatiotemporal algal dynamics to develop a mathematical model as algae behavior function.

### 5.3.3 Satellite data and pre-processing

The PlanetScope satellite constellation has more than 150 satellites in orbit, resulting in an unparalleled capacity to gather nearly daily photos on a global scale (Roy et al., 2021). The imagery has a Ground Sample Distance (GSD) of 4 m per pixel, which is better in spatial resolution compared to commonly utilized global-cover optical imagery from satellites like Landsat-8 (30 m) and Sentinel-2 (10 m). In recent years, PlanetScope launched Super Dove series of remote sensing satellites, which provides 8 bands multi-spectral imagery with precise and narrow spectrum coverage for each band, including costal blue (431 - 452 nm), blue (465 – 515 nm), green I (513 - 549 nm), green (547 – 583 nm), yellow (600 - 620 nm), red, (650 – 680 nm), red edge (697 – 713 nm), and Near Inferred (NIR, 845 – 885 nm). Compare with other multi-spectral remote sensing satellites as MODIS, Sentinel-2, or Worldview series, PlanetScope have a unique advantage to monitor algae dynamics in small regions within short time.

For GLSM region, since sonde measurements provide 24/7 Chl-a data, all potential PlanetScope images without cloud can be utilized within measuring period. 24 PlanetScope images for 2022 and 19 images for 2023 have been applied, and there are 76 measurements to establish monitoring method for GLSM region. For Delco region, we adapt the imagery at the same date or nearby filed measurement date, which get 37 images for 2023 (some date more than one images), and there are 333 measurements to establish monitoring method for Del-Co region.

Given the variability among the sensors across the PlanetScope Super Dove satellites (they are not exactly same for different satellite sensors), we undertake an additional reflectance calibration step. This process is meticulously designed to amend minor discrepancies in each Top of Atmospheric Reflectance (TOAR) image, ensuring consistency and accuracy across the dataset. We manually selected several impervious structures around study areas, and then calibrate spectrum from different images by adapting image in a certain date as reference.

## 5.4 Methodology

In the methodology part, the common methods for remote sensing Chl-a estimation, including spectral indices methods which utilize the spectrum feature between each satellite band, and machine learning methods which adapt deep features from spectral bands. Next, we introduce the definition and usage of algae diurnal vertical behavior function by analyzing the field measurements of Chl-a and the time of these measurements. Finally, a behavior function refined satellite-based Chl-a estimation strategy is applied to remove the time misalignment between field measurement and remote sensing imagery, to generate a refined estimation model.

### 5.4.1 Satellite-based Chl-a estimation method

Remote sensing-based algae monitoring utilizes distinct spectral signatures to differentiate algal blooms from water by capturing and analyzing their reflectance from satellite imagery. To assess algae concentration, we developed regression models correlating field-measured Chl-a with satellite spectral reflectance. These models fall into two categories: spectral indices and machine learning methods, and the regression models for study areas are developed separately based on their datasets.

Spectral indices employ empirical second-order polynomial regressions, using PlanetScope band ratios for Chl-a estimation. The spectral indices utilize different band combinations to monitor Chl-a based on sunlight absorption differences for Red, Red Edge, and Near Infrared bands. Notable Spectral indices methods for our study include the Normalized Difference Chlorophyll Index (NDCI) (Mishra and Mishra, 2012), the Surface Algal Bloom Index (SABI) (Alawadi, 2010), the Normalized Difference Vegetation Index (NDVI) (Rouse et al. 1974), and the Enhanced Vegetation Index (EVI) (Huete et al. 2002). For detail, NDCI focuses on the ratio between the near-infrared and red bands, advantageous for identifying chlorophyll concentration in water bodies. SABI is tailored to detect algal blooms by emphasizing specific spectral features associated with algae. NDVI uses the difference between the near-infrared and red reflectance to assess vegetation health, beneficial for areas with mixed land and water. EVI2 improves on NDVI by reducing atmospheric influences, offering better sensitivity in high biomass regions. In addition, NDCI and SABI are specifically developed for sea and lake algae monitoring via remote sensing, which have been widely applied to large scale estimation (Mishra and Mishra 2012; Alawadi 2010; Hu 2009; Lou and Hu 2014). The spectral indices for NDCI, SABI, NDVI, and EVI are described as following equations:

$$NDCI = \frac{Red\ Edge - Red}{Red\ Edge + Red} \tag{5.1}$$

$$SABI = \frac{NIR - Red}{Green + Blue} \tag{5.2}$$

$$NDVI = \frac{NIR - Red}{NIR + Red} \tag{5.3}$$

$$EVI = 2.5 * \frac{NIR - Red}{NIR + 6 * Red - 7.5 * Blue + 1} \tag{5.4}$$

Machine learning regression methods for Chl-a estimation employs Random Forest (Ho 1995), Support Vector Machine (SVM) (Cortes and Vapnik 1995; Gualtieri and Cromp 1999), and Neural Networks (Specht and others 1991), each offering unique advantages in discovering the deep features among spectrum bands. For regression tasks, Random Forest aggregates predictions from multiple decision trees to output a mean prediction for continuous variables, improving accuracy and reducing overfitting by averaging the results, enhancing predictive accuracy and robustness. SVM works by identifying the hyperplane that optimally separates different class labels with the maximum margin, making it powerful for classification problems. SVM for regression, known as SVR (Support Vector Regression) (Awad et al. 2015), applies a similar principle as SVM classification but focuses on fitting the error within a certain threshold. Neural Networks leverage a layered node structure to learn from complex patterns by weighting inputs and processing them through activation functions. They excel in identifying nonlinear relationships within data, optimizing the network's predictive capabilities for continuous outcomes by minimizing errors through backpropagation, which enabling complex regression from datasets. For the input features for machine learning methods, we include all 8 bands from PlanetScope Super Dove to build up the regression models.

**5.4.2 Algae diurnal vertical behavior in lake**

Remote sensing imagery predominantly captures water surface-level Chl-a, typically within a 1-meter depth (Lai et al. 2023). However, algae's diurnal vertical and minor horizontal movements can cause discrepancies between field and satellite data, affecting regression model accuracy. To address this, we introduce the 'Algal Behavior Function,' a mathematical model focusing on the time-based description of algae's diurnal vertical movement. This approach aims to mitigate the Chl-a observation misalignment caused by these movements, enhancing model precision of both validation and prediction.

GLSM dataset provides a time-series Chl-a observation of 2022 and 2023 for water surface, from **Figure 5.3 (a)**, it shows the Chl-a rfu measurements for GLSM Buoy 2022 sample point around 2022 May 28$^{th}$ to June 8$^{th}$, suggesting the algal movement, measured 1 meter underwater, follows a pattern resembling a sine or cosine function over time. Therefore, we use a cosin representative to represent the surface Chl-a changes with time diurnally:

$$Chla_{normalized}(t) = a_0 \cos(a_1 t + a_2) + a_3 t + a_4 \qquad (5.5)$$

Where $Chla_{normalized}(t)$ is the normalized Chl-a value ($\frac{chla(t)}{chla}$) in a water surface area at time $t$ (for $t \in [0,1]$ to represent one day), and $a_0 \sim a_4$ are coefficients that represent the combination of linear function and sine function, with $a_1 = 2\pi$ assigned as a constant to represent the phase equal to one day. The coefficients $a_0 \sim a_4$ are fitted by using the time series data of one day or a certain time range, as shown in **Figure 5.4**.

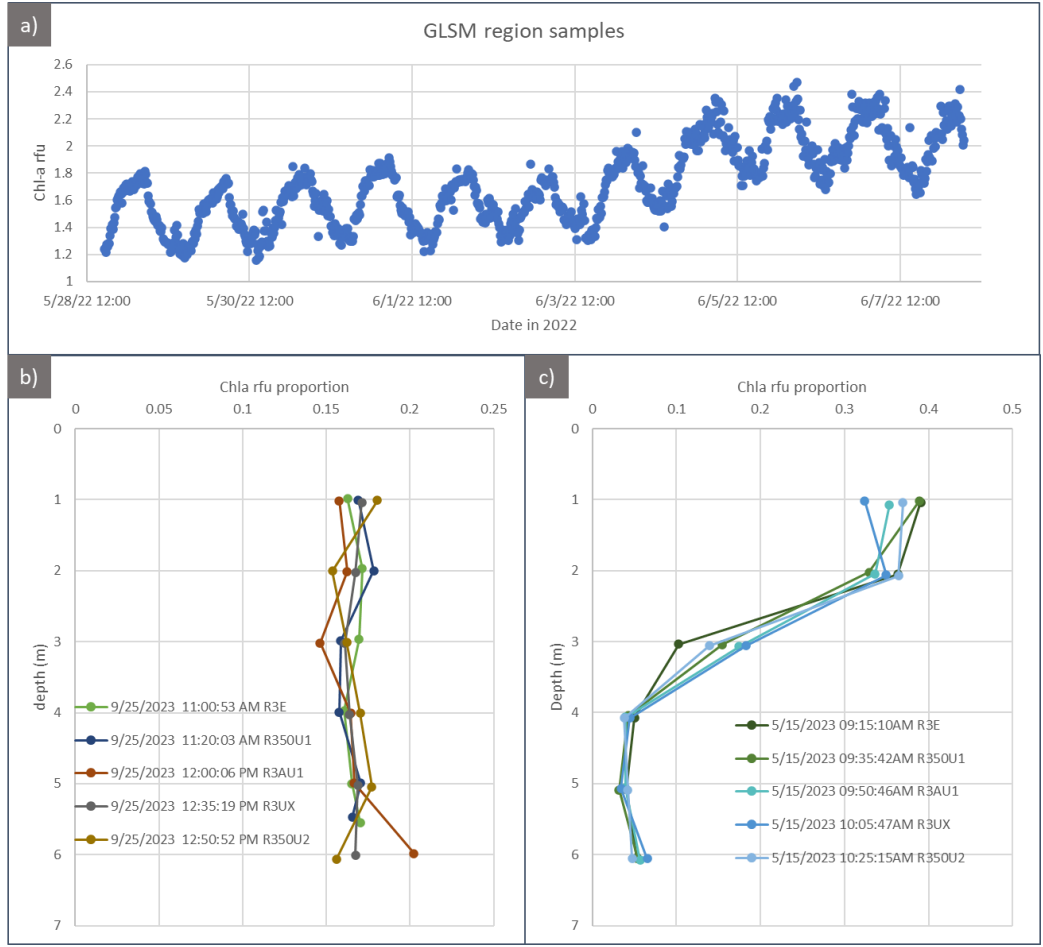

**Figure 5.3** The measured Chl-a rfu in temporal and spatial (vertical) level. **(a)** The Chl-a for GLSM region around 2022 May 28th to June 8th, show a sine or cosine function with time for algal movement at 1m depth under water surface; **(b)** The Chl-a for Del-Co region at September 25th 2023, show a vertical Chl-a distribution around noon time; **(c)** The Chl-a for Del-Co region at May 5th 2023, show a vertical Chl-a distribution around morning time

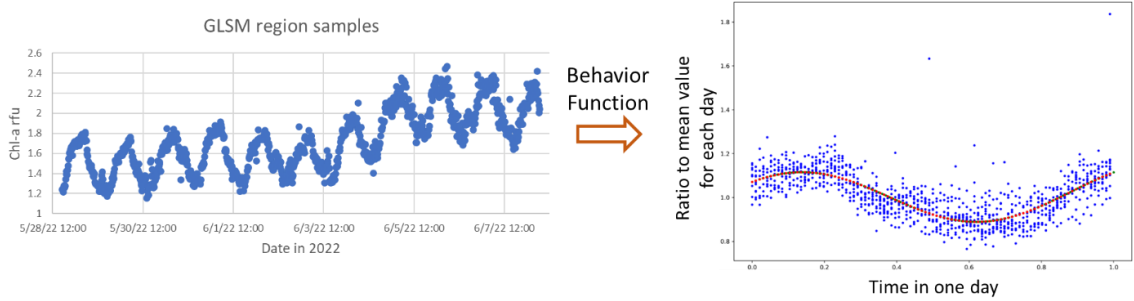

**Figure 5.4** The time-series data to build up Algal Behavior Function

The Del-Co dataset, which includes detailed Chl-a measurements from deeper water, allows us to model the relationship between surface and underwater Chl-a concentrations at any given time $t$ using the proportion of surface Chl-a, which is shown as **Figure 5.3 (b)** and **(c)**. This approach provides a quantitative method to describe how Chl-a levels at the water's surface relate to those at depth, and Algal Behavior Function can be modified as :

$$P_{chla}(t) = a_0 \cos(2\pi t + a_2) + a_3 t + a_4 \tag{5.6}$$

Where $P_{chla}$ is the proportion of surface Chl-a at time $t$. For Del-Co region, all field measurements with time are summarized to compute a set of best fitted coefficients $a_0 \sim a_4$ with Least Square algorithm, and generate an overall Algal Behavior Function to describe the surface Chl-a proportion for whole observation period.

**5.4.3 Refined satellite Chl-a estimation model with behavior function**

Once the Algal Behavior Functions are generated for study areas, it is used to reduce the Chl-a observation misalignment between field measurements and remote sensing imagery since time misalign effect. Several current research (Lai et al. 2023; D. Liu et al. 2021) have chosen the strategy to exclude data samples with a time discrepancy greater than 3 hours between field and remote observations. Nevertheless, the findings in **Section 5.4.2** indicate that algae's vertical movement within a three-hour window is still significant, suggesting that such filtering may compromise the robustness of regression analyses, particularly when the available dataset is limited.

Our Algal Behavior Functions incorporate field measurement time $t_0$ and remote sensing imagery record time $t_1$ to calculate the surface chl-a proportion at $t_1$. This method aligns time discrepancies, enabling the construction of an accurate regression model by adjusting for algae's temporal dynamics, ensuring precise modeling of Chl-a concentrations. As depicted in **Figure 5.5**, the Algal Behavior Function is effectively utilized during the prediction stage to align remote sensing imagery times with the algae's vertical movements. This alignment allows for the computation of total Chl-a at both surface and underwater levels, facilitating accurate biomass estimation. This approach is particularly advantageous for time series environmental monitoring, especially when the capturing time of remote sensing images varies significantly, ensuring a precise understanding of algal dynamics over time.

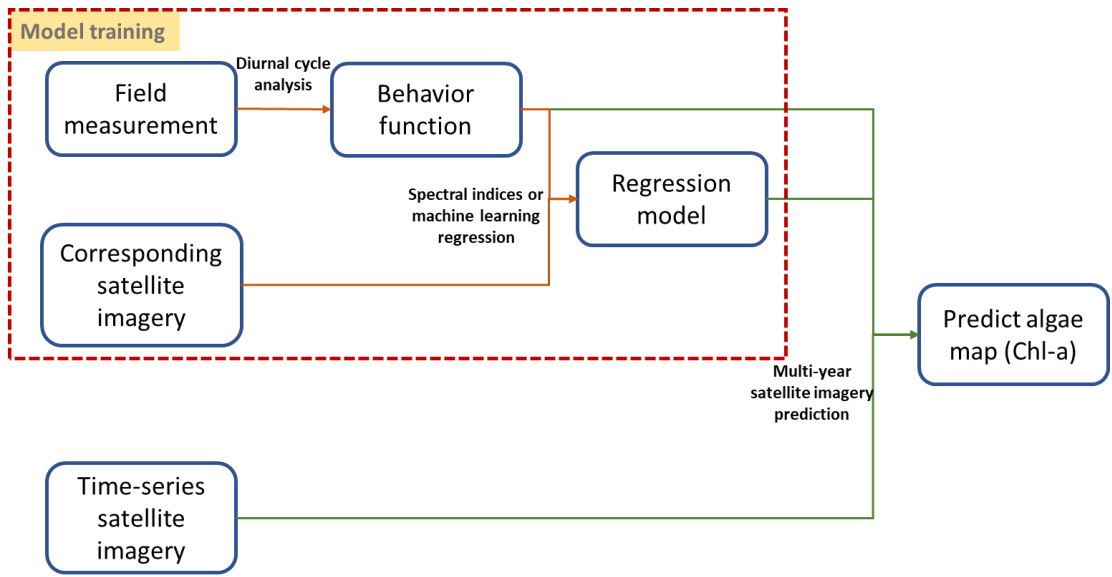

**Figure 5.5** The workflow of Algal Behavior Function utilization in regression model training and prediction for new data

### 5.5 Result

#### 5.5.1 Chl-a estimation for GLSM region

For GLSM region, four spectral indices and three machine learning methods have been applied to build up the Chl-a estimation models. For regression model establishing, 70% of data has been used for training and rest of 30% of data has been used for validation, and the validation error plots are shown in **Figure 5.6**, it indicates that machine learning methods are slightly better than spectral indices method if we adapt the original measurement data to build up the regression model, and $R^2$ for all methods are around 0.2-0.4.

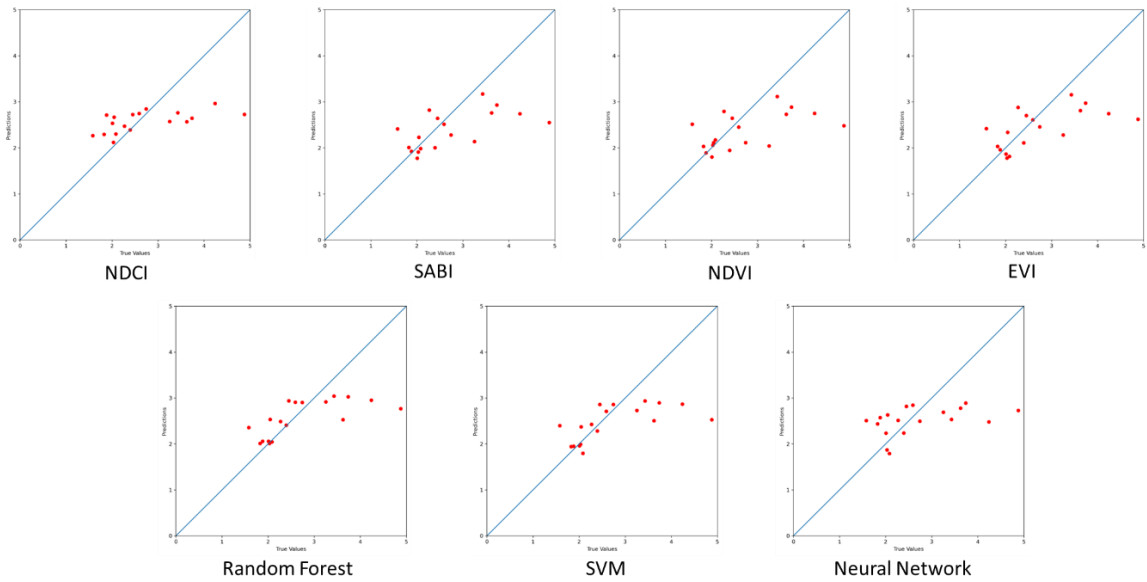

**Figure 5.6** Validation result of remote sensing imagery and Chl-a for Spectral indices methods and machine learning methods, for GLSM region

To validate the performance of adding Algal Behavior Function, we have done a comparative analysis with and without time alignment. The study comprised three groups of experiments: 1) using **noon-time** Chl-a field measurements to reflect typical satellite observation times (a consistent time) for regression models; 2) aligning Chl-a field measurements with **actual satellite observation times**; and 3) incorporating a daily fitted **behavior function** to refine Chl-a measurements. The comparison result is shown in **Table 5.1**, it shows that with algal behavior function strategy, accuracy for either spectral indices methods or machine learning methods can be improved.

**Table 5.1** The RMSE for seven regression methods by using consistent time, actual observation time, and behavior function with actual time.

| RMSE (Chl-a rfu) | | Noon time (12:00 pm) | Satellite observation time | Satellite observation time + behavior function |
|---|---|---|---|---|
| Spectral indices | NDCI | 0.550 | 0.644 | **0.420** |
| | SABI | 0.497 | 0.633 | **0.423** |
| | EVI | 0.470 | 0.611 | **0.370** |
| | NDVI | 0.560 | 0.713 | **0.477** |
| Machine learning | RF | 0.456 | 0.528 | **0.321** |
| | SVM | 0.526 | 0.611 | **0.382** |
| | NN | 0.620 | 0.730 | **0.513** |

**5.5.2 Chl-a estimation model for Del-Co region**

For Del-Co region, four spectral indices and three machine learning methods have been applied to build up the Chl-a estimation models as well as GLSM region. For regression model establishing, 70% of data has been used for training and rest of 30% of data has been used for validation, and the validation error plots are shown in **Figure 5.7**. In the Del-Co region, the presence of many weak texture signal images challenged the regression model's generation, leading to suboptimal fitting accuracy with the spectral indices methods, where $R^2$ values were below 0.1. Conversely, the application of three machine learning methods significantly outperformed the spectral indices, with Random Forest and SVM showcasing superior regression fitting and validation $R^2$ values exceeding 0.7, indicating a more robust predictive capability in weak texture water regions.

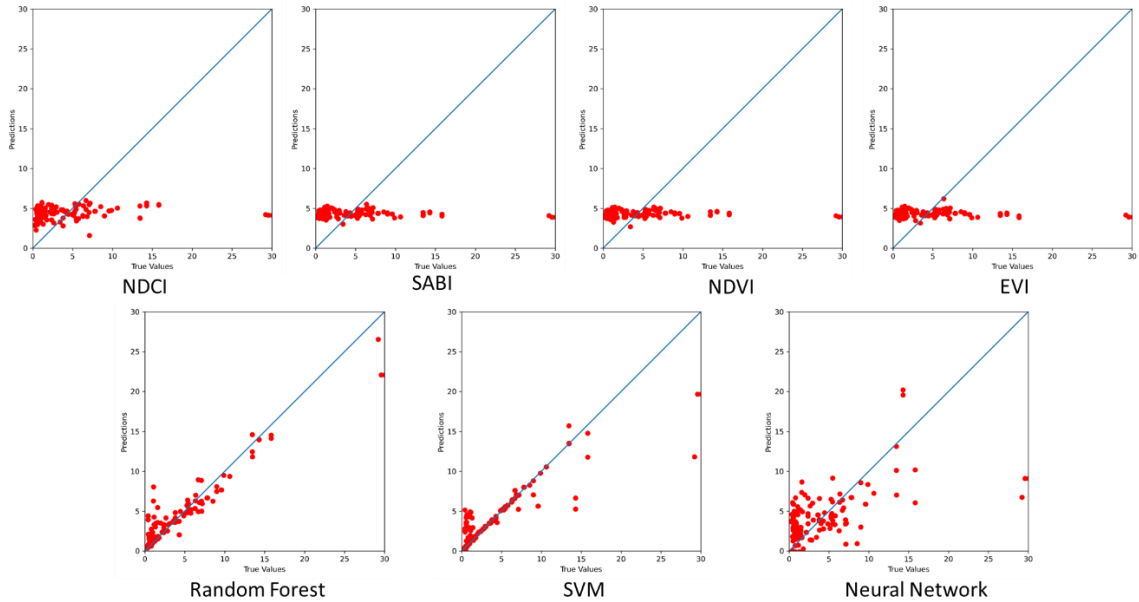

**Figure 5.7** Validation result of remote sensing imagery and Chl-a for Spectral indices methods and machine learning methods, for Del-Co region

To validate the performance of adding Algal Behavior Function, we have done a comparative analysis with and without time alignment. We use all field measurements with proportion of surface Chl-a from different time, and fit a behavior function to represent this proportion change diurnally. For the Delco region, incorporating depth data from field measurements, we designed two experiments: 1) A baseline approach, directly correlating field measurements with satellite imagery without considering temporal dynamics; 2) A refined method, integrating time with field measurements to assess surface Chl-a proportions, applying a cosine function to model algal movement, and calculating surface Chl-a at satellite observation times. This approach aims to enhance model accuracy by accounting for temporal variations in algae distribution. The experiment is shown as **Figure 5.8** and **Table 5.2**, it shows the similar result as the experiment in GLSM region, with algal behavior function strategy, result for all regression methods has been improved, even for spectral indices methods with not good regression. Besides, **Figure 5.9** shows a prediction result of using algal behavior function to estimate surface Chl-a in Del-Co region at August 23$^{rd}$ 2023.

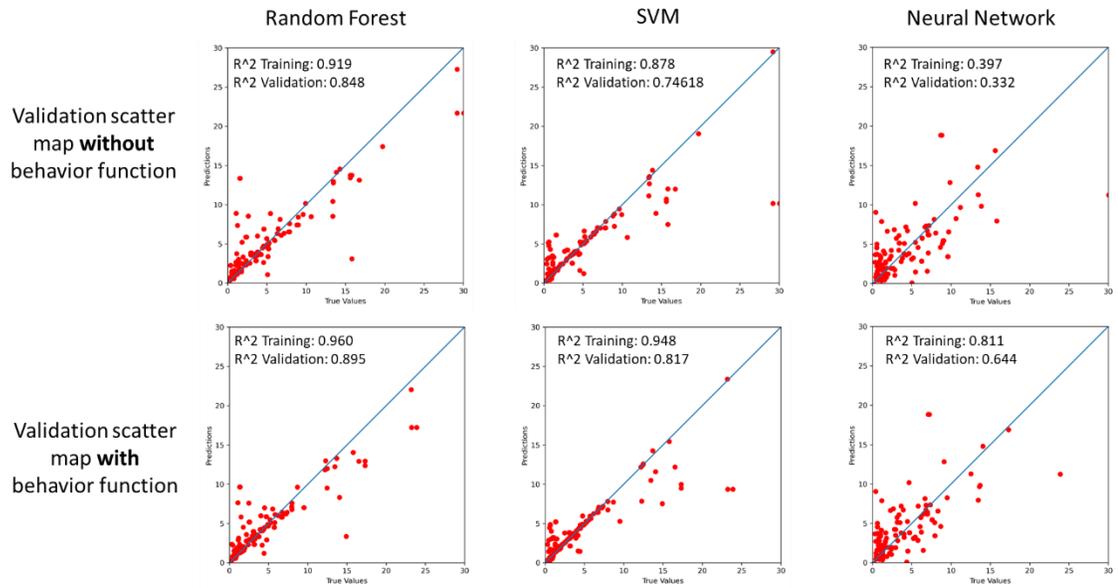

**Figure 5.8** Valid scatter map without and with Algal Behavior Function modification

**Table 5.2** The RMSE for seven regression methods by using original measurement, and measurements refined by behavior function.

| RMSE (Chl-a rfu) | | No behavior function | Behavior function |
|---|---|---|---|
| Spectral indices | NDCI | 25.958 | **18.084** |
| | SABI | 26.843 | **18.739** |
| | EVI | 26.944 | **18.814** |
| | NDVI | 26.786 | **18.699** |
| Machine learning | RF | 2.603 | **2.066** |
| | SVM | 6.095 | **3.373** |
| | NN | 17.760 | **6.568** |

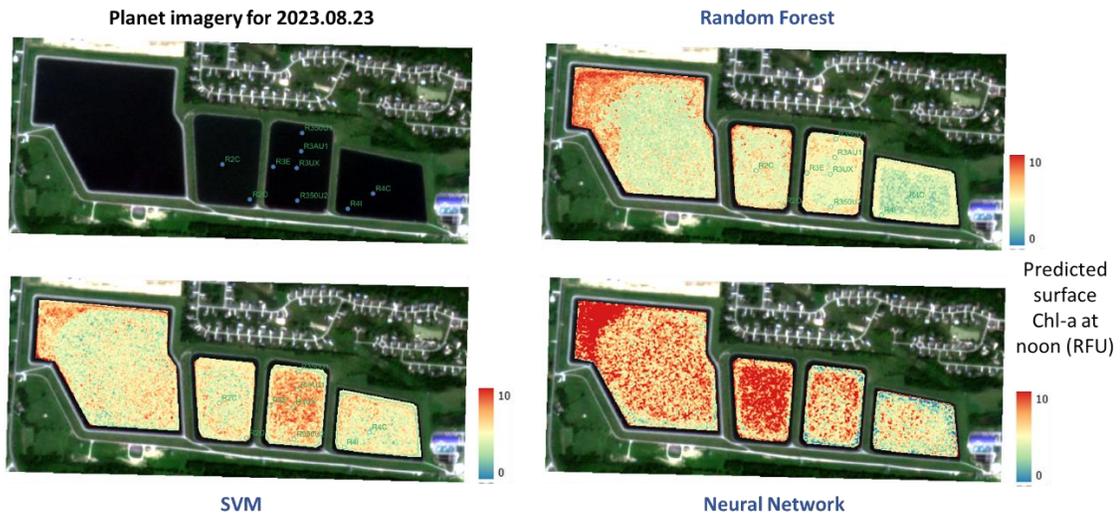

**Figure 5.9** Surface Chl-a estimation result for Del-Co region at noon (12:00 pm), August 23$^{rd}$, 2023, for machine learning methods.

**5.6 Discussion**

**5.6.1 Benefit of time alignment of field measurement and remote sensing observation**

The alignment of field measurement and remote sensing observation times significantly impacts the accuracy of algae monitoring. This approach or strategy for Algal Behavior Function can be directly added as a pre-processing step to any current method to remove the time difference effect and utilize more data to establish estimation model and improve the accuracy. Through examples like the GLSM and Del-Co regions, where models incorporating time alignment showcased superior performance, the importance of considering temporal discrepancies becomes evident. From the comparison result of GLSM region, the RMSE value for adapting actual time measurement for regression is worse than model with noon time, the reasons include 1) PlanetScope capturing time for Ohio region is around 11:00 – 12:30, which is close to noon; 2) sonde measurements in GLSM region have some noise and the validation samples contain these outliers to decrease the accuracy. While the Algal Behavior Function strategy enhances the training and prediction accuracy of satellite-based Chl-a models, it requires a substantial dataset of Chl-a depth measurements across various times. Acquiring such comprehensive data can be challenging for many researchers, highlighting a limitation in the widespread application of this approach.

**5.6.2 Implication for surface and underwater algae biomass estimation**

Leveraging the Algal Behavior Function allows for the estimation of both surface and underwater Chl-a concentrations, enhancing our ability to gauge the total Chl-a concentration throughout the water column. This method, by integrating multi-spectral satellite imagery with the precise timing of image capture, offers a comprehensive view of algal biomass. Previous research (D. Liu et al. 2021; Lai et al. 2023) utilized machine learning models and multi-spectral imagery to estimate underwater Chl-a, while the integration of multimodal input data—including both multi-spectral imagery and time through the Algal Behavior Function—significantly enhances the accuracy of total Chl-a estimation. This approach advancement facilitates a more precise quantification of Chl-a across the entire water column by using Algal Behavior Function.

### 5.6.3 Algae monitoring in weak signal remote sensing imagery

In the context of algae monitoring using remote sensing imagery with weak signals, spectral indices methods often fall short due to their limited capacity to handle nuanced variations in algal presence. Machine learning and deep learning techniques, on the other hand, show a promising improvement in regression performance. These advanced computational models can better decipher subtle differences in the data, leading to more accurate estimations of algal biomass even in challenging imaging conditions.

### 5.7 Conclusion

This study addresses the challenges of accurately quantifying chlorophyll-a concentrations from satellite imagery, particularly in non-bloom seasons with weak texture and areas requiring high-resolution analysis. By developing and applying an Algal Behavior Function that accounts for time differences and algal fluctuations, we enhance the precision of biomass estimations. Our novel approach significantly improves regression model accuracy and enables effective underwater algae biomass computation with multimodal data with remote sensing imagery and time information. The experiment, derived from comprehensive field measurements and PlanetScope imagery analysis in the Delco and GLSM regions, underscore the critical role of incorporating temporal dynamics into remote sensing methodologies for advanced environmental monitoring.

**Chapter 6. Conclusion and Future Work**

**6.1 Conclusion**

In this dissertation, we develop several novel approaches in satellite photogrammetry applications, including built-up area modeling and natural environment monitoring using stereo/multi-view satellite image-derived 3d data.

In Chapter 2, we developed a LoD-2 building reconstruction approach using high-resolution satellite images, integrating satellite-derived geospatial data for improved model accuracy, and summarized this approach into an open-source software SAT2LoD2. Our method, tested across diverse urban landscapes, demonstrated superior performance in both 2D and 3D evaluations compared to existing techniques. This work marks a significant contribution to satellite-based urban reconstruction, albeit with considerations for data resolution and building complexity.

In Chapter 3, we presented an advanced LoD2 building reconstruction approach focused on unit-level segmentation and the reconstruction of circular buildings using satellite Orthophotos and DSMs. Our method efficiently distinguishes and models both complex, closely situated buildings and circular structures within dense urban landscapes, demonstrating a clear advantage over standard building segmentation techniques in challenging environments.

In Chapter 4, we developed a photogrammetry pipeline has enabled detailed monitoring of glacier dynamics, revealing significant variations across different regions, with a novel multi-direction glacier flow tracking method. The Viedma Glacier showed a steady retreat with a 45-50 days delay in climate impact on velocity. The La Perouse Glacier experienced rapid volume loss from 2019 to 2021, followed by accumulation, with climate effects on velocity being immediate. Meanwhile, the Skamri Glacier displayed stability in volume, with a 40-45 days climate influence lag. These results highlight the nuanced responses of glaciers to climate changes, emphasizing the importance of tailored strategies for understanding and managing glacier dynamics in the face of global warming.

Lastly, in Chapter 5, we introduced an innovative approach termed 'Algal Behavior Function' to enhancing the accuracy of chlorophyll-a estimation from satellite imagery, particularly during non-bloom seasons and in high-resolution contexts. By incorporating an Algal Behavior Function that accounts for temporal dynamics and algal fluctuations, we significantly improve the precision of Chl-a estimations. Our findings from the Delco and GLSM regions highlight the importance of integrating temporal factors into environmental monitoring, demonstrating the potential for more accurate regional algal quantification and seasonal variation analysis.

## 6.2 Limitation

In Chapter 2, our approach can only deal with rectangular building, and does not perform well on very dense urban regions, and this limitation is solved via Chapter 3. Besides, the semantic segmentation is sensitive to resolution and sensors of input orthophoto.

In Chapter 3, our approach can reconstruct the complex and duplex building, but the roof classification and primitive computation can be further improved. For circular building reconstruction, it still needs an advanced circular detection method to robustly detect small circular shapes from satellite Orthophoto and building mask.

In Chapter 4, although our glacier tracking is efficient to compute the 3D dynamics, it still needs large computation and storage resources to process large scale datasets. Besides, the reasons make three study areas have different flowing and retreatment mode are still not clearly illustrated.

In Chapter 5, implementing the Algal Behavior Function, which requires sonde equipment for Chl-a measurements at various depths and times, presents a challenge for many research teams due to the need for consistent, detailed data collection. Furthermore, the necessity for lake-specific models due to varying depth information across different lakes adds an additional layer of complexity.

**6.3 Future work**

Future research for satellite photogrammetry and remote sensing application work may include:

**Roof identification and computation via Deep Learning**: Although the framework in Chapter 2 and Chapter 3 achieve a model-driven approach to compute the best-fitted roof structure, it still can use Deep learning networks to directly predict the structure and primitives for roof structure for both rectangular buildings and circular buildings.

**Complex building 3D structure recovery**: Our work in Chapter 2 and Chapter 3 can recovery the 3D structure of "L" shape building by merging candidate rectangular building, but it is challenge to recovery the shape for very complex building, this part can be studied by including more research of computer graphics or topology.

**Global glaciers surging tracking for climate change study**: Our approach in Chapter 4 provides a glacier tracking pipeline to detect the velocity and direction of glacier flow. Moreover, PlanetScope provides the global monthly Orthophoto that covers most glacier outside polar, and there is potential for using this data to map the global glacier surging and dynamics. Combining with global and local climate change data, the principle of glacier surging and retreatment can be studied at a large scale with very high resolution.

**Algae type and weather impact on Algal Behavior Function**: Currently, in Chapter 5, our approach consider the algae vertical movement is same for one study area, while algae type and climate conditions including temperature and wind speed will influence the algae distribution. Therefore, the next step is include these data and generate a more comprehensive algae vertical movement model.

*Pattern Recognition (CVPR)*, 1938–47. New Orleans, LA, USA: IEEE. https://doi.org/10.1109/CVPR52688.2022.00189.